\documentclass{article}

\usepackage{PRIMEarxiv}

\usepackage[utf8]{inputenc} 
\usepackage[T1]{fontenc}    

\usepackage{amsmath}        
\usepackage{amssymb}        
\usepackage{amsfonts}       
\usepackage{mathtools}      
\usepackage{amsthm}         
\usepackage{bm}             

\usepackage{graphicx}       
\usepackage{float}          
\usepackage{caption}        
\usepackage{subcaption}     
\usepackage{tikz}           
\usetikzlibrary{arrows.meta,positioning,shapes.geometric,calc}

\usepackage{booktabs}       
\usepackage{tabularx}       
\usepackage{multirow}       
\usepackage{adjustbox}      

\usepackage{algorithm}
\usepackage{algorithmic}

\usepackage{listings}

\usepackage{microtype}      
\usepackage{textcomp}       
\usepackage{nicefrac}       
\usepackage{siunitx}        

\usepackage{url}            
\usepackage{hyperref}       
\usepackage{xcolor}         
\usepackage[capitalize,noabbrev]{cleveref} 

\usepackage{geometry}       
\usepackage{fancyhdr}       

\usepackage{physics}        

\usepackage{cite}           
\usepackage{lipsum}         
\usepackage{pdfpages}       
\usepackage{orcidlink}      

\usepackage{array}
\newcolumntype{M}[1]{>{\centering\arraybackslash}m{#1}}

\theoremstyle{plain}

\theoremstyle{definition}

\theoremstyle{remark}

\definecolor{pBlue}{HTML}{D6E6F5}   \definecolor{pBlueE}{HTML}{6F9AC6}
\definecolor{pLav}{HTML}{E6DFF4}    \definecolor{pLavE}{HTML}{8E7DBE}
\definecolor{pYel}{HTML}{FFF1C1}    \definecolor{pYelE}{HTML}{C9A94A}
\definecolor{pPeach}{HTML}{FDE0CC}  \definecolor{pPeachE}{HTML}{D08B5E}
\definecolor{pGreen}{HTML}{D8EFD9}  \definecolor{pGreenE}{HTML}{6FAE72}
\definecolor{pPink}{HTML}{F8D7E1}   \definecolor{pPinkE}{HTML}{C9748F}
\definecolor{pGray}{HTML}{E9ECEF}   \definecolor{pGrayE}{HTML}{8A939C}
\definecolor{pInk}{HTML}{3F4A5A}

\title{Can AI Mature a Model's Geometry?\\Training-Free Weight Baking in LLMs}

\makeatletter
\newcommand{\affilmark}[1]{\textsuperscript{#1}}
\makeatother

\author{
  \normalfont \textbf{Amin Golnari}\,\text{\orcidlink{0000-0002-3841-0588}}\affilmark{a}
  \and
  \normalfont \textbf{Saeid Sanei}\,\text{\orcidlink{0000-0002-1446-5744}}\affilmark{b}
}

\begin{document}
\maketitle

\footnotetext[1]{%
  Corresponding author. Email: \texttt{amin.g@vinuni.edu.vn}. \\
  College of Engineering \& Computer Science, VinUniversity, Hanoi, Vietnam.
}

\footnotetext[2]{%
  Email: \texttt{saeid.s@vinuni.edu.vn}. \\
  College of Engineering \& Computer Science, VinUniversity, Hanoi, Vietnam. \\
  Also affiliated with College of Health Science, VinUniversity, Hanoi, Vietnam,
  and the Institute of Psychiatry, Psychology \& Neuroscience,
  King's College London, UK.
}

\begin{abstract}
  Can an AI make a trained language model more mature, in the sense of a better organized latent space, by looking only at the geometry of that space? An analyst model reads a fixed-size digest of topological descriptors of the layers of a frozen target and proposes a transformation, which closed-form ridge regression writes into the feed-forward weights without gradient training. Using lightweight models, 126 runs show that category purity rose on unseen words and sentences, benchmark accuracy stayed within sampling error, and the analyst matched a category-aware heuristic with gentler edits. In a five-model round robin approach, 16 pairs raised purity and four left the target unchanged. The results provide a baseline for future work on LLM-guided, training-free model editing without using raw weights.
\end{abstract}

\keywords{Latent geometry \and LLMs \and Topological data analysis \and Weight baking}

\section{Introduction}
\label{sec:intro}

The hidden feature spaces of language models have structure that a flat view hides: representations occupy a narrow cone~\cite{ethayarajh2019how}, and the intrinsic dimension changes with depth~\cite{ansuini2019intrinsic}. These studies diagnose the geometry; they do not repair it. Steering shifts hidden states at inference time~\cite{turner2023steering}, and closed-form weight edits change a single fact~\cite{meng2022locating} or erase a single linear concept~\cite{belrose2023leace}. Weight-Space Programming lets a frontier large language model (LLM) read and write raw weights as text, but a 1.5-million-parameter network already exceeds the model's context~\cite{takehi2026weightspace}.

We ask whether an AI can make a trained model more \emph{mature} by examining only a compact topological and geometric description of its latent space, and whether the result can be written into the weights without gradient descent. The latent space of a layer is the space of hidden representations that it produces for words. We call it more mature when words of the same concept sit closer together, different concepts are better separated, and fewer directions dominate, while the model's behavior is preserved. This concerns the geometry of the hidden representations, not of the weights, which are only where the change is written. We measure it on words and sentence frames that the editor never saw; whether it also improves behavior is a separate question.

Our approach, \emph{topology-aware weight baking}, splits the work. A fixed-size digest of topological descriptors tells an LLM \emph{analyst} where one layer of a frozen \emph{target} model is flawed, and the analyst decides what to change; closed-form ridge regression then computes the weight change (Figure~\ref{fig:concept}). Analyst and target are different models of a pool of five lightweight models (1.5 to 3.8 billion parameters), which let us study the idea efficiently on a single local GPU with fast experiment turnaround. Our contributions are as follows.

\begin{enumerate}
\item A training-free pipeline from topological descriptors to LLM-proposed, doubly verified, closed-form weight edits, whose analyst input does not grow with target size (Section~\ref{sec:method}).
\item A round-robin protocol with decoupled analyst and target roles: a main study with a heuristic, a parallel control, and ablations, and a study in which each of five models analyzes the other four (Sections~\ref{sec:exp} and~\ref{sec:results}).
\item Evidence of feasibility with lightweight models (Section~\ref{sec:results}): purity on unseen words in unseen frames rose in every default run and in every round-robin pair that changed its target, and the analyst matched the heuristic with gentler edits and depended on its descriptors. We also report costs and failures.
\end{enumerate}

Training or finetuning LLMs requires large curated datasets and substantial compute. To test whether an LLM can help improve a model from compressed topological descriptors alone, we instead worked within our hardware and time limits, using lightweight open-source models and a small dataset. Even with such lightweight analysts, the edits baked into the weights produced gains that were modest but consistent in direction, transferred to unseen vocabulary and sentence frames, and stayed within a divergence budget. Code and data are provided as a baseline for future work.

\section{Related work}
\label{sec:related}

On the geometry and topology of representations, Naitzat et al.~\cite{naitzat2020topology}, Gardinazzi et al.~\cite{gardinazzi2025persistent}, and Fay et al.~\cite{fay2026shape} study topological features across layers and under adversarial inputs, and Kushnareva et al.~\cite{kushnareva2021artificial} and Tulchinskii et al.~\cite{tulchinskii2023intrinsic} use topological and intrinsic-dimension features to detect generated text. Intrinsic dimension first rises and then falls with depth~\cite{ansuini2019intrinsic,valeriani2023geometry}. Cheng et al.~\cite{cheng2025emergence} find a high-dimensional abstraction phase, and Joshi et al.~\cite{joshi2025geometry} a consistent pattern across 28 open models. Gao et al.~\cite{gao2019representation} describe representation degeneration into a narrow cone, Ethayarajh~\cite{ethayarajh2019how}, Razzhigaev et al.~\cite{razzhigaev2024shape}, and Cai et al.~\cite{cai2021isotropy} profile anisotropy and isotropy across layers and clusters, and Rajaee and Pilehvar~\cite{rajaee2021cluster} propose cluster-based isotropy enhancement. Park et al.~\cite{park2024linear} formalize the linear representation hypothesis and extend it to hierarchical concepts~\cite{park2025geometry}, Engels et al.~\cite{engels2025not} show multi-dimensional features, and Marks and Tegmark~\cite{marks2024geometry} report linear structure in true and false statements. These results motivate our category descriptors and factual probe (Section~\ref{sec:data}). Kornblith et al.~\cite{kornblith2019similarity} introduce centered kernel alignment (CKA), which we use. These works measure geometry; none of them modifies the model.

Turning to gradient-free editing and steering, Li et al.~\cite{li2023inference}, Turner et al.~\cite{turner2023steering}, and Rimsky et al.~\cite{rimsky2024steering} shift activations at inference time along probe or contrasting-prompt directions, and Zou et al.~\cite{zou2023representation} and Beaglehole et al.~\cite{beaglehole2025toward} read and control concepts through representation engineering. Wu et al.~\cite{wu2024reft} learn interventions on frozen representations, and a companion benchmark~\cite{wu2025axbench} compares steering with prompting and finetuning. Raval et al.~\cite{raval2026curveball} report that linear steering behaves inconsistently and propose a kernel-PCA intervention, and Oozeer et al.~\cite{oozeer2026riemannian} steer along geodesics. In weight space, Belrose et al.~\cite{belrose2023leace} remove a linear concept in closed form, and Meng et al.~\cite{meng2022locating} change a factual association by a rank-one update.

On LLM-driven weight editing and the positioning of our work, Takehi et al.~\cite{takehi2026weightspace} let a frontier LLM read and write raw weights as text. We give the analyst a fixed-size digest instead. To our knowledge, no previous work lets one language model edit the global geometry of another from a compressed description and writes the edit into the weights. Unsupervised isotropy corrections~\cite{rajaee2021cluster} are category-agnostic, so our control is a category-aware deterministic heuristic under identical limits. Appendix~\ref{app:related} compares the closest works attribute by attribute.

\section{Method}
\label{sec:method}

\subsection{Overview and notation}
\label{sec:notation}

\begin{figure*}[t]
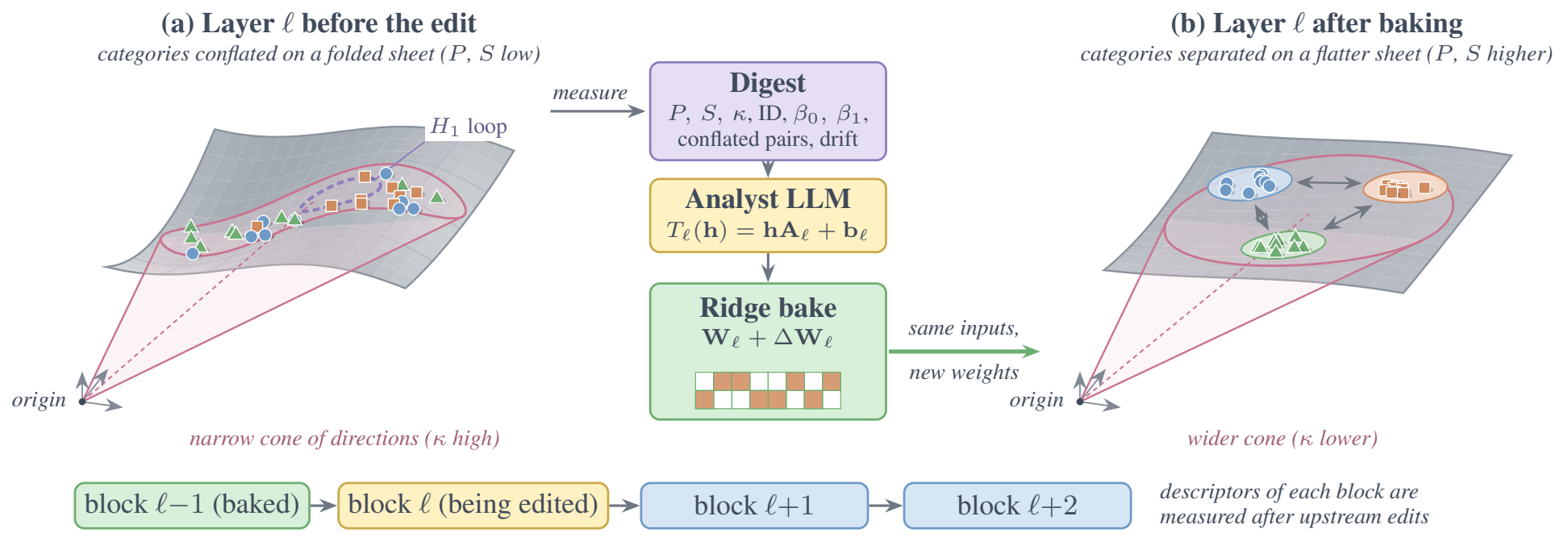

\centering
\resizebox{0.78\textwidth}{!}{%

}
\caption{Conceptual illustration of the proposed approach. (a)~Before the edit, the probe representations of block $\ell$ lie on a folded sheet: the three categories are intermixed, the directions fall in a narrow cone (high anisotropy $\kappa$), and a spurious loop ($H_1$) is present. The analyst answers with an affine map $T_\ell$, which closed-form ridge regression writes into the feed-forward weights (shaded cells: changed entries). (b)~For the same inputs, the baked block gives separated categories on a flatter sheet and a wider cone, without any temporary intervention. Bottom: blocks are edited in order, each described after all upstream edits are baked. ID: intrinsic dimension.}
\label{fig:concept}
\end{figure*}

The core idea is a division of labor. The analyst LLM decides only \emph{what} to change and \emph{where}: which operations to apply and which category pairs to pull apart. The system then computes the weight update $\Delta\mathbf{W}$ in closed form, with a single ridge-regression solve per layer (no gradients, no iteration), and processes the layers one after another (Figure~\ref{fig:concept}). The data flow has five steps.
\begin{enumerate}
\item The \emph{descriptors} (Table~\ref{tab:desctable}) summarize the geometry of one layer and are computed from the last-token hidden states of probe words in a fixed template.
\item They are sent to the analyst as a fixed-size text \emph{digest}, one message per layer.
\item We ask the analyst for at most four operations in a fixed format (JSON).
\item The operations are mapped to one affine map $T(\mathbf{h})=\mathbf{h}\mathbf{A}+\mathbf{b}$ of the residual stream.
\item Ridge regression writes the map into the feed-forward output projection in closed form.
\end{enumerate}
A frozen decoder has $L$ blocks and residual width $d$ (Figure~\ref{fig:pipeline}, Appendix~\ref{app:moved}, shows the cycle in detail). Let $\mathbf{W}_\ell\in\mathbb{R}^{d\times d_k}$ be the output projection of the feed-forward module of block $\ell$, and $\mathbf{X}_\ell\in\mathbb{R}^{n\times d}$ the last-token hidden states of the $n$ probe words ($C$ categories, labels $c_i$) after block $\ell$ of the current, already edited, target. $P$, $S$, and $\kappa$ denote purity, silhouette, and anisotropy.

As a running example, suppose the digest shows \emph{tool} and \emph{vehicle} as the most conflated category pair. The analyst can propose a centroid-axis scaling $T_{\mathrm{sep}}$ (Section~\ref{sec:ops}), which stretches the space along the line joining their centroids; the heuristic arm makes such edits. At the strict gain threshold, the analyst's accepted edits mostly used principal-component rescaling $T_{\mathrm{pcs}}$ (Appendix~\ref{app:ops}). In one recorded edit (Qwen2.5-1.5B, seed 2, layer 2), the system tested such a rescaling through a temporary intervention on the layer output during the forward pass, baked it by ridge regression, and repeated the tests. On that layer's probe words, purity rose from 0.578 to 0.584 and silhouette from 0.031 to 0.037 (Appendix~\ref{app:prompt}). The next layer was then analyzed on the edited model.

\subsection{Descriptors and edit language}
\label{sec:descriptors}
\label{sec:ops}

Table~\ref{tab:desctable} lists the descriptors, computed from $\mathbf{X}_\ell$ with cosine distance (full definitions in Appendix~\ref{app:descdefs}).

\begin{table*}[t]
\centering
\caption{Descriptors in the digest. $K=8$ nearest neighbors under cosine distance; full definitions in Appendix~\ref{app:descdefs}.}
\label{tab:desctable}
\small
\setlength{\tabcolsep}{4pt}
\begin{tabular}{>{\raggedright\arraybackslash}p{3.3cm}>{\raggedright\arraybackslash}p{6.4cm}>{\raggedright\arraybackslash}p{6.6cm}}
\toprule
Descriptor & Plain-language meaning & How it is computed \\
\midrule
Purity $P$ & How many of a word's nearest neighbors belong to its own category; high means clean groups. & Mean, over words, of the share of the $K$ nearest neighbors with the same label. \\
Silhouette $S$ & How much closer a word is to its own category than to the other categories. & Mean silhouette under cosine distance. \\
Anisotropy $\kappa$ & How narrow a cone the vectors fill; high means a few directions dominate. & Mean pairwise cosine similarity of the word vectors. \\
Category-pair conflation & How often words of two categories are each other's neighbors; high means the two are mixed. & $\mathrm{cross}(a,b)=\tfrac12(\rho_{ab}+\rho_{ba})$, where $\rho_{ab}$ is the mean share of the neighbors of words in $a$ that belong to $b$. \\
Intrinsic dimension & How many independent directions the point cloud actually uses. & Two-NN estimate $N/\sum_i\ln(r_i^{(2)}/r_i^{(1)})$ from first- and second-neighbor distances. \\
Betti number $\beta_0$ & How many separate pieces the point cloud falls into. & Connected components in the persistent homology of the cosine-distance filtration, at the 25th percentile of pairwise distances. \\
Betti number $\beta_1$ & How many loops the point cloud contains. & One-dimensional holes in the same filtration and at the same scale. \\
\bottomrule
\end{tabular}
\end{table*}

The digest has four cumulative levels, which the descriptor ablation (Section~\ref{sec:descablation}) switches on one at a time: (i)~$P$ and $S$, the six most conflated category pairs, the purity of each category, and the eight least pure words; (ii)~$\kappa$, the mean norm, and the first six spectral components; (iii)~the intrinsic dimensions and $\beta_0$, $\beta_1$; and (iv)~the intra- to inter-category distance ratio and the drift from upstream edits. The last five accepted edits and the remaining budgets are always shown, and the digest length depends on the number of categories, not on the parameter count of the target.

The analyst's reply holds at most four operations: centroid-axis scaling $T_{\mathrm{sep}}$ scales the space along the line joining two category centroids, principal-component rescaling $T_{\mathrm{pcs}}$ rescales listed principal components, partial spectral whitening $T_{\mathrm{wh}}$ equalizes the variance of the top components, and partial mean subtraction $T_{\mathrm{ctr}}$ subtracts part of the mean vector. The operations compile into one affine map,
\begin{equation}
T_\ell(\mathbf{h})=\mathbf{h}\mathbf{A}_\ell+\mathbf{b}_\ell,
\label{eq:affine}
\end{equation}
\noindent where $\mathbf{A}_\ell\in\mathbb{R}^{d\times d}$ is the product of the operations and $\mathbf{b}_\ell\in\mathbb{R}^{1\times d}$ their accumulated offset. The system tunes the overall strength $\alpha$ over six values from $0.25$ to $1.5$. Appendix~\ref{app:formal} gives every operation, the composition rule, the norm restoration, and the strength scaling.

\subsection{Closed-form baking}
\label{sec:bake}

An accepted map is written into $\mathbf{W}_\ell$. Let $\mathbf{K}\in\mathbb{R}^{m\times d_k}$ collect the inputs of the output projection and $\mathbf{R}\in\mathbb{R}^{m\times d}$ the residual change $\tilde{T}_\ell(\mathbf{H})-\mathbf{H}$ requested by the map, both over $m$ calibration tokens (Appendix~\ref{app:calib}). The update solves the ridge problem
\begin{equation}
\Delta\mathbf{W}_\ell=\arg\min_{\mathbf{\Delta}}\;\lVert\mathbf{K}\mathbf{\Delta}^{\top}-\mathbf{R}\rVert_F^{2}+\tilde{\lambda}\,\lVert\mathbf{\Delta}\rVert_F^{2},
\label{eq:ridge}
\end{equation}
\noindent where $\tilde{\lambda}$ is a ridge weight scaled by $\mathrm{tr}(\mathbf{K}^{\top}\mathbf{K})/d_k$. The problem has a closed-form solution (Appendix~\ref{app:methoddetails}). The smallest ridge weight that keeps the relative weight change $\lVert\Delta\mathbf{W}_\ell\rVert_F/\lVert\mathbf{W}_\ell\rVert_F$ at or below 0.10 is kept; if none does, stage two (Section~\ref{sec:accept}) moves to a weaker strength, and the edit is rejected if no strength fits.

\subsection{Two-stage verification}
\label{sec:accept}

A proposal is first simulated on $\mathbf{X}_\ell$ at every strength $\alpha$, and strengths within the size and CKA limits whose gain $\Delta P+\Delta S$ reaches 0.01 are ranked by gain. \emph{Stage one} tests them, strongest gain first, as an intervention test with the weights untouched. \emph{Stage two} bakes the first passing strength and repeats every test on the baked model. If a limit fails, the weights are reverted and the edit is baked again at half and at a quarter of its strength. Table~\ref{tab:limitsmain} lists the main limits (full list: Table~\ref{tab:limits}, Appendix~\ref{app:methoddetails}). The divergence, perplexity, and CKA limits are measured on a guard set of 24 held-out sentences, disjoint from the evaluation data (Section~\ref{sec:data}). A nonzero KL is expected, because an edit that changes the representations of a layer also changes the output distribution. We therefore treat KL as a budget, not as a quantity to minimize: an edit is accepted while its KL stays at or below 0.05 and the cumulative KL at or below 0.30 nats per token (Table~\ref{tab:limitsmain}).

\begin{table}[t]
\centering
\caption{Main acceptance limits, checked in the intervention test and on the baked model (the weight budget on the baked update only). KL: Kullback--Leibler divergence to the original model, in nats per token.}
\label{tab:limitsmain}
\small
\setlength{\tabcolsep}{4pt}
\begin{tabular}{lcc}
\toprule
Quantity & Per edit & Cumulative \\
\midrule
KL to the original model & $\le 0.05$ & $\le 0.30$ \\
Guard-set perplexity & $\le +2\%$ & $\le +5\%$ \\
CKA to the original model & $\ge 0.95$ & -- \\
Mean relative feature change & $\le 0.20$ & -- \\
Relative weight change & $\le 0.10$ & -- \\
\bottomrule
\end{tabular}
\end{table}

\subsection{Sequential loop and controls}
\label{sec:loop}

Editing block $\ell$ changes the input of block $\ell+1$, so the loop recomputes all descriptors on the target that contains the edits baked so far (Algorithm~\ref{alg:loop}, Figure~\ref{fig:pipeline}, Appendix~\ref{app:moved}); in our runs no layer was skipped and no budget was exhausted.

Each \emph{arm} produces its own edited models from the same original weights under the same limits. In the \emph{LLM arm}, only the analyst's proposals are applied. It proposes up to three times per layer and, after a rejection, receives the measured numbers of its proposal (Appendix~\ref{app:prompt}). The \emph{heuristic arm} uses no language model: a deterministic rule stretches the most conflated category pairs. The \emph{hybrid arm} falls back on the heuristic when the analyst fails (Appendix~\ref{app:extended}). In every arm, if no proposal passes, the best stored weak proposals are re-tested at a relaxed gain threshold. All other limits are never relaxed (Appendix~\ref{app:methoddetails}). The \emph{parallel control} removes the sequential loop and nothing else: every layer is analyzed, tested by intervention, and fitted on the original model, all weight updates are added together, and the result is evaluated once.

\section{Experimental setup}
\label{sec:exp}

\subsection{Round-robin protocol and models}
\label{sec:models}

In the round-robin protocol, one model of the pool is loaded read-only as the analyst and proposes the edits for every target of a round. A model is never its own analyst, so every experiment is cross-model. In a control round a deterministic heuristic replaces the analyst under the same targets, seeds, and limits.

The five lightweight instruction-tuned models allow fast turnaround on a single local GPU, an NVIDIA H100 with 80\,GB of memory (Table~\ref{tab:models}, Appendix~\ref{app:moved}). The \emph{main study} uses Gemma-2-2B~\cite{gemma2024gemma2}, with 2.6 billion parameters, as analyst and Qwen2.5-1.5B~\cite{qwen2024qwen25} and SmolLM2-1.7B~\cite{allal2025smollm2} as targets. The \emph{round-robin study} adds Llama-3.2-3B~\cite{llama2024herd} and Phi-3-mini~\cite{abdin2024phi3}, and every model analyzes the other four (20 runs; default LLM arm, seed 0; Appendix~\ref{app:rr}).

The main study comprises 126 runs: every configuration (arm, target, mode, digest level, strength, gain threshold, number of categories) was run with seeds 0, 1, and 2, and seven sub-studies share these runs (Appendix~\ref{app:design}): the default configuration of each arm (18 runs), the parallel control, and ablations of the descriptors, strength, gain threshold, and number of categories. The default is the sequential loop with adaptive strength, the full digest, a gain threshold of 0.01 with the relaxed stage, and eight categories. All runs completed but one fixed-strength run accepted no edit.

\subsection{Data and transfer conditions}
\label{sec:data}

The probe words are frequency-ranked WordNet lemmas~\cite{miller1995wordnet} of eight categories (animal, fruit, vehicle, tool, color, country, emotion, profession), extended to 16 categories for one ablation. For each seed, 20\% of the words of every category are set aside as \emph{unseen words}; edits are chosen and fitted on the other words only. Three further templates are never used for editing (Appendix~\ref{app:data}). Geometry is evaluated under four conditions: $[A]$ probe words in the probe template (in-sample), $[B]$ unseen words in the probe template, $[C]$ probe words in the unseen templates, and $[D]$ unseen words in the unseen templates (full out-of-distribution transfer).

Text sentences come from WikiText-103~\cite{merity2017pointer}: calibration sentences for the ridge fit, the 24-sentence guard set, and 476 held-out sentences used for evaluation only. Behavior is measured with a category cloze task and with three benchmarks of 200 examples each, redrawn per seed: HellaSwag~\cite{zellers2019hellaswag}, ARC-Easy~\cite{clark2018think}, and TruthfulQA~\cite{lin2022truthfulqa}. A factual-preference probe has 40 prompts with a true and a plausible false completion; its \emph{truth margin} is the mean log-probability difference between the two (Appendix~\ref{app:methoddetails}).

\subsection{Metrics and reporting}
\label{sec:metrics}

The changes $\Delta P$, $\Delta S$, and $\Delta\kappa$ of a run are differences (edited minus original) averaged over the layers edited in that run. We also report CKA~\cite{kornblith2019similarity} to the original model on probe words, unseen words, and guard-set tokens, the cumulative KL divergence on the guard set, held-out perplexity, and the Betti numbers and intrinsic dimension of the original and edited model (Appendix~\ref{app:metricdefs}). A statement about an arm gives the range over its three seeds, and a comparison of two arms counts seed-matched pairs. A seed redraws the probe/unseen split and the benchmark samples on the same frozen weights, so seeds are not independent replications and the tests we quote are descriptive (Appendix~\ref{app:stats}).

\section{Results}
\label{sec:results}

\subsection{The analyst matured the geometry of unseen material}
\label{sec:main}

Table~\ref{tab:headline} summarizes each arm per target as the range over its three seeds; every run is in Table~\ref{tab:main} (Appendix~\ref{app:moved}).

\begin{table*}[t]
\centering
\caption{Default configuration and two tuned heuristic configurations per target: range over the three seeds of the per-run values of Tables~\ref{tab:main}, \ref{tab:beh}, \ref{tab:ablstr}, and~\ref{tab:ablgain}. $\Delta P$, $\Delta S$: changes under $[D]$, averaged over the edited layers. CKA$_w$: CKA to the original model on unseen words. KL: cumulative KL (nats per token). $\Delta\mathrm{PPL}_h$: held-out perplexity change (\%). $\max\lvert\Delta\mathrm{Acc}\rvert$: largest absolute accuracy change on HellaSwag, ARC-Easy, and TruthfulQA. The unedited model has $\Delta=0$ and CKA $=1$; the hybrid arm is in Table~\ref{tab:main}. $^{\dagger}$Heuristic-only ablation configurations.}
\label{tab:headline}
\small
\setlength{\tabcolsep}{4pt}
\begin{adjustbox}{max width=\textwidth}
\begin{tabular}{llcccccc}
\toprule
Target & Arm & $\Delta P$ $\uparrow$ & $\Delta S$ $\uparrow$ & CKA$_w$ $\uparrow$ & KL $\downarrow$ & $\Delta\mathrm{PPL}_h$ $\downarrow$ & $\max\lvert\Delta\mathrm{Acc}\rvert$ $\downarrow$ \\
\midrule
\multirow{4}{*}{Qwen} 
 & LLM arm & 0.0196--0.0287 & 0.0235--0.0484 & 0.897--0.935 & 0.086--0.104 & 6.45--8.72 & $\le$0.030 \\
 & Heuristic (default) & 0.0162--0.0285 & 0.0110--0.0158 & 0.812--0.849 & 0.108--0.115 & 8.05--8.26 & $\le$0.015 \\
 & Heuristic, $\alpha=0.5^{\dagger}$ & 0.0275--0.0293 & 0.0169--0.0210 & 0.947--0.951 & 0.022--0.028 & 1.68--2.05 & -- \\
 & Heuristic, gain $\ge$0.02, no relaxed$^{\dagger}$ & 0.0261--0.0357 & 0.0133--0.0199 & 0.924--0.944 & 0.026--0.043 & 1.21--2.77 & -- \\
\midrule
\multirow{4}{*}{SmolLM2} 
 & LLM arm & 0.0104--0.0117 & 0.0163--0.0266 & 0.902--0.962 & 0.083--0.099 & 6.50--8.17 & $\le$0.015 \\
 & Heuristic (default) & 0.0087--0.0150 & 0.0173--0.0219 & 0.853--0.898 & 0.112--0.121 & 4.71--9.22 & $\le$0.040 \\
 & Heuristic, $\alpha=0.5^{\dagger}$ & 0.0146--0.0176 & 0.0090--0.0153 & 0.953--0.969 & 0.089--0.130 & 4.73--8.01 & -- \\
 & Heuristic, gain $\ge$0.02, no relaxed$^{\dagger}$ & 0.0097--0.0219 & 0.0095--0.0187 & 0.940--0.976 & 0.030--0.076 & 1.29--5.46 & -- \\
\bottomrule
\end{tabular}
\end{adjustbox}
\end{table*}

In terms of geometry, in all six default runs, the analyst's edits raised category purity on unseen words in unseen templates, which neither the analyst nor the ridge fit ever saw (Table~\ref{tab:headline}), more at the last layer than on average (Table~\ref{tab:main}). Silhouette rose and anisotropy fell in every run, and purity under $[D]$ rose at 150 of the 156 layers of the six edited models (intervals: Table~\ref{tab:stats}).

In topology as well, the Betti numbers $\beta_0$ and $\beta_1$ of the unseen words fell in all six runs (Table~\ref{tab:topo}, Appendix~\ref{app:traj}): the point cloud became less fragmented and lost loops (Figure~\ref{fig:concept}). The intrinsic dimension did not change consistently.

Beyond the edited layers, maturing propagates through depth. Edits also raised layers that were never edited: on Qwen, the purity of a layer before its own decision was higher than in the unedited model in all 80 decisions that followed a first edit, and on SmolLM2 in 57 of 66 (Figure~\ref{fig:cascade}, Appendix~\ref{app:moved}). In every run of every arm, the purity gain in the last third of the layers exceeded that in the first third (Table~\ref{tab:traj}).

Moreover, the change is in the weights. No intervention remains: every edit was written into one feed-forward output projection, with a relative weight change below the budget of 0.10 and a close ridge fit (median $R^2$ of 0.9995; on SmolLM2, 3 of the analyst's 44 edits had $R^2<0.9$ and still passed every limit; Table~\ref{tab:decisions}).

Behavior was largely retained within the divergence budget: the analyst’s final models kept a high CKA to the original model on unseen words (Table~\ref{tab:headline}) and on guard tokens (Table~\ref{tab:main}). No benchmark change was detectable at this sample size (standard error about 0.035 for 200 examples; every change was at most 0.030), whereas cloze accuracy rose on probe words in all six runs and on unseen words in five (Table~\ref{tab:beh}).

\subsection{Comparison with a category-aware heuristic}
\label{sec:heuristic}

On the same target and seed, the analyst's purity gain under $[D]$ was higher than the default heuristic's in three of the six pairs and lower in three, so the two arms are comparable in effectiveness. At a comparable purity gain, the analyst’s edits used less of the divergence budget: in all six pairs its cumulative KL was lower and its CKA to the original model higher, and each accepted edit contained one operation, against 2.5 to 2.8 for the heuristic (Tables~\ref{tab:headline}, \ref{tab:ops}, and~\ref{tab:decisions}).

Two tuned heuristic configurations from the ablations (Table~\ref{tab:headline}) matched or exceeded the default analyst on Qwen in purity gain, CKA, KL, and held-out perplexity, while the analyst kept the larger silhouette gain. They were not run on the LLM arm, so the comparisons above refer to the default configuration of both arms (Section~\ref{sec:limitations}). Most of the analyst's accepted edits passed only at the relaxed gain threshold, which admits very small gains and, in two cases of the main study, neutral edits (Table~\ref{tab:decisions}, Appendix~\ref{app:ops}). The hybrid arm and per-run values are in Appendix~\ref{app:extended}.

\subsection{The analyst uses the topological description}
\label{sec:descablation}

To test whether the analyst uses the descriptors, Table~\ref{tab:descsum} restricts the digest to its four cumulative levels (per-seed values: Table~\ref{tab:desc}). The full digest was the only level at which every run on both targets had a purity gain above 0.010, and with $P$ and $S$ alone the gain on SmolLM2 was negative in all three runs. The full digest also exceeded the $P$-and-$S$ digest and the digest with anisotropy in mean purity gain over the six runs (paired $p=0.009$ and $p=0.024$; Table~\ref{tab:pooled}). Level (iii) adds intrinsic dimension and Betti numbers together, so the homological contribution is not isolated.

\begin{table}[!tbp]
\centering
\caption{Descriptor ablation, LLM arm: range over three seeds of the purity change under $[D]$ and of the cumulative KL (nats per token). Each level adds descriptors to the level above it. ID: intrinsic dimension.}
\label{tab:descsum}
\footnotesize
\setlength{\tabcolsep}{2.5pt}
\begin{tabular}{lcccc}
\toprule
 & \multicolumn{2}{c}{Qwen} & \multicolumn{2}{c}{SmolLM2} \\
\cmidrule(lr){2-3}\cmidrule(lr){4-5}
Digest & $\Delta P$ & KL & $\Delta P$ & KL \\
\midrule
(i) $P$, $S$ & .010--.021 & .09--.11 & $-$.003--.000 & .07--.14 \\
(ii) + $\kappa$ & .004--.016 & .08--.09 & .001--.007 & .09--.17 \\
(iii) + ID, Betti & .020--.027 & .08--.10 & .001--.009 & .07--.14 \\
(iv) Full & .020--.029 & .09--.10 & .010--.012 & .08--.10 \\
\bottomrule
\end{tabular}
\end{table}

Beyond the digest levels, a boundary appears at 16 categories. With 16 categories, the analyst's purity gain under $[D]$ was negative in five of six runs, although its silhouette gain stayed positive (Table~\ref{tab:ablcatllm}). We attribute this to three properties: the composite gate $\Delta P+\Delta S$ can pass an edit on silhouette alone, the digest shows six of the 120 category pairs, and uncurated polysemous lemmas increase cluster overlap. With four categories, the analyst's purity gain was positive in all six runs, although the heuristic's was larger in every pair (Table~\ref{tab:ablcatrule}).

\subsection{The sequential loop is required in this setting}
\label{sec:parallel}

Table~\ref{tab:parallelsum} compares each sequential run with the parallel run of the same arm, target, and seed (Table~\ref{tab:parallel}, Appendix~\ref{app:moved}). For the LLM arm, the parallel mode raised the cumulative KL by a factor of 7.7 to 14.6 and held-out perplexity by more than 100\%, against single-digit increases in the sequential mode. For the heuristic, it reached a KL of several nats per token, lowered HellaSwag accuracy by 0.20 to 0.26, and left purity below that of the unedited model in all six runs.

\begin{table*}[!htbp]
\centering
\caption{Sequential versus parallel mode: range over six runs per arm. $\Delta P$: purity change under $[D]$; CKA$_t$: guard-token CKA; KL: cumulative KL (nats per token); $\Delta\mathrm{PPL}_h$: held-out perplexity change (\%); $\Delta\mathrm{HS}$: HellaSwag change.}
\label{tab:parallelsum}
\footnotesize
\setlength{\tabcolsep}{2pt}
\begin{tabular}{llccccc}
\toprule
Arm & Mode & $\Delta P$ & CKA$_t$ & KL & $\Delta\mathrm{PPL}_h$ & $\Delta\mathrm{HS}$ \\
\midrule
LLM & Seq. & .010--.029 & .96--.99 & .08--.10 & 6--9 & $-$.01--.01 \\
LLM & Par. & .004--.029 & .90--.92 & .70--1.23 & 104--185 & $-$.09--.01 \\
\addlinespace
Heur. & Seq. & .009--.029 & .94--.98 & .11--.12 & 5--9 & $-$.04--.01 \\
Heur. & Par. & $-$.107--$-$.037 & .48--.58 & 3.97--5.56 & $>$4{,}000 & $-$.26--$-$.20 \\
\bottomrule
\end{tabular}
\end{table*}

Every parallel edit passed the per-edit limits of the intervention test in isolation, so the divergence arises from the interaction of the edits: upstream edits change every later layer, so descriptors computed on the unedited model do not describe the layers the parallel control edits. In this setting, the sequential loop is therefore required. The analyst's parallel edits, mostly principal-component rescalings $T_{\mathrm{pcs}}$, kept a positive purity gain, whereas the heuristic's centroid-axis scalings $T_{\mathrm{sep}}$ did not (Appendix~\ref{app:ops}).

\subsection{Transfer across vocabulary and sentence template}
\label{sec:transfer}

Table~\ref{tab:transfer} (Appendix~\ref{app:moved}) separates the change of vocabulary ($[A]\rightarrow[B]$) from the change of sentence template ($[A]\rightarrow[C]$); retention is the gain under a condition divided by the in-sample gain. For the LLM arm, new words did not lower the purity gain in five of six runs, whereas a new template lowered it in all three Qwen runs but not on SmolLM2. Under $[D]$, the analyst retained 0.73 to 1.69 of its in-sample gain on Qwen, against 0.45 to 0.83 for the heuristic, so the remaining gap follows the sentence template; fitting on several templates is a next step.

\subsection{Costs}
\label{sec:costs}
\label{sec:identity}
\label{sec:behavior}

The edits moved the output distribution, as expected: the cumulative KL was 0.083 to 0.104 nats per token in the LLM arm, well within the budget of 0.30 (Table~\ref{tab:headline}). We read this value together with held-out perplexity and the benchmark changes below, because KL alone does not indicate whether the model degraded. Guard-set perplexity stayed within its $+5\%$ budget, but held-out perplexity rose (Table~\ref{tab:main}) and exceeded the guard-set increase in 17 of the 18 default runs; edits are selected adaptively against only 24 guard sentences, which a larger guard set should fix. The truth margin of the factual probe fell in all 18 default runs, by at most 8.1\% of the baseline margin, although the share of prompts with the true completion preferred did not change in any run (Table~\ref{tab:beh}).

\subsection{Round robin across five models}
\label{sec:roundrobin}

To test whether the result depends on the analyst and targets of the main study, each of the five models analyzed the four others (default configuration, seed 0; Table~\ref{tab:rrmatrix}, Appendix~\ref{app:rr}). The two pairs shared with the main study reproduced its outputs exactly. In 16 of the 20 pairs the analyst changed its target, and in all 16 purity under $[D]$ rose over the edited layers; it rose at the last layer in 15, and silhouette rose and $\beta_0$ fell in 15. The gain held on the two larger targets, Llama-3.2-3B and Phi-3-mini, and every model was matured by at least one other. No benchmark changed by more than 0.040, factual-preference accuracy never changed, and the cumulative KL stayed below 0.10 nats per token. In the other four pairs the target was left unchanged: on Gemma-2-2B, small edits exceeded the weight budget for three of four analysts, and the two accepted edits of one further pair were neutral.

\begin{table*}[!htbp]
\centering
\caption{Round robin (default configuration, seed 0): purity change under $[D]$ over the edited layers for every analyst (rows) and target (columns), with accepted edits in parentheses. ``0 (0)'': no accepted edit (the two SmolLM2 edits on Qwen were neutral). Means count an unchanged target as zero.}
\label{tab:rrmatrix}
\footnotesize
\setlength{\tabcolsep}{2pt}
\begin{tabular}{lcccccc}
\toprule
 & \multicolumn{5}{c}{Target} & \\
\cmidrule(lr){2-6}
Analyst & Gemma & Llama & Phi & Qwen & SmolLM2 & Mean \\
\midrule
Gemma & -- & .016 (20) & .022 (21) & .029 (13) & .011 (13) & .019 \\
Llama & .010 (6) & -- & .010 (20) & .022 (18) & .014 (13) & .014 \\
Phi & 0 (0) & .005 (27) & -- & .007 (21) & .007 (9) & .005 \\
Qwen & 0 (0) & .033 (1) & .003 (1) & -- & .006 (1) & .010 \\
SmolLM2 & 0 (0) & .002 (10) & .018 (7) & .000 (2) & -- & .005 \\
\bottomrule
\end{tabular}
\end{table*}

The analysts differed. Gemma-2-2B and Llama-3.2-3B changed all four of their targets and had the largest mean gains, whereas the two smallest models, Qwen2.5-1.5B and SmolLM2-1.7B, had few proposals accepted; the largest model, Phi-3-mini, had one of the smallest mean gains, so the gain did not grow with parameter count. Two observations qualify the result: one single-edit pair lowered purity at most downstream layers, and ARC-Easy accuracy fell in 11 of the 16 changed pairs.

\section{Discussion and limitations}
\label{sec:discussion}

\subsection{What the results show}
\label{sec:hypothesis}

With lightweight models, an analyst that reads only a fixed-size topological and geometric digest made the latent geometry of trained targets measurably more organized on vocabulary and sentence frames it never saw, and the change was written into the weights without gradient descent. The round robin suggests that this does not depend on one pair of models. Every model acted as analyst for at least two targets whose purity it raised, and every model was matured by at least one analyst. The gains are modest and come at a cost in held-out perplexity and factual margin.

We read these results as a proof of concept, not as a final editing method, and we do not claim that the edited models are better trained behaviorally in general. They show that, in our setting, a compressed topological description suffices for an LLM analyst to propose geometrically meaningful edits, that ridge regression bakes them into the weights, and that a sequential loop keeps the divergence bounded. The analyst's reading of the topology appears to matter, its gains depend on the descriptors shown, its edits are gentler than those of a heuristic of equal effectiveness, and they interfere less across layers in the parallel control. The open questions are how far the paradigm scales and how it compares with the steering and closed-form editing baselines of Section~\ref{sec:related}, which we did not run.

A possible mechanism concerns the operation types: 26 of the 30 analyst edits that passed the strict gain threshold used $T_{\mathrm{pcs}}$, whereas the heuristic only uses $T_{\mathrm{sep}}$ (Appendix~\ref{app:ops}). We hypothesize that edits along such global directions depend less on the fitted vocabulary and interfere less across layers. Operation type remains confounded with the number of operations per edit and with the adaptive strength, and the analyst almost never used the other two operations (no $T_{\mathrm{ctr}}$ and one rank-one $T_{\mathrm{wh}}$, which leaves a layer unchanged, across 126 runs); restricting the edit language to one operation type at a time will test the hypothesis.

\subsection{Limitations and future work}
\label{sec:limitations}
\label{sec:roadmap}

The main study uses one analyst and two targets; the round robin covers all 20 pairs of five models of up to 3.8 billion parameters, with one seed and without the heuristic arm on the three additional targets. The lightweight LLMs allowed fast experiments on a single local GPU; analysts and targets larger than those tested remain to be evaluated. Our hypothesis is that a more capable analyst with stronger reasoning would propose better edits for baking into the weights, with higher acceptance at the strict gain threshold and better transfer. Testing this hypothesis is a direction for future work. Four further limits apply: the analyst's written rationales are short and often generic (Appendix~\ref{app:prompt}), so our conclusions rest on measured outcomes; the three seeds resample data splits and benchmark samples on the same frozen weights, so the statistics are descriptive; the strength and gain-threshold ablations covered the heuristic arm only; and the WordNet probe vocabulary was not sense-curated, which lowers the attainable purity. Next steps are (a)~running the tuned configurations on the LLM arm and testing single-operation edit languages; (b)~separate non-negativity gates, a guard set of 100 to 200 sentences, a factual guard, fitting on several templates, and sense curation; (c)~a multi-seed round robin with the heuristic on all five targets, larger analysts (7B to 70B+ parameters), and comparison with the steering and closed-form editing methods of Section~\ref{sec:related}; and (d)~starting from earlier checkpoints or degraded models to test whether an analyst can move an immature model toward its mature counterpart, and letting a model edit itself.

\section{Conclusion}
\label{sec:conclusion}

Topology-aware weight baking lets one language model refine the latent geometry of another from compressed topological descriptors alone, and writes each verified edit into the weights in closed form, without gradient descent. With lightweight models, the analyst's edits made categories purer and better separated on unseen words in unseen sentence frames in every default run and left benchmark accuracy within sampling error. The analyst matched a category-aware heuristic with gentler edits, and applying the edits sequentially kept the divergence within its budget while benchmark accuracy stayed within sampling error. The edits cost held-out perplexity and factual margin, some round-robin pairs left the target unchanged, and larger models, steering baselines, and immature starting points remain untested (Section~\ref{sec:limitations}). We hope this work opens a research direction in which trained models are refined by other models through compact summaries of their internal state, without gradient descent or raw weight access.

\section*{Data and Code Availability}

The code and the data generated and analyzed in this study are openly available at \url{https://github.com/amingolnari/weight-baking}. All five language models used in this work are publicly available on Hugging Face (repository identifiers in Table~\ref{tab:models}), and the text corpus and benchmarks used for evaluation are the public datasets cited in Section~\ref{sec:data}.

\section*{Declaration of Competing Interest}

The authors declare that they have no known competing financial interests or personal relationships that could have appeared to influence the work reported in this paper.

\section*{Declaration of using Generative AI}

During the preparation of this work, the authors used generative AI (Claude Opus 5.5) (i) for language editing, to improve the English language, grammar, and flow of the text; and (ii) for assisting the analysis of the large volume of run outputs. All content was verified and edited by the authors, who take full responsibility for the content of the publication.

\section*{Authorship Contribution Statement}

\textbf{Amin Golnari:} Conceptualization, Methodology, Software, Investigation, Writing -- Original Draft, Writing -- Review \& Editing. \textbf{Saeid Sanei:} Supervision, Writing -- Review \& Editing.

\bibliographystyle{unsrt}  
\bibliography{references}  

\clearpage
\appendix

\noindent\textit{This appendix is supplementary: it documents definitions, the analyst prompt, per-run values, and extended tables, and it is not required for the core claims of the paper. Every table states the study and the runs from which it is computed and lists every run separately; the statistical tests of Appendix~\ref{app:stats} are computed separately for each arm. All tables concern the 126 runs of the main study (Gemma-2-2B as analyst; Qwen2.5-1.5B and SmolLM2-1.7B as targets), except those of Appendix~\ref{app:rr}, which concern the 20 runs of the round-robin study.}
\section{Supplementary Figures and Tables}
\label{app:moved}

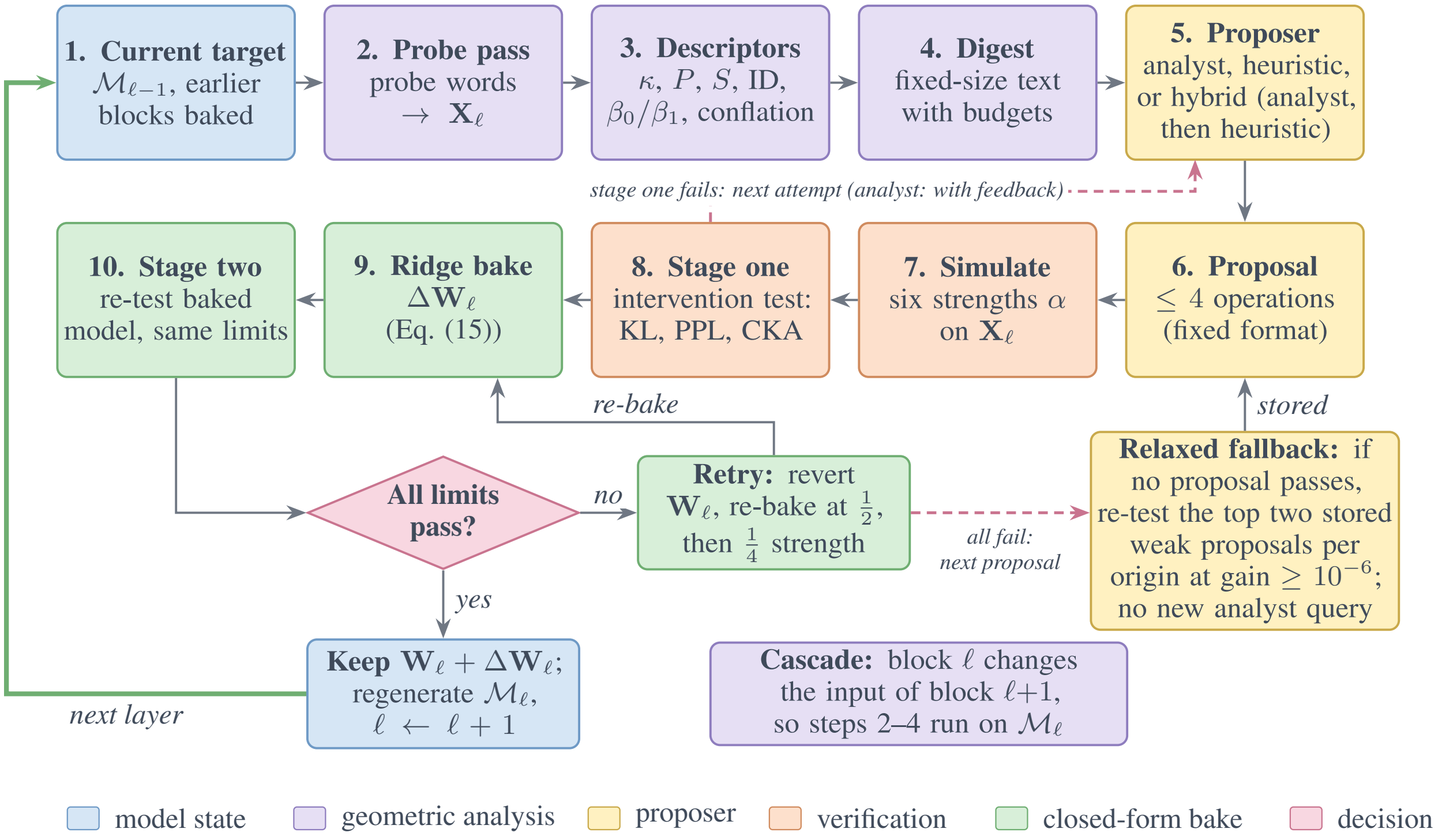
\begin{figure}[!tbp]
\centering
\resizebox{\textwidth}{!}{%
\begin{tikzpicture}[
  font=\footnotesize, >={Stealth[length=2.2mm]}, text=pInk,
  box/.style 2 args={rounded corners=3pt, draw=#2, fill=#1, line width=0.7pt,
                     minimum width=2.62cm, minimum height=1.7cm, text width=2.48cm, align=center, inner sep=2pt,
                     execute at begin node={\hyphenpenalty=10000\exhyphenpenalty=10000}},
  flow/.style={->, line width=0.7pt, draw=pInk!75},
  fb/.style={->, line width=0.9pt, dash pattern=on 3pt off 2pt, draw=pPinkE},
  regen/.style={->, line width=1.5pt, draw=pGreenE},
  lab/.style={font=\footnotesize\itshape, text=pInk, fill=white, inner sep=1.5pt}]
\node[box={pBlue}{pBlueE}]   (n1) at (0,0)     {\textbf{1. Current target}\\ $\mathcal{M}_{\ell-1}$, earlier\\ blocks baked};
\node[box={pLav}{pLavE}]     (n2) at (2.95,0)  {\textbf{2. Probe pass}\\ probe words\\ $\rightarrow \mathbf{X}_\ell$};
\node[box={pLav}{pLavE}]     (n3) at (5.9,0)   {\textbf{3. Descriptors}\\ $\kappa$, $P$, $S$, ID,\\ $\beta_0/\beta_1$, conflation};
\node[box={pLav}{pLavE}]     (n4) at (8.85,0)  {\textbf{4. Digest}\\ fixed-size text\\ with budgets};
\node[box={pYel}{pYelE}]     (n5) at (11.8,0)  {\textbf{5. Proposer}\\ analyst, heuristic,\\ or hybrid (analyst,\\ then heuristic)};
\node[box={pYel}{pYelE}]     (n6) at (11.8,-2.4) {\textbf{6. Proposal}\\ $\le 4$ operations\\ (fixed format)};
\node[box={pPeach}{pPeachE}] (n7) at (8.85,-2.4) {\textbf{7. Simulate}\\ six strengths $\alpha$\\ on $\mathbf{X}_\ell$};
\node[box={pPeach}{pPeachE}] (n8) at (5.9,-2.4)  {\textbf{8. Stage one}\\ intervention test:\\ KL, PPL, CKA};
\node[box={pGreen}{pGreenE}] (n9) at (2.95,-2.4) {\textbf{9. Ridge bake}\\ $\Delta\mathbf{W}_\ell$\\ (Eq.~\eqref{eq:dual})};
\node[box={pGreen}{pGreenE}] (n10) at (0,-2.4)   {\textbf{10. Stage two}\\ re-test baked\\ model, same limits};
\node[draw=pPinkE, fill=pPink, line width=0.7pt, diamond, aspect=2.4, inner sep=1pt,
      align=center, font=\footnotesize\bfseries] (d) at (2.95,-4.75) {All limits\\ pass?};
\node[box={pGreen}{pGreenE}, minimum width=3.0cm, text width=2.85cm, minimum height=1.25cm] (retry) at (6.6,-4.75)
      {\textbf{Retry:} revert $\mathbf{W}_\ell$, re-bake at $\tfrac12$, then $\tfrac14$ strength};
\node[box={pYel}{pYelE}, minimum width=3.4cm, text width=3.25cm, minimum height=1.55cm] (relax) at (11.8,-4.95)
      {\textbf{Relaxed fallback:} if no proposal passes, re-test the top two stored weak proposals per origin at gain $\ge 10^{-6}$; no new analyst query};
\node[box={pBlue}{pBlueE}, minimum width=3.0cm, text width=2.8cm, minimum height=1.2cm] (keep) at (2.95,-6.75)
      {\textbf{Keep} $\mathbf{W}_\ell+\Delta\mathbf{W}_\ell$;\\ regenerate $\mathcal{M}_\ell$, $\ell \leftarrow \ell+1$};
\node[box={pLav}{pLavE}, minimum width=4.6cm, text width=4.4cm, minimum height=1.0cm] at (8.2,-6.75)
      {\textbf{Cascade:} block $\ell$ changes the input of block $\ell{+}1$, so steps 2--4 run on $\mathcal{M}_\ell$};
\draw[flow] (n1) -- (n2); \draw[flow] (n2) -- (n3); \draw[flow] (n3) -- (n4); \draw[flow] (n4) -- (n5);
\draw[flow] (n5) -- (n6); \draw[flow] (n6) -- (n7); \draw[flow] (n7) -- (n8); \draw[flow] (n8) -- (n9); \draw[flow] (n9) -- (n10);
\draw[flow] (n10.south) |- (d.west);
\draw[flow] (d.south) -- node[lab, right=2pt] {yes} (keep.north);
\draw[flow] (d.east) -- node[lab, above=1pt] {no} (retry.west);
\draw[flow] (retry.north) |- node[lab, pos=0.75, above=1pt] {re-bake} ([xshift=0.6cm]n9.south |- 0,-3.75) -- ([xshift=0.6cm]n9.south);
\draw[fb] (n8.north) |- node[lab, pos=0.62, font=\scriptsize\itshape] {stage one fails: next attempt (analyst: with feedback)} (11.25,-1.2) -- (11.25,-0.85);
\draw[flow] (relax.north) -- node[lab, right=2pt] {stored} (n6.south);
\draw[fb] (retry.east) -- node[lab, below=3pt, align=center, font=\scriptsize\itshape] {all fail:\\ next proposal} (retry.east -| relax.west);
\draw[regen] (keep.west) -| node[lab, pos=0.3, below=1pt] {next layer} ([xshift=-0.55cm]n1.west) -- (n1.west);
\begin{scope}[shift={(-1.2,-8.25)}, every node/.style={font=\footnotesize, text=pInk, anchor=west}]
\foreach \x/\c/\e/\t in {0.0/pBlue/pBlueE/model state, 2.5/pLav/pLavE/geometric analysis, 5.75/pYel/pYelE/proposer, 7.75/pPeach/pPeachE/verification, 10.25/pGreen/pGreenE/closed-form bake, 13.5/pPink/pPinkE/decision}{
  \fill[\c, draw=\e, rounded corners=1pt] (\x,0) rectangle ++(0.35,0.25);
  \node at (\x+0.4,0.125) {\t};
}
\end{scope}
\end{tikzpicture}}
\caption{The sequential regeneration cycle of topology-aware weight baking. For block $\ell$, the current target is probed, and the descriptors of its feature space are summarized as a fixed-size digest (steps 1--5). The proposer supplies a list of operations, which the system simulates at six strengths and tests through a temporary intervention on the layer output (stage one, steps 6--8). In the LLM arm, a proposal that fails stage one is sent back to the analyst with the simulated numbers of the proposal as written ($\alpha=1$) and, if an intervention test ran, the measurements of the last strength tested; at most three analyst proposals are made per layer. In the heuristic arm, the next attempt uses fewer category pairs, and in the hybrid arm, the heuristic follows the analyst. If no proposal passes, the two best stored weak proposals of each origin are re-tested at a relaxed gain threshold without a new analyst query. A proposal that passes stage one is baked by closed-form ridge regression and re-tested on the baked model (stage two, steps 9--10); if a limit fails, the weights are reverted and the edit is baked again at half and at a quarter of its strength. If all limits pass (solid green arrow), the baked weights are kept and the cycle restarts at block $\ell+1$ on the regenerated model. ID: intrinsic dimension; $\beta_0/\beta_1$: Betti numbers.}
\label{fig:pipeline}
\end{figure}

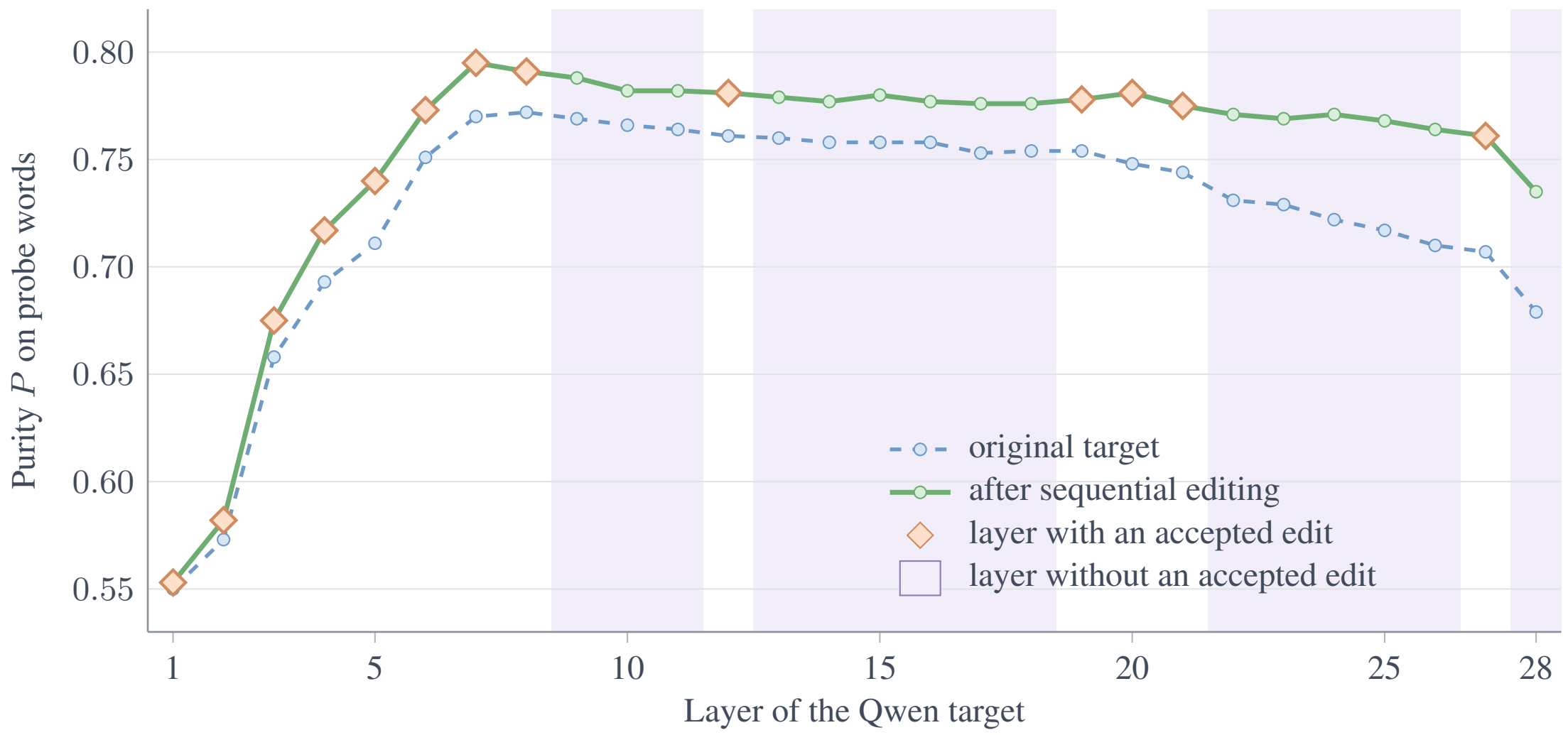
\begin{figure}[!tbp]
\centering
\begin{tikzpicture}[x=0.47cm, y=20cm, text=pInk, font=\footnotesize]
\begin{scope}[yshift=-10.6cm] 
  \foreach \a/\b in {8.5/11.5, 12.5/18.5, 21.5/26.5, 27.5/28.5} \fill[pLav!55] (\a,0.53) rectangle (\b,0.82);
  \foreach \y in {0.55,0.60,...,0.801} { \draw[pInk!15] (0.5,\y) -- (28.5,\y); \node[left] at (0.5,\y) {\pgfmathprintnumber[fixed,precision=2,zerofill]{\y}}; }
  \foreach \x in {1,5,10,15,20,25,28} { \draw[pInk!40] (\x,0.53) -- (\x,0.525); \node[below] at (\x,0.525) {\x}; }
  \draw[pInk!60, line width=0.5pt] (0.5,0.53) -- (28.5,0.53);
  \draw[pInk!60, line width=0.5pt] (0.5,0.53) -- (0.5,0.82);
  \node at (14.5,0.492) {Layer of the Qwen target};
  \node[rotate=90] at (-1.9,0.675) {Purity $P$ on probe words};
  \draw[pBlueE, line width=1.0pt, dashed] plot coordinates {(1,0.550) (2,0.573) (3,0.658) (4,0.693) (5,0.711) (6,0.751) (7,0.770) (8,0.772) (9,0.769) (10,0.766) (11,0.764) (12,0.761) (13,0.760) (14,0.758) (15,0.758) (16,0.758) (17,0.753) (18,0.754) (19,0.754) (20,0.748) (21,0.744) (22,0.731) (23,0.729) (24,0.722) (25,0.717) (26,0.710) (27,0.707) (28,0.679)};
  \foreach \p in {(1,0.550),(2,0.573),(3,0.658),(4,0.693),(5,0.711),(6,0.751),(7,0.770),(8,0.772),(9,0.769),(10,0.766),(11,0.764),(12,0.761),(13,0.760),(14,0.758),(15,0.758),(16,0.758),(17,0.753),(18,0.754),(19,0.754),(20,0.748),(21,0.744),(22,0.731),(23,0.729),(24,0.722),(25,0.717),(26,0.710),(27,0.707),(28,0.679)} \fill[pBlue, draw=pBlueE] \p circle (1.6pt);
  \draw[pGreenE, line width=1.3pt] plot coordinates {(1,0.553) (2,0.582) (3,0.675) (4,0.717) (5,0.740) (6,0.773) (7,0.795) (8,0.791) (9,0.788) (10,0.782) (11,0.782) (12,0.781) (13,0.779) (14,0.777) (15,0.780) (16,0.777) (17,0.776) (18,0.776) (19,0.778) (20,0.781) (21,0.775) (22,0.771) (23,0.769) (24,0.771) (25,0.768) (26,0.764) (27,0.761) (28,0.735)};
  \foreach \p in {(1,0.553),(2,0.582),(3,0.675),(4,0.717),(5,0.740),(6,0.773),(7,0.795),(8,0.791),(9,0.788),(10,0.782),(11,0.782),(12,0.781),(13,0.779),(14,0.777),(15,0.780),(16,0.777),(17,0.776),(18,0.776),(19,0.778),(20,0.781),(21,0.775),(22,0.771),(23,0.769),(24,0.771),(25,0.768),(26,0.764),(27,0.761),(28,0.735)} \fill[pGreen, draw=pGreenE] \p circle (1.6pt);
  \foreach \p in {(1,0.553),(2,0.582),(3,0.675),(4,0.717),(5,0.740),(6,0.773),(7,0.795),(8,0.791),(12,0.781),(19,0.778),(20,0.781),(21,0.775),(27,0.761)} \node[diamond, minimum size=7pt, inner sep=0pt, fill=pPeach, draw=pPeachE, line width=0.8pt] at \p {};
  \begin{scope}[shift={(15.2,0.615)}]
    \draw[pBlueE, dashed, line width=1.0pt] (0,0) -- (1.2,0); \fill[pBlue, draw=pBlueE] (0.6,0) circle (1.6pt); \node[anchor=west] at (1.3,0) {original target};
    \draw[pGreenE, line width=1.3pt] (0,-0.02) -- (1.2,-0.02); \fill[pGreen, draw=pGreenE] (0.6,-0.02) circle (1.6pt); \node[anchor=west] at (1.3,-0.02) {after sequential editing};
    \node[diamond, minimum size=7pt, inner sep=0pt, fill=pPeach, draw=pPeachE] at (0.6,-0.04) {}; \node[anchor=west] at (1.3,-0.04) {layer with an accepted edit};
    \fill[pLav!55, draw=pLavE, line width=0.4pt] (0.2,-0.068) rectangle (1.0,-0.052); \node[anchor=west] at (1.3,-0.06) {layer without an accepted edit};
  \end{scope}
\end{scope}
\end{tikzpicture}
\caption{Propagation of upstream edits, shown for the Qwen target edited in the LLM arm (seed 0). Purity on probe words for each layer of the original target (dashed) and of the final edited target (solid). Diamonds mark the 13 layers with an accepted edit (1--8, 12, 19--21, and 27). The 15 layers without an accepted edit (shaded) end 0.016 to 0.056 above the original value, because the edits of earlier blocks change the input of every later block.}
\label{fig:cascade}
\end{figure}

This section holds the figures and tables that the main text references: the regeneration cycle (Figure~\ref{fig:pipeline}) and its pseudocode (Algorithm~\ref{alg:loop}), the models (Table~\ref{tab:models}), every default run (Tables~\ref{tab:main}, \ref{tab:transfer}, and~\ref{tab:beh}), the parallel control (Table~\ref{tab:parallel}), the propagation of upstream edits (Figure~\ref{fig:cascade}), and the descriptor ablation (Table~\ref{tab:desc}).

\begin{algorithm}[!tbp]
\caption{Sequential Topology-Aware Weight Baking}
\label{alg:loop}
\begin{algorithmic}[1]
\small
\REQUIRE Frozen target $\mathcal{M}$ with $L$ blocks; probe set; arm (LLM, heuristic, or hybrid); limits (Table~\ref{tab:limits})
\FOR{$\ell=1$ \textbf{to} $L$}
  \STATE Compute $\mathbf{X}_\ell$ and the descriptors on the current $\mathcal{M}$
  \IF{a budget is spent \textbf{or} the quality screen skips $\ell$}
    \STATE \textbf{continue}
  \ENDIF
  \STATE $\mathcal{W}\leftarrow$ empty store of weak proposals per origin
  \STATE Proposal sequence: LLM arm: up to three analyst attempts; heuristic arm: $q=3,2,1$ pairs; hybrid arm: analyst attempts, then $q=3,2,1$
  \FOR{each proposal $p$ in the sequence}
    \STATE Simulate $p$ at all $\alpha$; keep strengths within size and CKA limits
    \IF{no kept strength reaches gain $\ge 0.01$}
      \STATE Store the best positive-gain strength of $p$ in $\mathcal{W}$
    \ELSE
      \STATE Stage one: intervention-test kept strengths by descending gain
      \IF{a strength passes}
        \STATE accept $p$ at that strength; \textbf{break}
      \ENDIF
    \ENDIF
    \STATE Analyst proposals: give feedback (Appendix~\ref{app:prompt})
  \ENDFOR
  \IF{nothing accepted \textbf{and} the relaxed stage is on}
    \STATE For each origin, re-test the top two proposals in $\mathcal{W}$ at gain $\ge 10^{-6}$ (no new analyst query); accept the first that passes stage one
  \ENDIF
  \IF{a proposal was accepted}
    \FOR{$a\in\{1,\tfrac12,\tfrac14\}$}
      \STATE Bake $\Delta\mathbf{W}_\ell$ at strength $a$ (Eq.~\eqref{eq:dual})
      \STATE Stage two: re-test the baked model
      \IF{all limits pass}
        \STATE Keep $\mathbf{W}_\ell+\Delta\mathbf{W}_\ell$; \textbf{break}
      \ELSE
        \STATE Revert $\mathbf{W}_\ell$
      \ENDIF
    \ENDFOR
  \ENDIF
\ENDFOR
\ENSURE Edited target $\mathcal{M}'$
\end{algorithmic}
\end{algorithm}

\begin{table}[!tbp]
\centering
\caption{Models of the round-robin pool. Main study: role in the 126 runs of the main study; in the round-robin study, every model is the analyst of the four others and a target of each of them. $L$: number of transformer blocks; $d$: residual width; $d_k$: input width of the feed-forward output projection. Parameter counts include embeddings. A model is loaded in bfloat16 as analyst and in 32-bit floating point as target.}
\label{tab:models}
\small
\begin{tabular}{lllrrrr}
\toprule
Model & Main study & Hugging Face repository & $L$ & $d$ & $d_k$ & Params (M) \\
\midrule
Gemma-2-2B   & Analyst & unsloth/gemma-2-2b-it               & 26 & 2304 & 9216 & 2,614 \\
Qwen2.5-1.5B & Target  & Qwen/Qwen2.5-1.5B-Instruct          & 28 & 1536 & 8960 & 1,544 \\
SmolLM2-1.7B & Target  & HuggingFaceTB/SmolLM2-1.7B-Instruct & 24 & 2048 & 8192 & 1,711 \\
\midrule
Llama-3.2-3B & -- & unsloth/Llama-3.2-3B-Instruct  & 28 & 3072 & 8192 & 3,213 \\
Phi-3-mini   & -- & microsoft/Phi-3-mini-4k-instruct & 32 & 3072 & 8192 & 3,821 \\
\bottomrule
\end{tabular}
\end{table}

\begin{table}[!tbp]
\centering
\caption{Geometry, identity preservation, and divergence of every default run (sequential loop; 18 runs). Changes are edited minus original and are averaged over the edited layers, except $\Delta P_L$, which is the change at the last layer. $[A]$: probe words in the probe template; $[D]$: unseen words in unseen templates. CKA$_p$, CKA$_w$, CKA$_t$: CKA to the original model on probe words, unseen words, and guard-set tokens. KL: cumulative KL (nats per token). $\Delta\mathrm{PPL}_g$, $\Delta\mathrm{PPL}_h$: perplexity change on the guard set and on the held-out set (\%). $\uparrow$: higher is better; $\downarrow$: lower is better.}
\label{tab:main}
\begin{adjustbox}{max width=\textwidth}
\small
\setlength{\tabcolsep}{3.5pt}
\begin{tabular}{lllrrrrrrrrrrrr}
\toprule
Target & Arm & Seed & Edits & $\Delta P^{[A]}$ $\uparrow$ & $\Delta P^{[D]}$ $\uparrow$ & $\Delta P_L^{[D]}$ $\uparrow$ & $\Delta S^{[D]}$ $\uparrow$ & $\Delta\kappa^{[D]}$ $\downarrow$ & CKA$_p$ $\uparrow$ & CKA$_w$ $\uparrow$ & CKA$_t$ $\uparrow$ & KL $\downarrow$ & $\Delta\mathrm{PPL}_g$ $\downarrow$ & $\Delta\mathrm{PPL}_h$ $\downarrow$ \\
\midrule
\multirow{10}{*}{Qwen} & Unedited & -- & 0 & 0 & 0 & 0 & 0 & 0 & 1 & 1 & 1 & 0 & 0 & 0 \\
\cmidrule(l){2-15}
 & \multirow{3}{*}{LLM arm} & 0 & 13 & 0.0239 & 0.0287 & 0.061 & 0.0235 & $-$0.0300 & 0.873 & 0.897 & 0.975 & 0.089 & 5.00 & 6.45 \\
 &  & 1 & 20 & 0.0269 & 0.0196 & 0.060 & 0.0484 & $-$0.0435 & 0.923 & 0.935 & 0.981 & 0.086 & 4.28 & 6.96 \\
 &  & 2 & 23 & 0.0156 & 0.0264 & 0.060 & 0.0453 & $-$0.0684 & 0.909 & 0.919 & 0.962 & 0.104 & 4.99 & 8.72 \\
\cmidrule(l){2-15}
 & \multirow{3}{*}{Heuristic} & 0 & 23 & 0.0344 & 0.0285 & 0.065 & 0.0158 & $-$0.0213 & 0.800 & 0.837 & 0.982 & 0.115 & 4.09 & 8.14 \\
 &  & 1 & 21 & 0.0315 & 0.0235 & 0.059 & 0.0153 & $-$0.0229 & 0.818 & 0.849 & 0.982 & 0.110 & 4.70 & 8.05 \\
 &  & 2 & 20 & 0.0360 & 0.0162 & 0.055 & 0.0110 & $-$0.0260 & 0.776 & 0.812 & 0.981 & 0.108 & 3.27 & 8.26 \\
\cmidrule(l){2-15}
 & \multirow{3}{*}{Hybrid} & 0 & 23 & 0.0316 & 0.0232 & 0.066 & 0.0115 & $-$0.0177 & 0.794 & 0.835 & 0.982 & 0.114 & 2.62 & 7.28 \\
 &  & 1 & 24 & 0.0389 & 0.0263 & 0.070 & 0.0269 & $-$0.0370 & 0.834 & 0.861 & 0.979 & 0.111 & 4.97 & 8.20 \\
 &  & 2 & 26 & 0.0340 & 0.0169 & 0.062 & 0.0125 & $-$0.0275 & 0.831 & 0.857 & 0.982 & 0.102 & 4.84 & 8.81 \\
\midrule
\multirow{10}{*}{SmolLM2} & Unedited & -- & 0 & 0 & 0 & 0 & 0 & 0 & 1 & 1 & 1 & 0 & 0 & 0 \\
\cmidrule(l){2-15}
 & \multirow{3}{*}{LLM arm} & 0 & 13 & 0.0017 & 0.0111 & 0.026 & 0.0266 & $-$0.0210 & 0.902 & 0.902 & 0.984 & 0.092 & 4.81 & 6.82 \\
 &  & 1 & 15 & 0.0067 & 0.0104 & 0.050 & 0.0166 & $-$0.0127 & 0.959 & 0.962 & 0.988 & 0.083 & 4.99 & 6.50 \\
 &  & 2 & 16 & 0.0095 & 0.0117 & 0.052 & 0.0163 & $-$0.0135 & 0.950 & 0.953 & 0.989 & 0.099 & 4.90 & 8.17 \\
\cmidrule(l){2-15}
 & \multirow{3}{*}{Heuristic} & 0 & 15 & 0.0097 & 0.0150 & 0.057 & 0.0173 & $-$0.0240 & 0.881 & 0.898 & 0.977 & 0.121 & 4.96 & 4.71 \\
 &  & 1 & 19 & 0.0131 & 0.0139 & 0.061 & 0.0203 & $-$0.0217 & 0.867 & 0.885 & 0.942 & 0.112 & 4.88 & 6.04 \\
 &  & 2 & 17 & 0.0064 & 0.0087 & 0.043 & 0.0219 & $-$0.0270 & 0.833 & 0.853 & 0.964 & 0.119 & 4.95 & 9.22 \\
\cmidrule(l){2-15}
 & \multirow{3}{*}{Hybrid} & 0 & 19 & 0.0100 & 0.0147 & 0.057 & 0.0191 & $-$0.0190 & 0.870 & 0.892 & 0.978 & 0.115 & 4.91 & 5.47 \\
 &  & 1 & 18 & 0.0128 & 0.0143 & 0.058 & 0.0208 & $-$0.0252 & 0.906 & 0.919 & 0.940 & 0.112 & 4.83 & 5.63 \\
 &  & 2 & 20 & 0.0044 & 0.0070 & 0.042 & 0.0175 & $-$0.0203 & 0.855 & 0.874 & 0.972 & 0.102 & 3.90 & 9.25 \\
\bottomrule
\end{tabular}
\end{adjustbox}
\end{table}

\begin{table}[!tbp]
\centering
\caption{Sequential versus parallel reconstruction for every seed (12 sequential and 12 parallel default runs of the LLM and heuristic arms). In the parallel mode, every edit is planned, verified, and fitted on the unedited model, and all edits are applied together. Edits: accepted layer edits; $\Delta P$: purity change under $[D]$; CKA$_t$: CKA on guard-set tokens; KL: cumulative KL (nats per token); $\Delta\mathrm{PPL}$: held-out perplexity change (\%); $\Delta\mathrm{HS}$: HellaSwag accuracy change.}
\label{tab:parallel}
\begin{adjustbox}{max width=\textwidth}
\small
\setlength{\tabcolsep}{3.5pt}
\begin{tabular}{lllrrrrrrrrrrrr}
\toprule
 & & & \multicolumn{6}{c}{Sequential} & \multicolumn{6}{c}{Parallel} \\
\cmidrule(lr){4-9}\cmidrule(lr){10-15}
Target & Arm & Seed & Edits & $\Delta P$ $\uparrow$ & CKA$_t$ $\uparrow$ & KL $\downarrow$ & $\Delta\mathrm{PPL}$ $\downarrow$ & $\Delta\mathrm{HS}$ $\uparrow$ & Edits & $\Delta P$ $\uparrow$ & CKA$_t$ $\uparrow$ & KL $\downarrow$ & $\Delta\mathrm{PPL}$ $\downarrow$ & $\Delta\mathrm{HS}$ $\uparrow$ \\
\midrule
\multirow{6}{*}{Qwen} & \multirow{3}{*}{LLM arm} & 0 & 13 & 0.029 & 0.975 & 0.089 & 6.5 & 0.010 & 28 & 0.029 & 0.921 & 0.702 & 111.6 & $-$0.020 \\
 &  & 1 & 20 & 0.020 & 0.981 & 0.086 & 7.0 & 0.005 & 28 & 0.015 & 0.923 & 0.696 & 103.6 & 0.005 \\
 &  & 2 & 23 & 0.026 & 0.962 & 0.104 & 8.7 & 0.010 & 27 & 0.027 & 0.909 & 0.801 & 134.4 & 0.005 \\
\cmidrule(l){2-15}
 & \multirow{3}{*}{Heuristic} & 0 & 23 & 0.028 & 0.982 & 0.115 & 8.1 & 0.010 & 28 & $-$0.084 & 0.492 & 5.559 & 19{,}045 & $-$0.230 \\
 &  & 1 & 21 & 0.024 & 0.982 & 0.110 & 8.1 & $-$0.005 & 28 & $-$0.106 & 0.491 & 4.965 & 12{,}334 & $-$0.235 \\
 &  & 2 & 20 & 0.016 & 0.981 & 0.108 & 8.3 & $-$0.005 & 28 & $-$0.107 & 0.483 & 5.241 & 15{,}423 & $-$0.245 \\
\midrule
\multirow{6}{*}{SmolLM2} & \multirow{3}{*}{LLM arm} & 0 & 13 & 0.011 & 0.984 & 0.092 & 6.8 & 0.005 & 19 & 0.011 & 0.908 & 1.136 & 165.8 & $-$0.090 \\
 &  & 1 & 15 & 0.010 & 0.988 & 0.083 & 6.5 & $-$0.010 & 20 & 0.004 & 0.902 & 1.205 & 179.3 & $-$0.075 \\
 &  & 2 & 16 & 0.012 & 0.989 & 0.099 & 8.2 & 0.000 & 20 & 0.010 & 0.900 & 1.234 & 185.1 & $-$0.065 \\
\cmidrule(l){2-15}
 & \multirow{3}{*}{Heuristic} & 0 & 15 & 0.015 & 0.977 & 0.121 & 4.7 & $-$0.040 & 20 & $-$0.037 & 0.577 & 3.974 & 4{,}367 & $-$0.210 \\
 &  & 1 & 19 & 0.014 & 0.942 & 0.112 & 6.0 & $-$0.020 & 20 & $-$0.040 & 0.502 & 4.091 & 4{,}824 & $-$0.260 \\
 &  & 2 & 17 & 0.009 & 0.964 & 0.119 & 9.2 & 0.000 & 20 & $-$0.059 & 0.553 & 4.013 & 4{,}905 & $-$0.200 \\
\bottomrule
\end{tabular}
\end{adjustbox}
\end{table}

\begin{table}[!tbp]
\centering
\caption{Transfer of the geometric gain under the four conditions for every default run (18 sequential runs). $[A]$: probe words, probe template; $[B]$: unseen words, probe template; $[C]$: probe words, unseen templates; $[D]$: unseen words, unseen templates. $\tau_P^{[X]}=\Delta P^{[X]}/\Delta P^{[A]}$ in the same run (Section~\ref{sec:metrics}); ratios are computed from unrounded values, so they can differ slightly from ratios of the rounded entries. $^{\dagger}$In-sample change below 0.005; the ratio is unstable.}
\label{tab:transfer}
\begin{adjustbox}{max width=\textwidth}
\small
\setlength{\tabcolsep}{3.2pt}
\begin{tabular}{lllccccccccccc}
\toprule
 & & & \multicolumn{4}{c}{$\Delta P$} & \multicolumn{3}{c}{Retention $\tau_P$} & \multicolumn{4}{c}{$\Delta S$} \\
\cmidrule(lr){4-7}\cmidrule(lr){8-10}\cmidrule(lr){11-14}
Target & Arm & Seed & $[A]$ & $[B]$ & $[C]$ & $[D]$ & $[B]$ & $[C]$ & $[D]$ & $[A]$ & $[B]$ & $[C]$ & $[D]$ \\
\midrule
\multirow{9}{*}{Qwen} & \multirow{3}{*}{LLM arm} & 0 & 0.0239 & 0.0349 & 0.0206 & 0.0287 & 1.46 & 0.86 & 1.20 & 0.0258 & 0.0221 & 0.0277 & 0.0235 \\
 &  & 1 & 0.0269 & 0.0224 & 0.0216 & 0.0196 & 0.83 & 0.80 & 0.73 & 0.0422 & 0.0487 & 0.0426 & 0.0484 \\
 &  & 2 & 0.0156 & 0.0333 & 0.0117 & 0.0264 & 2.14 & 0.75 & 1.69 & 0.0445 & 0.0451 & 0.0441 & 0.0453 \\
\cmidrule(l){2-14}
 & \multirow{3}{*}{Heuristic} & 0 & 0.0344 & 0.0373 & 0.0262 & 0.0285 & 1.08 & 0.76 & 0.83 & 0.0209 & 0.0120 & 0.0250 & 0.0158 \\
 &  & 1 & 0.0315 & 0.0273 & 0.0230 & 0.0235 & 0.87 & 0.73 & 0.75 & 0.0138 & 0.0137 & 0.0167 & 0.0153 \\
 &  & 2 & 0.0360 & 0.0259 & 0.0290 & 0.0162 & 0.72 & 0.81 & 0.45 & 0.0146 & 0.0079 & 0.0190 & 0.0110 \\
\cmidrule(l){2-14}
 & \multirow{3}{*}{Hybrid} & 0 & 0.0316 & 0.0313 & 0.0244 & 0.0232 & 0.99 & 0.77 & 0.74 & 0.0143 & 0.0077 & 0.0185 & 0.0115 \\
 &  & 1 & 0.0389 & 0.0320 & 0.0286 & 0.0263 & 0.82 & 0.73 & 0.68 & 0.0286 & 0.0257 & 0.0306 & 0.0269 \\
 &  & 2 & 0.0340 & 0.0269 & 0.0280 & 0.0169 & 0.79 & 0.82 & 0.50 & 0.0170 & 0.0099 & 0.0207 & 0.0125 \\
\midrule
\multirow{9}{*}{SmolLM2} & \multirow{3}{*}{LLM arm} & 0 & 0.0017 & 0.0145 & 0.0025 & 0.0111 & 8.62$^{\dagger}$ & 1.49$^{\dagger}$ & 6.58$^{\dagger}$ & 0.0285 & 0.0281 & 0.0281 & 0.0266 \\
 &  & 1 & 0.0067 & 0.0137 & 0.0071 & 0.0104 & 2.04 & 1.06 & 1.55 & 0.0151 & 0.0175 & 0.0152 & 0.0166 \\
 &  & 2 & 0.0095 & 0.0097 & 0.0096 & 0.0117 & 1.02 & 1.01 & 1.23 & 0.0169 & 0.0138 & 0.0179 & 0.0163 \\
\cmidrule(l){2-14}
 & \multirow{3}{*}{Heuristic} & 0 & 0.0097 & 0.0147 & 0.0108 & 0.0150 & 1.52 & 1.12 & 1.56 & 0.0243 & 0.0146 & 0.0260 & 0.0173 \\
 &  & 1 & 0.0131 & 0.0179 & 0.0144 & 0.0139 & 1.37 & 1.10 & 1.06 & 0.0221 & 0.0202 & 0.0238 & 0.0203 \\
 &  & 2 & 0.0064 & 0.0096 & 0.0103 & 0.0087 & 1.51 & 1.62 & 1.36 & 0.0229 & 0.0161 & 0.0250 & 0.0219 \\
\cmidrule(l){2-14}
 & \multirow{3}{*}{Hybrid} & 0 & 0.0100 & 0.0183 & 0.0119 & 0.0147 & 1.83 & 1.20 & 1.47 & 0.0254 & 0.0164 & 0.0283 & 0.0191 \\
 &  & 1 & 0.0128 & 0.0173 & 0.0135 & 0.0143 & 1.35 & 1.05 & 1.11 & 0.0222 & 0.0212 & 0.0227 & 0.0208 \\
 &  & 2 & 0.0044 & 0.0045 & 0.0078 & 0.0070 & 1.04$^{\dagger}$ & 1.80$^{\dagger}$ & 1.60$^{\dagger}$ & 0.0180 & 0.0122 & 0.0198 & 0.0175 \\
\bottomrule
\end{tabular}
\end{adjustbox}
\end{table}

\begin{table}[!tbp]
\centering
\caption{Behavioral changes of every default run (sequential loop; 18 runs). Cloze: category cloze accuracy on probe words and on unseen words. HS, ARC, TQA: accuracy on HellaSwag, ARC-Easy, and TruthfulQA (200 examples each, redrawn per seed). Fact acc.: share of the 40 factual prompts in which the true completion is preferred. TM: truth margin (nats).}
\label{tab:beh}
\begin{adjustbox}{max width=\textwidth}
\small
\begin{tabular}{lllrrrrrrr}
\toprule
Target & Arm & Seed & $\Delta$cloze, probe & $\Delta$cloze, unseen & $\Delta\mathrm{HS}$ & $\Delta\mathrm{ARC}$ & $\Delta\mathrm{TQA}$ & $\Delta$fact acc. & $\Delta\mathrm{TM}$ \\
\midrule
\multirow{9}{*}{Qwen} & \multirow{3}{*}{LLM arm} & 0 & 0.068 & 0.050 & 0.010 & $-$0.030 & 0.000 & 0.000 & $-$0.256 \\
 &  & 1 & 0.065 & $-$0.003 & 0.005 & 0.015 & $-$0.010 & 0.000 & $-$0.412 \\
 &  & 2 & 0.084 & 0.026 & 0.010 & $-$0.005 & $-$0.005 & 0.000 & $-$0.563 \\
\cmidrule(l){2-10}
 & \multirow{3}{*}{Heuristic} & 0 & 0.058 & 0.023 & 0.010 & $-$0.010 & $-$0.005 & 0.000 & $-$0.198 \\
 &  & 1 & 0.042 & 0.029 & $-$0.005 & 0.005 & $-$0.010 & 0.000 & $-$0.237 \\
 &  & 2 & 0.043 & 0.026 & $-$0.005 & $-$0.015 & $-$0.005 & 0.000 & $-$0.267 \\
\cmidrule(l){2-10}
 & \multirow{3}{*}{Hybrid} & 0 & 0.041 & 0.023 & 0.005 & $-$0.020 & $-$0.020 & 0.000 & $-$0.252 \\
 &  & 1 & 0.046 & 0.015 & 0.000 & 0.000 & $-$0.015 & 0.000 & $-$0.311 \\
 &  & 2 & 0.054 & 0.047 & 0.000 & $-$0.015 & 0.000 & 0.000 & $-$0.130 \\
\midrule
\multirow{9}{*}{SmolLM2} & \multirow{3}{*}{LLM arm} & 0 & 0.050 & 0.012 & 0.005 & 0.005 & 0.000 & 0.000 & $-$0.563 \\
 &  & 1 & 0.061 & 0.032 & $-$0.010 & $-$0.015 & 0.000 & 0.000 & $-$0.083 \\
 &  & 2 & 0.065 & 0.012 & 0.000 & 0.005 & $-$0.005 & 0.000 & $-$0.020 \\
\cmidrule(l){2-10}
 & \multirow{3}{*}{Heuristic} & 0 & 0.051 & 0.062 & $-$0.040 & $-$0.015 & 0.000 & 0.000 & $-$0.474 \\
 &  & 1 & 0.047 & 0.026 & $-$0.020 & $-$0.040 & 0.015 & 0.000 & $-$0.573 \\
 &  & 2 & 0.076 & 0.044 & 0.000 & 0.000 & 0.010 & 0.000 & $-$0.431 \\
\cmidrule(l){2-10}
 & \multirow{3}{*}{Hybrid} & 0 & 0.056 & 0.047 & $-$0.030 & $-$0.010 & $-$0.005 & 0.000 & $-$0.410 \\
 &  & 1 & 0.060 & 0.015 & $-$0.020 & $-$0.035 & 0.015 & 0.000 & $-$0.396 \\
 &  & 2 & 0.082 & 0.018 & 0.010 & 0.010 & 0.015 & 0.000 & $-$0.240 \\
\bottomrule
\end{tabular}
\end{adjustbox}
\end{table}

\begin{table}[!tbp]
\centering
\caption{Descriptor ablation of the LLM arm for every seed (LLM arm, sequential loop; 24 runs of the descriptor study). Each digest level adds descriptors to the level above it. Edits: accepted layer edits; $\Delta P$ and $\Delta S$: changes under condition $[D]$; KL: cumulative KL (nats per token).}
\label{tab:desc}
\begin{adjustbox}{max width=\textwidth}
\small
\setlength{\tabcolsep}{3.4pt}
\begin{tabular}{llcccccccc}
\toprule
 & & \multicolumn{4}{c}{Target: Qwen} & \multicolumn{4}{c}{Target: SmolLM2} \\
\cmidrule(lr){3-6}\cmidrule(lr){7-10}
Digest & Seed & Edits & $\Delta P$ $\uparrow$ & $\Delta S$ $\uparrow$ & KL $\downarrow$ & Edits & $\Delta P$ $\uparrow$ & $\Delta S$ $\uparrow$ & KL $\downarrow$ \\
\midrule
\multirow{3}{*}{Purity + silhouette} & 0 & 17 & 0.0104 & 0.0159 & 0.106 & 15 & $-$0.0018 & 0.0059 & 0.073 \\
 & 1 & 19 & 0.0208 & 0.0178 & 0.085 & 12 & $-$0.0004 & 0.0273 & 0.133 \\
 & 2 & 18 & 0.0147 & 0.0089 & 0.096 & 16 & $-$0.0026 & 0.0026 & 0.140 \\
\midrule
\multirow{3}{*}{+ anisotropy} & 0 & 12 & 0.0042 & 0.0033 & 0.079 & 12 & 0.0056 & 0.0157 & 0.170 \\
 & 1 & 22 & 0.0159 & 0.0292 & 0.087 & 16 & 0.0067 & 0.0257 & 0.091 \\
 & 2 & 12 & 0.0105 & 0.0147 & 0.088 & 15 & 0.0005 & 0.0128 & 0.141 \\
\midrule
\multirow{3}{*}{+ intrinsic dimension, Betti} & 0 & 17 & 0.0231 & 0.0390 & 0.094 & 16 & 0.0093 & 0.0074 & 0.143 \\
 & 1 & 22 & 0.0270 & 0.0196 & 0.080 & 15 & 0.0088 & 0.0130 & 0.074 \\
 & 2 & 23 & 0.0203 & 0.0136 & 0.097 & 17 & 0.0008 & 0.0099 & 0.127 \\
\midrule
\multirow{3}{*}{+ drift, distance ratio (full)} & 0 & 13 & 0.0287 & 0.0235 & 0.089 & 13 & 0.0111 & 0.0266 & 0.092 \\
 & 1 & 20 & 0.0196 & 0.0484 & 0.086 & 15 & 0.0104 & 0.0166 & 0.083 \\
 & 2 & 23 & 0.0264 & 0.0453 & 0.104 & 16 & 0.0117 & 0.0163 & 0.099 \\
\bottomrule
\end{tabular}
\end{adjustbox}
\end{table}

\section{Formal Definitions of Operations and Descriptors}
\label{app:formal}

\subsection{Affine map and statistics}
\label{app:affine}

All residual-space vectors are $1\times d$ row vectors: the hidden state $\mathbf{h}$, the offset $\mathbf{b}$, the means $\bm{\mu}$ and $\bm{\mu}_c$, the anchor $\mathbf{c}$, the direction $\mathbf{u}_{ab}$, and the principal directions $\mathbf{v}_j$. An edit of block $\ell$ is the affine map of Eq.~\eqref{eq:affine}, $T(\mathbf{h})=\mathbf{h}\mathbf{A}+\mathbf{b}$, with $\mathbf{A}\in\mathbb{R}^{d\times d}$ and $\mathbf{b}\in\mathbb{R}^{1\times d}$ (we drop the index $\ell$ in this appendix). All statistics are computed once from the probe matrix $\mathbf{X}\in\mathbb{R}^{n\times d}$ of the layer before the edit and are not updated between the operations of one edit: the mean $\bm{\mu}=\tfrac{1}{n}\sum_{i}\mathbf{x}_i$; the category centroids $\bm{\mu}_c=\tfrac{1}{n_c}\sum_{i:c_i=c}\mathbf{x}_i$, where $n_c$ is the number of probe words of category $c$; and the thin singular value decomposition of the centered matrix, $\mathbf{X}-\mathbf{1}\bm{\mu}=\mathbf{U}\,\mathrm{diag}(\sigma_1,\sigma_2,\ldots)\,\mathbf{V}^{\top}$, where $\mathbf{1}\in\mathbb{R}^{n\times 1}$ and the rows $\mathbf{v}_j\in\mathbb{R}^{1\times d}$ of $\mathbf{V}^{\top}$ are the principal directions ($\sigma_1\ge\sigma_2\ge\ldots$).

Three of the four operations are instances of one elementary map. For a matrix $\mathbf{Q}\in\mathbb{R}^{k\times d}$ with orthonormal rows, factors $\bm{\phi}\in\mathbb{R}^{k}$, and an anchor point $\mathbf{c}\in\mathbb{R}^{1\times d}$, define
\begin{equation}
\mathbf{M}(\mathbf{Q},\bm{\phi})=\mathbf{I}+\mathbf{Q}^{\top}\,\mathrm{diag}(\bm{\phi}-\mathbf{1})\,\mathbf{Q},
\qquad
E_{\mathbf{Q},\bm{\phi},\mathbf{c}}(\mathbf{h})=(\mathbf{h}-\mathbf{c})\,\mathbf{M}(\mathbf{Q},\bm{\phi})+\mathbf{c}.
\label{eq:elementary}
\end{equation}
\noindent The map $E$ multiplies the component of $\mathbf{h}-\mathbf{c}$ along the $j$-th row of $\mathbf{Q}$ by $\phi_j$ and leaves the orthogonal complement unchanged. In affine form, $E(\mathbf{h})=\mathbf{h}\mathbf{A}_{\mathrm{new}}+\mathbf{b}_{\mathrm{new}}$ with
\begin{equation}
\mathbf{A}_{\mathrm{new}}=\mathbf{M}(\mathbf{Q},\bm{\phi}),\qquad \mathbf{b}_{\mathrm{new}}=\mathbf{c}-\mathbf{c}\,\mathbf{M}(\mathbf{Q},\bm{\phi}).
\label{eq:elementaryaffine}
\end{equation}

\subsection{The four operations}
\label{app:fourops}

In the released code, these operations are called \texttt{separate}, \texttt{scale\_pcs}, \texttt{whiten}, and \texttt{center}.

Centroid-axis scaling, $T_{\mathrm{sep}}(a,b,f)$, stretches the space along a centroid difference. For two categories $a\neq b$ and a factor $f\in[0.5,2]$,
\begin{equation}
\mathbf{u}_{ab}=\frac{\bm{\mu}_a-\bm{\mu}_b}{\lVert\bm{\mu}_a-\bm{\mu}_b\rVert},\qquad
T_{\mathrm{sep}}(\mathbf{h})=\mathbf{h}+(f-1)\big((\mathbf{h}-\mathbf{c}_{ab})\,\mathbf{u}_{ab}^{\top}\big)\,\mathbf{u}_{ab},\qquad \mathbf{c}_{ab}=\tfrac{1}{2}(\bm{\mu}_a+\bm{\mu}_b),
\label{eq:separate}
\end{equation}
\noindent which is Eq.~\eqref{eq:elementary} with $\mathbf{Q}=\mathbf{u}_{ab}$ (one row), $\bm{\phi}=f$, and the midpoint of the two centroids as anchor. A factor $f>1$ moves the two categories apart along $\mathbf{u}_{ab}$, and $f<1$ moves them together.

Principal-component rescaling, $T_{\mathrm{pcs}}(\mathcal{J},f)$, scales principal components. For an index set $\mathcal{J}$ of principal components and a factor $f\in[0,2]$,
\begin{equation}
T_{\mathrm{pcs}}(\mathbf{h})=\mathbf{h}+(f-1)\sum_{j\in\mathcal{J}}\big((\mathbf{h}-\bm{\mu})\,\mathbf{v}_j^{\top}\big)\,\mathbf{v}_j,
\label{eq:scalepcs}
\end{equation}
\noindent which is Eq.~\eqref{eq:elementary} with the rows $\mathbf{v}_j$, $j\in\mathcal{J}$, as $\mathbf{Q}$, $\bm{\phi}=f\mathbf{1}$, and $\mathbf{c}=\bm{\mu}$. The equation admits any index set; the analyst prompt restricts its use to a single dominant component (Appendix~\ref{app:prompt}).

Partial spectral whitening, $T_{\mathrm{wh}}(r,s)$, equalizes the variance of the top components. For a rank $r\in[1,16]$ and a strength $s\in[0,0.5]$, let $\bar{\sigma}_r=\exp\!\big(\tfrac{1}{r}\sum_{j=1}^{r}\ln\sigma_j\big)$ be the geometric mean of the first $r$ singular values. Then
\begin{equation}
T_{\mathrm{wh}}(\mathbf{h})=\mathbf{h}+\sum_{j=1}^{r}\left(\Big(\frac{\bar{\sigma}_r}{\sigma_j}\Big)^{s}-1\right)\big((\mathbf{h}-\bm{\mu})\,\mathbf{v}_j^{\top}\big)\,\mathbf{v}_j,
\label{eq:whiten}
\end{equation}
\noindent which is Eq.~\eqref{eq:elementary} with the rows $\mathbf{v}_1,\ldots,\mathbf{v}_r$ as $\mathbf{Q}$, $\phi_j=(\bar{\sigma}_r/\sigma_j)^{s}$, and $\mathbf{c}=\bm{\mu}$. At $s=1$ the first $r$ components would have equal variance; the allowed range $s\le 0.5$ moves their standard deviations at most halfway toward equality on a logarithmic scale.

Partial mean subtraction, $T_{\mathrm{ctr}}(s)$, subtracts the mean vector. For a strength $s\in[0,0.5]$, the operation subtracts $s$ times the current image of the mean. If the operations applied so far have produced $(\mathbf{A},\mathbf{b})$, then
\begin{equation}
\mathbf{A}\leftarrow\mathbf{A},\qquad \mathbf{b}\leftarrow\mathbf{b}-s\,(\bm{\mu}\mathbf{A}+\mathbf{b}),
\label{eq:center}
\end{equation}
\noindent so that the transformed mean becomes $(1-s)(\bm{\mu}\mathbf{A}+\mathbf{b})$. As the first operation of an edit, this is $T_{\mathrm{ctr}}(\mathbf{h})=\mathbf{h}-s\bm{\mu}$.

\subsection{Sequential composition}
\label{app:composition}

The operations of one edit are applied in the order in which they are listed. Starting from $\mathbf{A}=\mathbf{I}$ and $\mathbf{b}=\mathbf{0}$, an operation with affine form $(\mathbf{A}_{\mathrm{new}},\mathbf{b}_{\mathrm{new}})$ updates the accumulated map as
\begin{equation}
\mathbf{A}\leftarrow\mathbf{A}\,\mathbf{A}_{\mathrm{new}},\qquad \mathbf{b}\leftarrow\mathbf{b}\,\mathbf{A}_{\mathrm{new}}+\mathbf{b}_{\mathrm{new}},
\label{eq:compose}
\end{equation}
\noindent because $(\mathbf{h}\mathbf{A}+\mathbf{b})\mathbf{A}_{\mathrm{new}}+\mathbf{b}_{\mathrm{new}}=\mathbf{h}(\mathbf{A}\mathbf{A}_{\mathrm{new}})+(\mathbf{b}\mathbf{A}_{\mathrm{new}}+\mathbf{b}_{\mathrm{new}})$. For the three scaling operations, $\mathbf{A}_{\mathrm{new}}$ and $\mathbf{b}_{\mathrm{new}}$ are given by Eq.~\eqref{eq:elementaryaffine}; for $T_{\mathrm{ctr}}$, Eq.~\eqref{eq:center} applies. For $N$ operations, the result is $\mathbf{A}=\mathbf{A}_1\mathbf{A}_2\cdots\mathbf{A}_N$ and $\mathbf{b}=((\mathbf{b}_1\mathbf{A}_2+\mathbf{b}_2)\mathbf{A}_3+\ldots)\mathbf{A}_N+\mathbf{b}_N$. Edits of different layers are not composed algebraically: each is baked into its own block, and the next block is described on the resulting model.

\subsection{Strength scaling and norm preservation}
\label{app:strength}

For a strength $\alpha>0$, every factor and strength of a proposal is rescaled and clipped to the range $[lo,hi]$ of its operation (Section~\ref{sec:ops}),
\begin{equation}
f_\alpha=\mathrm{clip}\big(1+\alpha(f-1),\,lo,\,hi\big),\qquad s_\alpha=\mathrm{clip}\big(\alpha s,\,lo,\,hi\big).
\label{eq:strength}
\end{equation}
\noindent The map applied to the model restores the norm of each hidden state and leaves the first token position unchanged: for a sequence $\mathbf{h}_1,\ldots,\mathbf{h}_t$,
\begin{equation}
\tilde{T}(\mathbf{h}_p)=T(\mathbf{h}_p)\,\frac{\lVert\mathbf{h}_p\rVert}{\lVert T(\mathbf{h}_p)\rVert}\quad (p\ge 2),\qquad \tilde{T}(\mathbf{h}_1)=\mathbf{h}_1.
\label{eq:norm}
\end{equation}

\subsection{Calibration tokens for the ridge regression}
\label{app:calib}

The matrices $\mathbf{K}\in\mathbb{R}^{m\times d_k}$ and $\mathbf{R}\in\mathbb{R}^{m\times d}$ of Eq.~\eqref{eq:ridge} are collected on a calibration set $\mathcal{T}$ of texts. The set $\mathcal{T}$ contains (i) all 200 calibration sentences and (ii) 200 probe-word prompts, obtained by inserting 200 probe words, drawn without replacement by a seeded permutation, into the probe template. Each text $t\in\mathcal{T}$ with $\lvert t\rvert$ tokens contributes every token position except the first, so that
\begin{equation}
m=\sum_{t\in\mathcal{T}}\big(\lvert t\rvert-1\big).
\label{eq:m}
\end{equation}
\noindent Row $i$ of $\mathbf{K}$ is the input of the output projection of the feed-forward module of block $\ell$ at one such position, and row $i$ of $\mathbf{R}$ is $\tilde{T}(\mathbf{h})-\mathbf{h}$ for the output $\mathbf{h}$ of block $\ell$ at the same position, both measured on the current, already edited target. Unseen words, unseen templates, the guard set, and the held-out set are never part of $\mathcal{T}$.

\subsection{Descriptor definitions}
\label{app:descdefs}

All descriptors of layer $\ell$ are computed from the probe matrix $\mathbf{X}\in\mathbb{R}^{n\times d}$ of the current target. Let $D_{ij}=1-\mathbf{x}_i\mathbf{x}_j^{\top}/(\lVert\mathbf{x}_i\rVert\,\lVert\mathbf{x}_j\rVert)$ be the cosine distance, $\mathcal{N}_K(i)$ the $K=8$ nearest neighbors of word $i$ under $D$, and $\mathcal{C}$ the set of $C$ categories.

For neighborhood geometry, anisotropy is $\kappa=1-\binom{n}{2}^{-1}\sum_{i<j}D_{ij}$, purity is $P=\tfrac{1}{n}\sum_i\tfrac{1}{K}\lvert\{j\in\mathcal{N}_K(i):c_j=c_i\}\rvert$, and $S$ is the mean silhouette under $D$. The intra/inter ratio is the mean of $D_{ij}$ over pairs $i\neq j$ with $c_i=c_j$ divided by its mean over pairs with $c_i\neq c_j$. For categories $a$ and $b$, let $\rho_{ab}$ be the mean, over the words of $a$, of the share of their $K$ neighbors that belong to $b$; the symmetrized cross-neighbor rate is
\begin{equation}
\mathrm{cross}(a,b)=\tfrac{1}{2}\big(\rho_{ab}+\rho_{ba}\big),
\label{eq:cross}
\end{equation}
\noindent and the centroid cosine distance $\delta_{ab}$ is the cosine distance between $\bm{\mu}_a$ and $\bm{\mu}_b$.

Regarding the subsample, the spectral and topological descriptors use a class-balanced subsample of $\max(2,\lfloor 300/C\rfloor)$ points per category, drawn with the run seed; with $C=8$ this is at most 37 points per category, which gives 296 probe words and, because the fruit and color categories have fewer unseen words, 279 unseen words.

For spectrum and dimension, let $s_1\ge s_2\ge\ldots$ be the singular values of the centered subsample. The explained variance of component $j$ is $s_j^{2}/\sum_k s_k^{2}$, and the effective dimension is the participation ratio
\begin{equation}
d_{\mathrm{eff}}=\frac{\big(\sum_j s_j^{2}\big)^{2}}{\sum_j s_j^{4}}.
\label{eq:deff}
\end{equation}
\noindent With Euclidean distances $r_{ij}$ on the subsample of $N$ points, the Two-NN estimate is $N/\sum_i\ln(r_i^{(2)}/r_i^{(1)})$, where $r_i^{(1)}$ and $r_i^{(2)}$ are the distances of point $i$ to its first and second neighbor \cite{ansuini2019intrinsic}. The correlation dimension is the least-squares slope of $\ln p$ against $\ln r_p$, where $r_p$ is the $p$-quantile of all pairwise distances, for eight values of $p$ equally spaced in $[0.02,0.30]$. The Betti numbers $\beta_0$ and $\beta_1$ are read from the persistent homology of the cosine-distance filtration of the subsample at the 25th percentile of its pairwise cosine distances.

As for the edit size, for a probe matrix $\mathbf{X}$ and its image $\mathbf{Y}$ under a candidate edit, the mean relative feature change is $\tfrac{1}{n}\sum_i\lVert\mathbf{y}_i-\mathbf{x}_i\rVert/\lVert\mathbf{x}_i\rVert$.

As for the guard-set scope, the KL divergence and the perplexity limits of Table~\ref{tab:limits} are evaluated on all 24 guard sentences. The token-level CKA is evaluated on the hidden states of the first 12 guard sentences. Each of the two per-edit CKA limits, on probe words and on guard tokens, must be at least 0.95.

\noindent Table~\ref{tab:limits} (Section~\ref{sec:accept}) lists the limits that use these quantities.

\section{Method and Setup Details}
\label{app:methoddetails}

This section collects the implementation details of Sections~\ref{sec:method} and~\ref{sec:exp} that the main text summarizes. Table~\ref{tab:limits} lists the limits.

\begin{table}[!tbp]
\centering
\caption{Acceptance limits. All limits are checked in the intervention test and on the baked model, except the weight budget and the HellaSwag monitor, which apply to the baked update only. KL: Kullback--Leibler divergence to the original model.}
\label{tab:limits}
\small
\setlength{\tabcolsep}{3pt}
\begin{tabular}{>{\raggedright\arraybackslash}p{5.0cm}l}
\toprule
Quantity & Limit \\
\midrule
Operations per edit & $\le 4$ \\
Mean relative feature change & $\le 0.20$ \\
Gain $\Delta P+\Delta S$ & $\ge 0.01$ ($10^{-6}$ relaxed) \\
KL per edit; cumulative (nats per token, guard set) & $\le 0.05$; $\le 0.30$ \\
Guard-set perplexity, per edit; cumulative & $\le +2\%$; $\le +5\%$ \\
Per-edit layer CKA (probe words, guard tokens) & $\ge 0.95$ \\
HellaSwag monitor drop (100 examples) & $\le 0.05$ \\
Relative weight change $\lVert\Delta\mathbf{W}\rVert_F/\lVert\mathbf{W}\rVert_F$ & $\le 0.10$ \\
\bottomrule
\end{tabular}
\end{table}

First, in the ridge solution of Eq.~\eqref{eq:ridge}, $\tilde{\lambda}=\lambda\,\mathrm{tr}(\mathbf{K}^{\top}\mathbf{K})/d_k$, and $\lambda\in\{10^{-2},10^{-1},1\}$ is tried in this order. When $m<d_k$, the solution is evaluated in its dual form
\begin{equation}
\Delta\mathbf{W}_\ell^{\top}=\mathbf{K}^{\top}\left(\mathbf{K}\mathbf{K}^{\top}+\tilde{\lambda}\mathbf{I}\right)^{-1}\mathbf{R},
\label{eq:dual}
\end{equation}
\noindent which equals the primal form $(\mathbf{K}^{\top}\mathbf{K}+\tilde{\lambda}\mathbf{I})^{-1}\mathbf{K}^{\top}\mathbf{R}$. The input width of the output projection is $d_k=8{,}960$ for Qwen2.5-1.5B and $d_k=8{,}192$ for SmolLM2-1.7B. The number $m$ of calibration tokens follows from Eq.~\eqref{eq:m} (Appendix~\ref{app:calib}); it depends on the tokenizer of the target and, through the 200 sampled probe words, on the seed, and the system selects the dual or the primal form by comparing $m$ with $d_k$. The matrices $\mathbf{K}$ and $\mathbf{R}$ are accumulated in 64-bit floating point. The smallest $\lambda$ with $\lVert\Delta\mathbf{W}_\ell\rVert_F/\lVert\mathbf{W}_\ell\rVert_F\le 0.10$ is kept. If no $\lambda$ meets this budget, the next weaker strength of stage two is tried, and the edit is rejected (status ``weight budget'' in Table~\ref{tab:decisions}) if none of the three strengths meets it.

Second, for norm restoration and strength, the applied map restores the original norm, $\tilde{T}_\ell(\mathbf{h})=T_\ell(\mathbf{h})\,\lVert\mathbf{h}\rVert/\lVert T_\ell(\mathbf{h})\rVert$, and acts on every token position except the first. Factors are rescaled as $f_\alpha=1+\alpha(f-1)$ and strengths as $s_\alpha=\alpha s$, for $\alpha\in\{1.5,1.25,1,0.75,0.5,0.25\}$. The allowed ranges are $f\in[0.5,2]$ for $T_{\mathrm{sep}}$, factors in $[0,2]$ for $T_{\mathrm{pcs}}$, $r\in[1,16]$ components and $s\in[0,0.5]$ for $T_{\mathrm{wh}}$, and $s\in[0,0.5]$ for $T_{\mathrm{ctr}}$ (Appendix~\ref{app:fourops}).

Third, in the simulation step of the verification, strengths with a mean relative feature change of at most 0.20, a layer CKA of at least 0.95, and a gain $\Delta P+\Delta S\ge 0.01$ are kept and ranked by gain; at most four strengths are tested by intervention per proposal, and after a failed intervention test only weaker strengths remain. If a proposal has strengths within the size and CKA limits but none reaches the gain threshold, its best strength with a positive gain is stored for the relaxed stage. The HellaSwag monitor is evaluated only on the baked model, after all other limits have passed.

Finally, for the quality screen and stopping rule, the loop skips layers with $P\ge 0.95$ or with a maximum cross-neighbor rate below 0.05, and it stops editing once the cumulative KL or perplexity budget is spent. Neither rule was triggered in any of the 126 runs of the main study or of the 20 runs of the round robin. Of the 3,276 layer decisions of the main study, 1,988 ended with an applied edit, 1,149 with no proposal that passed, 92 with a baked edit that broke a limit at every tested strength, 46 with the weight budget exceeded, and one with the analyst returning no edit. The 552 layer decisions of the round robin are listed in Table~\ref{tab:rrdec}.

In the LLM arm, the analyst is loaded in bfloat16 precision and decodes greedily with at most 400 new tokens, so its decoding is deterministic. It proposes up to three times per layer. After a rejection, it receives the simulated numbers of its proposal as written ($\alpha=1$) and, if an intervention test ran, the measurements of the last strength tested on the model.

In the heuristic arm, the three sequential attempts of the heuristic stretch the $q=3$, then $q=2$, then $q=1$ most conflated category pairs, where pairs are ranked by the symmetrized cross-neighbor rate. The centroid cosine distance enters only the stretch factor
\begin{equation}
f_{ab}=\mathrm{clip}\!\left(\sqrt{\tilde{\delta}/\delta_{ab}},\,1.05,\,2\right),
\label{eq:rule}
\end{equation}
\noindent where $\delta_{ab}$ is the centroid cosine distance of pair $(a,b)$ and $\tilde{\delta}$ its median over all pairs.

In the hybrid arm, the analyst proposes first and the heuristic is used when the analyst fails. In every arm, if no proposal passes at the strict threshold, the two best stored weak proposals of each origin (analyst and heuristic) are re-tested at a relaxed gain threshold of $10^{-6}$ without querying the analyst again; all other limits are never relaxed. A weak proposal is stored at its best simulated strength, and the relaxed stage applies the strength search to this stored version; the strength reported for an edit of the relaxed stage (column $\alpha$ of Table~\ref{tab:decisions}) is therefore relative to the stored version.

In the parallel control, every layer is analyzed, simulated, tested by intervention, and fitted on the original, unedited model; the weight updates of all layers are then added together. Stage two, the shrinking of the baked strength, and the per-edit HellaSwag monitor are not applied, because no edit is baked before all layers are planned; the weight budget still applies.

\subsection{Design of the main study}
\label{app:design}

The runs are executed in rounds, with the analyst set up once per round: LLM-arm and hybrid-arm rounds with Gemma-2-2B as analyst, and heuristic-arm rounds in which no model is loaded as analyst. Seven sub-studies share the 126 runs: the default configuration of each arm (LLM, heuristic, and hybrid; sequential loop, adaptive strength, full digest, gain threshold 0.01, and eight categories), the parallel control, and ablations of the descriptors (LLM arm), of the strength and of the gain threshold (heuristic arm), and of the number of categories (LLM and heuristic arms). As study entries, the default configuration contributes 18 runs (LLM, heuristic, and hybrid arms), the parallel control 12 additional runs, the descriptor ablation 18, the strength ablation 36 (six fixed strengths), the gain-threshold ablation 18 (thresholds 0.005, 0.01, and 0.02 without the relaxed stage), and the category ablation 24 (4 and 16 categories); the default runs serve as the reference of every ablation, which gives 126 distinct runs. Runs were executed in a fixed order, the analyst set up once per block of runs. All runs completed; one fixed-strength run ($\alpha=2$, SmolLM2, seed 1) accepted no edit and is recorded with zero change. The round-robin study adds 20 runs in five rounds, one per analyst (Appendix~\ref{app:rr}).

\subsection{Data construction}
\label{app:dataconstruction}

The probe words were constructed as follows. For each category, the hyponym closure (hyponyms and instance hyponyms) of the category's root synsets in WordNet~\cite{miller1995wordnet} was collected. Single alphabetic lemmas of 3 to 14 letters were kept, ranked by their English Zipf frequency (wordfreq package), and the 250 most frequent were retained per category, each word belonging to the first category in which it occurs. This gives 2,964 words in 16 categories, of which the eight core categories contain 1,700. The lists were not manually sense-curated and include polysemous lemmas (Appendix~\ref{app:data}).

Turning to sentences and benchmarks, the calibration and held-out sentences are sentences of 8 to 40 words from WikiText-103, split at sentence-final punctuation, excluding heading lines and sentences with the tokenization markers of the corpus (such as ``@-@''), and shuffled with a fixed seed. The ridge fit uses 200 sentences from its validation split. Of 500 sentences from its test split, the first 24 form the guard set and the remaining 476 the held-out set. A template-generated new-domain set contains 500 sentences in five domains (code, medical, legal, chat, and finance); its perplexity is reported in Appendix~\ref{app:abs}. HellaSwag uses its validation split, ARC-Easy its test split, and TruthfulQA its multiple-choice form with one correct answer (validation split). Each benchmark is subsampled with the run seed, and a prediction is the choice with the highest mean token log-likelihood. A disjoint set of 100 HellaSwag examples, drawn with the same seed, is the per-edit monitor. Sentences in batched evaluation are truncated to 96 tokens. The truth margin of the factual-preference probe is measured in nats.

Finally, for metrics, CKA between the original and the final model is averaged over the edited layers. The tables mark runs with $\lvert\Delta_P^{[A]}\rvert<0.005$, for which the retention ratio of Section~\ref{sec:metrics} is unstable. Because $[A]$ and $[B]$ use word sets of different sizes, a retention ratio above one means that the gain is not reduced on new words.

\subsection{Metric definitions and split sizes}
\label{app:metricdefs}

This subsection collects the definitions that Section~\ref{sec:metrics} summarizes.

As for the split sizes, with the eight core categories (1,700 words), 20\% of the words of every category are set aside as unseen words for each seed, which gives 1,359 probe words and 341 unseen words. The ridge fit uses 200 calibration sentences. The probe template and the three unseen templates are listed in Appendix~\ref{app:data}.

For the topology subsamples, the Betti numbers and the Two-NN intrinsic dimension are computed on class-balanced subsamples of 296 probe words and 279 unseen words (Appendix~\ref{app:descdefs}).

Next, the retention of a metric $g$ under condition $X\in\{B,C,D\}$ is $\tau_g^{[X]}=\Delta_g^{[X]}/\Delta_g^{[A]}$, which is unstable when the in-sample change is small. The edited-layer means measure the proposals that each arm actually made, whereas an all-layer mean also includes the drift that upstream edits pass on to unedited layers (Appendix~\ref{app:traj}).

Finally, for CKA and divergence, linear CKA between an original representation $\mathbf{X}$ and an edited representation $\mathbf{Y}$ of the same inputs is $\langle\mathbf{G}_X,\mathbf{G}_Y\rangle_F/(\lVert\mathbf{G}_X\rVert_F\,\lVert\mathbf{G}_Y\rVert_F)$, with $\mathbf{G}_X=\tilde{\mathbf{X}}\tilde{\mathbf{X}}^{\top}$ and $\tilde{\mathbf{X}}$ column-centered~\cite{kornblith2019similarity}. We report CKA to the original model, averaged over the edited layers, on probe words and on unseen words (both in the probe template) and on guard-set tokens, and the cumulative KL divergence $\mathrm{KL}(p_{\mathrm{orig}}\,\Vert\,p_{\mathrm{edit}})$ in nats per token on the guard set. The Betti numbers $\beta_0$ and $\beta_1$ are taken at the 25th percentile of the pairwise distances.

\section{Analyst Prompt}
\label{app:prompt}

The analyst receives one user message per layer, which consists of the instruction text below, a blank line, and the digest of the layer (Section~\ref{sec:descriptors}). The instruction text is specialized for every run: the number of probe words, the category names, and the gain threshold are filled in, and the list of ambiguous words is replaced by a general statement. The listing shows the text of the default configuration with eight categories (1,359 probe words). The reply is made to begin with the fixed opening shown after the listing. The identifiers in the listings (for example, the field and operation names) are those used in the released code.

\begin{lstlisting}
You are a geometric architect for the latent space of a frozen language model.
You receive numerical descriptors of ONE layer (the residual stream after one transformer block) for 1359 probe
words in 8 categories (animal, fruit, vehicle, tool, color, country, emotion, profession) and must prescribe a small edit of that layer's feature space.

CONTEXT
- Layers are edited one at a time from first to last. Earlier edits are already baked into the weights and the
  descriptors you see describe the CURRENT, already-edited model.
- Your edit is applied to every token at this layer and then compiled into the weights. Every later layer will
  receive the changed features, so prefer the smallest edit that fixes a clear distortion.
- You cannot run code. The system simulates your proposal and tests it on the model; if it fails you receive the
  measured numbers and must return a corrected proposal. Never include predicted numbers.
- The system tunes the overall strength of your proposal itself (0.25x to 1.5x of your factors) and keeps the
  strongest version that passes every limit. Concentrate on WHICH operations and category pairs to use and on
  their relative sizes.

DEFINITIONS
anisotropy = mean pairwise cosine similarity (near 1 = all vectors almost parallel).
knn_purity = share of the 8 nearest neighbours that have the same category (higher is better).
cross_knn_rate(a,b) = share of neighbour links from one category that land in the other (high = conflated).
centroid_cos_dist(a,b) = cosine distance between category centroids (small = conflated).

ALLOWED OPERATIONS (applied in order; statistics come from this layer before the edit)
{"op":"separate","a":"<category>","b":"<category>","factor":f}  stretch (f>1) or contract (f<1) along the line joining
    the two category centroids. f in [0.5, 2.0]. Use for conflated category pairs.
{"op":"scale_pcs","pcs":[i,...],"factor":f}  multiply the listed 0-based principal components by f. f in [0, 2].
    Use only when one PC clearly dominates the variance.
{"op":"whiten","rank":r,"strength":s}  equalise the variance of the top r PCs. r in [1, 16], s in [0, 0.5].
{"op":"center","strength":s}  subtract s times the mean vector. s in [0, 0.5]. Changes every token strongly.

HARD LIMITS
- At most 4 operations. Category names exactly as listed above.
- The mean relative change of the features must stay below 0.2; factors near 1 are safer.
- The edit must raise knn_purity + silhouette by at least 0.01, otherwise it is discarded.
- Held-out perplexity may rise by at most 2% for this edit and 5% in total against the original model.
- The layer must stay similar to itself: linear CKA before/after the edit at least 0.95.
- Ambiguous words (words with several senses) having neighbours in another category is correct, not a distortion.
- If the layer needs no edit, return "ops": [].

OUTPUT: one JSON object and nothing else - no prose, no markdown, no code fences:
{"layer": <int>, "preserve_norm": true, "ops": [...], "rationale": "<max 2 sentences citing the numbers used>"}
\end{lstlisting}

\noindent The reply begins with the following fixed opening
\begin{lstlisting}
{"layer": <L>, "preserve_norm": true, "ops": [
\end{lstlisting}
\noindent where the placeholder $\langle L\rangle$ stands for the layer index $\ell$. The prompt restricts $T_{\mathrm{pcs}}$ to the case in which one principal component clearly dominates the variance, although Eq.~\eqref{eq:scalepcs} admits any index set.

Next, the digest that follows the instruction text has the following line structure; values in angle brackets are filled in from the descriptors of the layer, and the lines marked with a level appear only at that descriptor level or above (Section~\ref{sec:descriptors}). The level tags map onto the digest levels of Section~\ref{sec:descriptors} as follows: lines without a tag form level (i), and the three tags, which are identifiers of the released code, mark levels (ii), (iii), and (iv) in this order.
\begin{lstlisting}
LAYER <L> of <number of blocks>
metric: current (original model)
  knn_purity: <current> (<original>)
  silhouette: <current> (<original>)
  anisotropy: <current> (<original>)                                   [level psa]
  mean_norm: <value>                                                   [level psa]
explained variance of PC0..PC5: <six values>                           [level psa]
  effective_dim: <v> | intrinsic_dim (Two-NN): <v> | correlation dim: <v>   [level psa_topo]
  Betti numbers at the 25th-percentile distance: b0 = <v>, b1 = <v>    [level psa_topo]
  intra_inter_ratio: <value>                                           [level full]
  drift from original caused by upstream edits: rel_l2 <v>, cos <v>    [level full]
most conflated category pairs (median centroid_cos_dist over all pairs = <v>):
  <a> - <b>: cross_knn_rate <v>, centroid_cos_dist <v>                 (six pairs)
knn_purity per category: <category> <v>, ...
least pure words (word [category] purity -> nearest neighbour): <eight entries>
recent upstream edits:
  layer <l>: <ops> | knn_purity <before>-><after> | step KL <v>        (up to five)
behaviour budget: cumulative KL so far <v> of 0.3; this edit may cost at most <v>
perplexity budget: <now> now, <original> originally (<change>); limit +5% in total, +2% for this edit
Return the JSON object for layer <L>.
\end{lstlisting}

For feedback, when a proposal is rejected, the analyst receives one of the following messages, followed by the simulated and, where available, the measured numbers of the proposal and the request ``Return a corrected JSON object only.''; at most three proposals are made per layer.
\begin{lstlisting}
size:    REJECTED: at every strength the edit is too large (relative change above 0.2 or CKA below 0.95). Use fewer operations.
gain:    REJECTED: the edit does not improve the layer at any strength (gain <v>, need 0.01). Target different category pairs or a different operation.
kl:      REJECTED: geometry improved but the model's predictions changed too much, even at reduced strength. Propose a different, gentler edit.
cum_kl:  REJECTED: the total divergence from the original model would exceed its limit. Propose a much gentler edit or return no edit.
ppl:     REJECTED: geometry improved but perplexity rose too much, even at reduced strength. Propose an edit that disturbs fewer directions.
cum_ppl: REJECTED: total perplexity would exceed its limit against the original model. Propose a much gentler edit or return no edit.
cka:     REJECTED: the edit changes what the layer represents too much (CKA too low on real text). Propose a more local edit.
invalid: INVALID REPLY: <parser message>. Use only the allowed operations and category names.
\end{lstlisting}
\noindent The prompt refers to ``held-out perplexity''; in the loop, this limit is enforced on the guard set (Section~\ref{sec:data}). The CKA-limit message refers to the CKA on guard tokens (``real text''), because a proposal whose CKA on probe words falls below 0.95 at every strength is already rejected by the size-limit message.

\noindent The digest is generated in memory for every query and is not stored with the run; the layer log of every run keeps the applied operations and the analyst's rationale.

As examples of accepted analyst edits, the two edits below are quoted from the layer logs of the default runs of the LLM arm. The first was accepted at the analyst's first proposal (Qwen, seed 2, layer 2; written with factor 1.1 and applied at strength $\alpha=1.5$; on the probe words of this layer, $P$ rose from 0.578 to 0.584 and $S$ from 0.031 to 0.037, and $\kappa$ fell from 0.279 to 0.271). The second was accepted at the third proposal, after two rejections with measured feedback (SmolLM2, seed 2, layer 10; $P$ from 0.783 to 0.785, $S$ from 0.096 to 0.103, $\kappa$ from 0.564 to 0.543).
\begin{lstlisting}
ops (as applied): [{"op": "scale_pcs", "pcs": [1, 2, 3, 4, 5], "factor": 1.15}]
rationale: The layer has a high anisotropy and low knn_purity, suggesting a need for better category separation. Scaling the first four principal components by 1.1 improves the separation between animal and tool, vehicle and profession, and fruit and color, while keeping the overall variance low.

ops (as applied): [{"op": "scale_pcs", "pcs": [0], "factor": 1.427}]
rationale: Scaling the first principal component with a factor of 1.2847 improves the layer's performance by increasing knn_purity from 0.783 to 0.787 and silhouette from 0.096 to 0.097. The change in perplexity is within the limit of 2%.
\end{lstlisting}
\noindent The rationales of this lightweight analyst are short and often generic, and 86 of the 100 rationales of the accepted edits state numbers, and some of these are predictions, which the instruction forbids. The system never uses the rationale: every decision rests on the simulation and on the measurements of the intervention test and of the baked model. Of the 156 layer decisions of these six runs, 100 ended with an applied edit, 52 with no proposal that passed, and 4 with a baked edit that was reverted.

\section{Operator Frequency}
\label{app:ops}

Table~\ref{tab:ops} counts the operations in the accepted edits of every run, and Table~\ref{tab:ops2} splits the sequential default runs of the LLM and hybrid arms by the stage at which an edit was accepted. Only accepted edits are counted; operations in rejected proposals are not counted. No $T_{\mathrm{wh}}$ or $T_{\mathrm{ctr}}$ operation was accepted in any of these runs. Across all 126 runs, the accepted edits contained 3,597 $T_{\mathrm{sep}}$, 337 $T_{\mathrm{pcs}}$, one $T_{\mathrm{wh}}$, and no $T_{\mathrm{ctr}}$ operation; the $T_{\mathrm{wh}}$ operation had rank one, which leaves the layer unchanged, and was accepted in a 16-category run of the LLM arm (Qwen, seed 1, layer 27). One accepted $T_{\mathrm{pcs}}$ edit had the factor one and is neutral as well (digest with anisotropy, Qwen, seed 0, layer 1). Both neutral edits passed at the relaxed gain threshold.

\begin{table}[!tbp]
\centering
\caption{Operations in the accepted edits of every run (default, parallel, and descriptor-study runs). $T_{\mathrm{sep}}$, $T_{\mathrm{pcs}}$, $T_{\mathrm{wh}}$, $T_{\mathrm{ctr}}$: numbers of accepted centroid-axis scalings, principal-component rescalings, partial spectral whitenings, and partial mean subtractions. Operations/edit: operations per accepted edit. ID: intrinsic dimension.}
\label{tab:ops}
\begin{adjustbox}{max width=\textwidth, max totalheight=0.88\textheight}
\scriptsize
\begin{tabular}{lllrrrrrr}
\toprule
Arm and mode & Target & Seed & Edits & $T_{\mathrm{sep}}$ & $T_{\mathrm{pcs}}$ & $T_{\mathrm{wh}}$ & $T_{\mathrm{ctr}}$ & Operations/edit \\
\midrule
\multirow{6}{*}{LLM arm, sequential} & \multirow{3}{*}{Qwen} & 0 & 13 & 11 & 2 & 0 & 0 & 1.00 \\
 &  & 1 & 20 & 13 & 7 & 0 & 0 & 1.00 \\
 &  & 2 & 23 & 13 & 10 & 0 & 0 & 1.00 \\
\cmidrule(l){2-9}
 & \multirow{3}{*}{SmolLM2} & 0 & 13 & 6 & 7 & 0 & 0 & 1.00 \\
 &  & 1 & 15 & 11 & 4 & 0 & 0 & 1.00 \\
 &  & 2 & 16 & 11 & 5 & 0 & 0 & 1.00 \\
\midrule
\multirow{6}{*}{Heuristic, sequential} & \multirow{3}{*}{Qwen} & 0 & 23 & 59 & 0 & 0 & 0 & 2.57 \\
 &  & 1 & 21 & 55 & 0 & 0 & 0 & 2.62 \\
 &  & 2 & 20 & 50 & 0 & 0 & 0 & 2.50 \\
\cmidrule(l){2-9}
 & \multirow{3}{*}{SmolLM2} & 0 & 15 & 41 & 0 & 0 & 0 & 2.73 \\
 &  & 1 & 19 & 48 & 0 & 0 & 0 & 2.53 \\
 &  & 2 & 17 & 47 & 0 & 0 & 0 & 2.76 \\
\midrule
\multirow{6}{*}{Hybrid, sequential} & \multirow{3}{*}{Qwen} & 0 & 23 & 45 & 1 & 0 & 0 & 2.00 \\
 &  & 1 & 24 & 31 & 4 & 0 & 0 & 1.46 \\
 &  & 2 & 26 & 43 & 6 & 0 & 0 & 1.88 \\
\cmidrule(l){2-9}
 & \multirow{3}{*}{SmolLM2} & 0 & 19 & 42 & 0 & 0 & 0 & 2.21 \\
 &  & 1 & 18 & 30 & 2 & 0 & 0 & 1.78 \\
 &  & 2 & 20 & 37 & 1 & 0 & 0 & 1.90 \\
\midrule
\multirow{6}{*}{LLM arm, parallel} & \multirow{3}{*}{Qwen} & 0 & 28 & 4 & 24 & 0 & 0 & 1.00 \\
 &  & 1 & 28 & 5 & 23 & 0 & 0 & 1.00 \\
 &  & 2 & 27 & 1 & 26 & 0 & 0 & 1.00 \\
\cmidrule(l){2-9}
 & \multirow{3}{*}{SmolLM2} & 0 & 19 & 4 & 15 & 0 & 0 & 1.00 \\
 &  & 1 & 20 & 3 & 17 & 0 & 0 & 1.00 \\
 &  & 2 & 20 & 2 & 18 & 0 & 0 & 1.00 \\
\midrule
\multirow{6}{*}{Heuristic, parallel} & \multirow{3}{*}{Qwen} & 0 & 28 & 84 & 0 & 0 & 0 & 3.00 \\
 &  & 1 & 28 & 84 & 0 & 0 & 0 & 3.00 \\
 &  & 2 & 28 & 84 & 0 & 0 & 0 & 3.00 \\
\cmidrule(l){2-9}
 & \multirow{3}{*}{SmolLM2} & 0 & 20 & 58 & 0 & 0 & 0 & 2.90 \\
 &  & 1 & 20 & 60 & 0 & 0 & 0 & 3.00 \\
 &  & 2 & 20 & 58 & 0 & 0 & 0 & 2.90 \\
\midrule
\multirow{6}{*}{LLM arm, digest: purity + silhouette} & \multirow{3}{*}{Qwen} & 0 & 17 & 14 & 3 & 0 & 0 & 1.00 \\
 &  & 1 & 19 & 18 & 1 & 0 & 0 & 1.00 \\
 &  & 2 & 18 & 16 & 2 & 0 & 0 & 1.00 \\
\cmidrule(l){2-9}
 & \multirow{3}{*}{SmolLM2} & 0 & 15 & 9 & 6 & 0 & 0 & 1.00 \\
 &  & 1 & 12 & 6 & 6 & 0 & 0 & 1.00 \\
 &  & 2 & 16 & 15 & 1 & 0 & 0 & 1.00 \\
\midrule
\multirow{6}{*}{LLM arm, digest: + anisotropy} & \multirow{3}{*}{Qwen} & 0 & 12 & 11 & 1 & 0 & 0 & 1.00 \\
 &  & 1 & 22 & 15 & 7 & 0 & 0 & 1.00 \\
 &  & 2 & 12 & 10 & 2 & 0 & 0 & 1.00 \\
\cmidrule(l){2-9}
 & \multirow{3}{*}{SmolLM2} & 0 & 12 & 5 & 7 & 0 & 0 & 1.00 \\
 &  & 1 & 16 & 8 & 8 & 0 & 0 & 1.00 \\
 &  & 2 & 15 & 10 & 5 & 0 & 0 & 1.00 \\
\midrule
\multirow{6}{*}{LLM arm, digest: + ID, Betti} & \multirow{3}{*}{Qwen} & 0 & 17 & 11 & 6 & 0 & 0 & 1.00 \\
 &  & 1 & 22 & 19 & 3 & 0 & 0 & 1.00 \\
 &  & 2 & 23 & 21 & 2 & 0 & 0 & 1.00 \\
\cmidrule(l){2-9}
 & \multirow{3}{*}{SmolLM2} & 0 & 16 & 13 & 3 & 0 & 0 & 1.00 \\
 &  & 1 & 15 & 11 & 4 & 0 & 0 & 1.00 \\
 &  & 2 & 17 & 14 & 3 & 0 & 0 & 1.00 \\
\bottomrule
\end{tabular}
\end{adjustbox}
\end{table}

\begin{table}[!tbp]
\centering
\caption{Operations in the accepted edits of the sequential default runs of the LLM and hybrid arms, by proposer and stage of acceptance, for every seed. Strict: gain $\ge 0.01$; relaxed: gain $\ge 10^{-6}$. $T_{\mathrm{sep}}$, $T_{\mathrm{pcs}}$: numbers of accepted centroid-axis scalings and principal-component rescalings.}
\label{tab:ops2}
\scriptsize
\begin{tabular}{lllrrrr}
\toprule
Arm & Target & Proposer and stage & Seed & Edits & $T_{\mathrm{sep}}$ & $T_{\mathrm{pcs}}$ \\
\midrule
\multirow{12}{*}{LLM arm} & \multirow{6}{*}{Qwen} & \multirow{3}{*}{analyst, strict} & 0 & 3 & 1 & 2 \\
 &  &  & 1 & 7 & 0 & 7 \\
 &  &  & 2 & 8 & 0 & 8 \\
\cmidrule(l){3-7}
 &  & \multirow{3}{*}{analyst, relaxed} & 0 & 10 & 10 & 0 \\
 &  &  & 1 & 13 & 13 & 0 \\
 &  &  & 2 & 15 & 13 & 2 \\
\cmidrule(l){2-7}
 & \multirow{6}{*}{SmolLM2} & \multirow{3}{*}{analyst, strict} & 0 & 5 & 0 & 5 \\
 &  &  & 1 & 3 & 1 & 2 \\
 &  &  & 2 & 4 & 2 & 2 \\
\cmidrule(l){3-7}
 &  & \multirow{3}{*}{analyst, relaxed} & 0 & 8 & 6 & 2 \\
 &  &  & 1 & 12 & 10 & 2 \\
 &  &  & 2 & 12 & 9 & 3 \\
\midrule
\multirow{24}{*}{Hybrid} & \multirow{12}{*}{Qwen} & \multirow{3}{*}{analyst, strict} & 0 & 2 & 2 & 0 \\
 &  &  & 1 & 5 & 2 & 3 \\
 &  &  & 2 & 1 & 1 & 0 \\
\cmidrule(l){3-7}
 &  & \multirow{3}{*}{analyst, relaxed} & 0 & 8 & 7 & 1 \\
 &  &  & 1 & 13 & 14 & 1 \\
 &  &  & 2 & 16 & 16 & 6 \\
\cmidrule(l){3-7}
 &  & \multirow{3}{*}{heuristic, strict} & 0 & 7 & 20 & 0 \\
 &  &  & 1 & 3 & 9 & 0 \\
 &  &  & 2 & 6 & 18 & 0 \\
\cmidrule(l){3-7}
 &  & \multirow{3}{*}{heuristic, relaxed} & 0 & 6 & 16 & 0 \\
 &  &  & 1 & 3 & 6 & 0 \\
 &  &  & 2 & 3 & 8 & 0 \\
\cmidrule(l){2-7}
 & \multirow{12}{*}{SmolLM2} & \multirow{3}{*}{analyst, strict} & 0 & 1 & 1 & 0 \\
 &  &  & 1 & 2 & 0 & 2 \\
 &  &  & 2 & 1 & 1 & 0 \\
\cmidrule(l){3-7}
 &  & \multirow{3}{*}{analyst, relaxed} & 0 & 6 & 8 & 0 \\
 &  &  & 1 & 7 & 7 & 0 \\
 &  &  & 2 & 9 & 8 & 1 \\
\cmidrule(l){3-7}
 &  & \multirow{3}{*}{heuristic, strict} & 0 & 7 & 21 & 0 \\
 &  &  & 1 & 6 & 16 & 0 \\
 &  &  & 2 & 7 & 19 & 0 \\
\cmidrule(l){3-7}
 &  & \multirow{3}{*}{heuristic, relaxed} & 0 & 5 & 12 & 0 \\
 &  &  & 1 & 3 & 7 & 0 \\
 &  &  & 2 & 3 & 9 & 0 \\
\bottomrule
\end{tabular}
\end{table}

In discussing these counts, we note that the main study contains one analyst model (Gemma-2-2B), so the tables compare the LLM arm with the heuristic and the hybrid arm rather than several analysts with one another. In the sequential runs, the analyst's accepted edits contained one operation each; 2 to 10 of 13 to 23 per run (Qwen) and 4 to 7 of 13 to 16 per run (SmolLM2) were $T_{\mathrm{pcs}}$, and the rest $T_{\mathrm{sep}}$. The heuristic used $T_{\mathrm{sep}}$ only, with 2.5 to 2.8 operations per edit. Among the analyst's edits that passed the strict gain threshold, 17 of 18 (Qwen) and 9 of 12 (SmolLM2) were spectral, whereas the edits accepted at the relaxed threshold were mostly centroid-axis scalings (36 of 38 and 25 of 32). All 35 $T_{\mathrm{pcs}}$ edits of the sequential runs amplified leading components (factors above one): 14 scaled the first component alone, 11 the first six, and 10 the second to the sixth. Of the 65 $T_{\mathrm{sep}}$ edits, 34 stretched the tool--vehicle pair and 18 the animal--profession pair. Of the 30 edits accepted at the strict threshold, 8 were accepted at the first proposal, 13 at the second, and 9 at the third. In the parallel runs, 23 to 26 of 27 to 28 (Qwen) and 15 to 18 of 19 to 20 (SmolLM2) of the analyst's edits were spectral, and these runs kept a token-level CKA of 0.900 to 0.923, against 0.483 to 0.577 for the heuristic's centroid-axis scalings (Table~\ref{tab:parallel}).

These patterns are consistent with the hypothesis of Section~\ref{sec:hypothesis} that the higher retention and identity preservation of the analyst's edits may be related to its choice of fewer operations along global spectral directions, which are less tied to the probe vocabulary than centroid-axis scaling. The data do not establish this explanation: the operation type is confounded with the number of operations per edit and with the strength chosen by the adaptive search, the comparison within the main study involves a single analyst, and the only spectral operation that was accepted is $T_{\mathrm{pcs}}$.

\section{Layer Decisions and Per-Edit Statistics}
\label{app:decisions}

Table~\ref{tab:decisions} lists the layer decisions and the per-edit statistics of every default sequential run.

\begin{table}[!tbp]
\centering
\caption{Layer decisions and per-edit statistics of every default sequential run. Dec.: layer decisions; App.: applied edits; Str., Rel.: applied at the strict and at the relaxed gain threshold; No safe: no proposal passed; Rev.: passed the intervention test (stage one) but broke the KL or perplexity limit once baked (stage two) at every tested strength; W: weight budget exceeded; Prop.: proposals tested per layer; Operations: operations per applied edit; $\alpha$: mean strength; $\Delta P$, KL, $\Delta\mathrm{PPL}$ (\%, guard set): mean per applied edit; $R^2$: ridge fit on calibration tokens; $\lVert\Delta\mathbf{W}\rVert$: $\lVert\Delta\mathbf{W}\rVert_F/\lVert\mathbf{W}\rVert_F$; CKA: lowest per-edit CKA on guard tokens.}
\label{tab:decisions}
\begin{adjustbox}{max width=\textwidth}
\scriptsize
\setlength{\tabcolsep}{2.8pt}
\begin{tabular}{lllrrrrrrrrrrrrrrrr}
\toprule
Target & Arm & Seed & Dec. & App. & Str. & Rel. & No safe & Rev. & W & Prop. & Operations & $\alpha$ & $\Delta P$ & KL & $\Delta\mathrm{PPL}$ & $R^2$ & $\lVert\Delta\mathbf{W}\rVert$ & CKA \\
\midrule
\multirow{9}{*}{Qwen} & \multirow{3}{*}{LLM arm} & 0 & 28 & 13 & 3 & 10 & 15 & 0 & 0 & 3.46 & 1.00 & 1.16 & 0.0034 & 0.0049 & 0.38 & 0.999 & 0.036 & 0.983 \\
 &  & 1 & 28 & 20 & 7 & 13 & 8 & 0 & 0 & 3.46 & 1.00 & 1.20 & 0.0026 & 0.0013 & 0.21 & 0.998 & 0.030 & 0.993 \\
 &  & 2 & 28 & 23 & 8 & 15 & 5 & 0 & 0 & 3.43 & 1.00 & 1.15 & 0.0015 & 0.0013 & 0.21 & 0.999 & 0.031 & 0.992 \\
\cmidrule(l){2-19}
 & \multirow{3}{*}{Heuristic} & 0 & 28 & 23 & 6 & 17 & 5 & 0 & 0 & 3.21 & 2.57 & 1.07 & 0.0024 & 0.0017 & 0.18 & 0.999 & 0.024 & 0.996 \\
 &  & 1 & 28 & 21 & 6 & 15 & 6 & 1 & 0 & 3.14 & 2.62 & 1.10 & 0.0036 & 0.0023 & 0.22 & 1.000 & 0.022 & 0.995 \\
 &  & 2 & 28 & 20 & 5 & 15 & 8 & 0 & 0 & 3.18 & 2.50 & 1.18 & 0.0039 & 0.0023 & 0.16 & 0.999 & 0.027 & 0.994 \\
\cmidrule(l){2-19}
 & \multirow{3}{*}{Hybrid} & 0 & 28 & 23 & 9 & 14 & 5 & 0 & 0 & 5.79 & 2.00 & 1.04 & 0.0029 & 0.0020 & 0.11 & 0.999 & 0.023 & 0.994 \\
 &  & 1 & 28 & 24 & 8 & 16 & 4 & 0 & 0 & 5.75 & 1.46 & 1.17 & 0.0032 & 0.0022 & 0.20 & 0.999 & 0.035 & 0.992 \\
 &  & 2 & 28 & 26 & 7 & 19 & 1 & 1 & 0 & 6.36 & 1.88 & 1.17 & 0.0033 & 0.0019 & 0.18 & 0.999 & 0.025 & 0.994 \\
\midrule
\multirow{9}{*}{SmolLM2} & \multirow{3}{*}{LLM arm} & 0 & 24 & 13 & 5 & 8 & 9 & 2 & 0 & 3.46 & 1.00 & 1.07 & 0.0002 & 0.0052 & 0.36 & 0.968 & 0.036 & 0.989 \\
 &  & 1 & 24 & 15 & 3 & 12 & 8 & 1 & 0 & 3.96 & 1.00 & 1.18 & 0.0011 & 0.0069 & 0.33 & 0.955 & 0.034 & 0.988 \\
 &  & 2 & 24 & 16 & 4 & 12 & 7 & 1 & 0 & 3.88 & 1.00 & 1.14 & 0.0018 & 0.0059 & 0.30 & 0.958 & 0.027 & 0.992 \\
\cmidrule(l){2-19}
 & \multirow{3}{*}{Heuristic} & 0 & 24 & 15 & 9 & 6 & 6 & 1 & 2 & 2.79 & 2.73 & 0.89 & 0.0022 & 0.0056 & 0.32 & 0.952 & 0.048 & 0.993 \\
 &  & 1 & 24 & 19 & 6 & 13 & 4 & 1 & 0 & 3.12 & 2.53 & 0.86 & 0.0022 & 0.0046 & 0.25 & 0.963 & 0.034 & 0.994 \\
 &  & 2 & 24 & 17 & 7 & 10 & 5 & 2 & 0 & 2.88 & 2.76 & 0.79 & 0.0023 & 0.0048 & 0.29 & 0.975 & 0.035 & 0.990 \\
\cmidrule(l){2-19}
 & \multirow{3}{*}{Hybrid} & 0 & 24 & 19 & 8 & 11 & 2 & 2 & 1 & 5.54 & 2.21 & 0.88 & 0.0019 & 0.0043 & 0.25 & 0.974 & 0.034 & 0.991 \\
 &  & 1 & 24 & 18 & 8 & 10 & 5 & 1 & 0 & 6.42 & 1.78 & 0.89 & 0.0025 & 0.0055 & 0.26 & 0.957 & 0.034 & 0.994 \\
 &  & 2 & 24 & 20 & 8 & 12 & 1 & 3 & 0 & 5.71 & 1.90 & 0.94 & 0.0017 & 0.0039 & 0.19 & 0.963 & 0.037 & 0.994 \\
\bottomrule
\end{tabular}
\end{adjustbox}
\end{table}

Stage two rejected edits that stage one had accepted: in every SmolLM2 run, 1 to 3 layer decisions ended with reverted weights (1 to 2 per run for the analyst and the heuristic, 1 to 3 for the hybrid arm), as did one decision in each of two Qwen runs (heuristic, seed 1; hybrid, seed 2), so an intervention test alone would have accepted these edits. The relaxed stage supplied 10 to 15 of the analyst's edits per run on Qwen and 8 to 12 on SmolLM2.

\section{Per-Run Results}
\label{app:runs}

Table~\ref{tab:runs} lists every individual run of the default configuration and of the parallel control.

\begin{table}[!tbp]
\centering
\caption{Individual runs of the default configuration (18 sequential runs) and of the parallel control (12 runs). Ed.: edited layers; $\Delta P$ under the four conditions and $\Delta S$ under $[D]$; CKA$_w$ and CKA$_t$: CKA on unseen words and on guard-set tokens; KL: cumulative KL (nats per token); $\Delta\mathrm{PPL}$: held-out perplexity change (\%); $\Delta\mathrm{HS}$, $\Delta\mathrm{ARC}$, $\Delta\mathrm{TQA}$: accuracy changes on HellaSwag, ARC-Easy, and TruthfulQA; $\Delta\mathrm{TM}$: truth-margin change (nats).}
\label{tab:runs}
\begin{adjustbox}{max width=\textwidth}
\scriptsize
\setlength{\tabcolsep}{3pt}
\begin{tabular}{lllrrrrrrrrrrrrrr}
\toprule
Target, mode & Arm & Seed & Ed. & $\Delta P^{[A]}$ & $\Delta P^{[B]}$ & $\Delta P^{[C]}$ & $\Delta P^{[D]}$ & $\Delta S^{[D]}$ & CKA$_w$ & CKA$_t$ & KL & $\Delta\mathrm{PPL}$ & $\Delta\mathrm{HS}$ & $\Delta\mathrm{ARC}$ & $\Delta\mathrm{TQA}$ & $\Delta\mathrm{TM}$ \\
\midrule
\multirow{9}{*}{Qwen, seq.} & \multirow{3}{*}{LLM arm} & 0 & 13 & 0.0239 & 0.0349 & 0.0206 & 0.0287 & 0.0235 & 0.897 & 0.975 & 0.089 & 6.45 & 0.010 & $-$0.030 & 0.000 & $-$0.256 \\
 &  & 1 & 20 & 0.0269 & 0.0224 & 0.0216 & 0.0196 & 0.0484 & 0.935 & 0.981 & 0.086 & 6.96 & 0.005 & 0.015 & $-$0.010 & $-$0.412 \\
 &  & 2 & 23 & 0.0156 & 0.0333 & 0.0117 & 0.0264 & 0.0453 & 0.919 & 0.962 & 0.104 & 8.72 & 0.010 & $-$0.005 & $-$0.005 & $-$0.563 \\
\cmidrule(l){2-17}
 & \multirow{3}{*}{Heuristic} & 0 & 23 & 0.0344 & 0.0373 & 0.0262 & 0.0285 & 0.0158 & 0.837 & 0.982 & 0.115 & 8.14 & 0.010 & $-$0.010 & $-$0.005 & $-$0.198 \\
 &  & 1 & 21 & 0.0315 & 0.0273 & 0.0230 & 0.0235 & 0.0153 & 0.849 & 0.982 & 0.110 & 8.05 & $-$0.005 & 0.005 & $-$0.010 & $-$0.237 \\
 &  & 2 & 20 & 0.0360 & 0.0259 & 0.0290 & 0.0162 & 0.0110 & 0.812 & 0.981 & 0.108 & 8.26 & $-$0.005 & $-$0.015 & $-$0.005 & $-$0.267 \\
\cmidrule(l){2-17}
 & \multirow{3}{*}{Hybrid} & 0 & 23 & 0.0316 & 0.0313 & 0.0244 & 0.0232 & 0.0115 & 0.835 & 0.982 & 0.114 & 7.28 & 0.005 & $-$0.020 & $-$0.020 & $-$0.252 \\
 &  & 1 & 24 & 0.0389 & 0.0320 & 0.0286 & 0.0263 & 0.0269 & 0.861 & 0.979 & 0.111 & 8.20 & 0.000 & 0.000 & $-$0.015 & $-$0.311 \\
 &  & 2 & 26 & 0.0340 & 0.0269 & 0.0280 & 0.0169 & 0.0125 & 0.857 & 0.982 & 0.102 & 8.81 & 0.000 & $-$0.015 & 0.000 & $-$0.130 \\
\midrule
\multirow{9}{*}{SmolLM2, seq.} & \multirow{3}{*}{LLM arm} & 0 & 13 & 0.0017 & 0.0145 & 0.0025 & 0.0111 & 0.0266 & 0.902 & 0.984 & 0.092 & 6.82 & 0.005 & 0.005 & 0.000 & $-$0.563 \\
 &  & 1 & 15 & 0.0067 & 0.0137 & 0.0071 & 0.0104 & 0.0166 & 0.962 & 0.988 & 0.083 & 6.50 & $-$0.010 & $-$0.015 & 0.000 & $-$0.083 \\
 &  & 2 & 16 & 0.0095 & 0.0097 & 0.0096 & 0.0117 & 0.0163 & 0.953 & 0.989 & 0.099 & 8.17 & 0.000 & 0.005 & $-$0.005 & $-$0.020 \\
\cmidrule(l){2-17}
 & \multirow{3}{*}{Heuristic} & 0 & 15 & 0.0097 & 0.0147 & 0.0108 & 0.0150 & 0.0173 & 0.898 & 0.977 & 0.121 & 4.71 & $-$0.040 & $-$0.015 & 0.000 & $-$0.474 \\
 &  & 1 & 19 & 0.0131 & 0.0179 & 0.0144 & 0.0139 & 0.0203 & 0.885 & 0.942 & 0.112 & 6.04 & $-$0.020 & $-$0.040 & 0.015 & $-$0.573 \\
 &  & 2 & 17 & 0.0064 & 0.0096 & 0.0103 & 0.0087 & 0.0219 & 0.853 & 0.964 & 0.119 & 9.22 & 0.000 & 0.000 & 0.010 & $-$0.431 \\
\cmidrule(l){2-17}
 & \multirow{3}{*}{Hybrid} & 0 & 19 & 0.0100 & 0.0183 & 0.0119 & 0.0147 & 0.0191 & 0.892 & 0.978 & 0.115 & 5.47 & $-$0.030 & $-$0.010 & $-$0.005 & $-$0.410 \\
 &  & 1 & 18 & 0.0128 & 0.0173 & 0.0135 & 0.0143 & 0.0208 & 0.919 & 0.940 & 0.112 & 5.63 & $-$0.020 & $-$0.035 & 0.015 & $-$0.396 \\
 &  & 2 & 20 & 0.0044 & 0.0045 & 0.0078 & 0.0070 & 0.0175 & 0.874 & 0.972 & 0.102 & 9.25 & 0.010 & 0.010 & 0.015 & $-$0.240 \\
\midrule
\multirow{6}{*}{Qwen, par.} & \multirow{3}{*}{LLM arm} & 0 & 28 & 0.0172 & 0.0375 & 0.0123 & 0.0285 & 0.0771 & 0.890 & 0.921 & 0.702 & 111.60 & $-$0.020 & $-$0.060 & 0.010 & $-$0.798 \\
 &  & 1 & 28 & 0.0219 & 0.0189 & 0.0153 & 0.0151 & 0.0977 & 0.897 & 0.923 & 0.696 & 103.61 & 0.005 & $-$0.025 & $-$0.005 & $-$1.013 \\
 &  & 2 & 27 & 0.0093 & 0.0318 & 0.0055 & 0.0268 & 0.0899 & 0.890 & 0.909 & 0.801 & 134.40 & 0.005 & 0.000 & $-$0.005 & $-$1.194 \\
\cmidrule(l){2-17}
 & \multirow{3}{*}{Heuristic} & 0 & 28 & $-$0.0439 & $-$0.0665 & $-$0.0544 & $-$0.0837 & $-$0.1220 & 0.394 & 0.492 & 5.559 & 19{,}045 & $-$0.230 & $-$0.430 & 0.040 & $-$4.002 \\
 &  & 1 & 28 & $-$0.0380 & $-$0.0940 & $-$0.0482 & $-$0.1064 & $-$0.1368 & 0.392 & 0.491 & 4.965 & 12{,}334 & $-$0.235 & $-$0.395 & 0.090 & $-$4.396 \\
 &  & 2 & 28 & $-$0.0419 & $-$0.0849 & $-$0.0543 & $-$0.1070 & $-$0.1066 & 0.394 & 0.483 & 5.241 & 15{,}423 & $-$0.245 & $-$0.410 & 0.085 & $-$4.025 \\
\midrule
\multirow{6}{*}{SmolLM2, par.} & \multirow{3}{*}{LLM arm} & 0 & 19 & 0.0069 & 0.0109 & 0.0080 & 0.0110 & 0.0483 & 0.919 & 0.908 & 1.136 & 165.79 & $-$0.090 & $-$0.050 & $-$0.030 & $-$1.660 \\
 &  & 1 & 20 & 0.0017 & 0.0107 & 0.0031 & 0.0035 & 0.0698 & 0.926 & 0.902 & 1.205 & 179.33 & $-$0.075 & $-$0.110 & 0.010 & $-$2.198 \\
 &  & 2 & 20 & 0.0047 & 0.0074 & 0.0042 & 0.0102 & 0.0571 & 0.923 & 0.900 & 1.234 & 185.10 & $-$0.065 & $-$0.050 & $-$0.025 & $-$1.765 \\
\cmidrule(l){2-17}
 & \multirow{3}{*}{Heuristic} & 0 & 20 & $-$0.0087 & $-$0.0356 & $-$0.0109 & $-$0.0366 & $-$0.1023 & 0.500 & 0.577 & 3.974 & 4{,}367 & $-$0.210 & $-$0.105 & 0.105 & $-$5.266 \\
 &  & 1 & 20 & $-$0.0157 & $-$0.0374 & $-$0.0151 & $-$0.0403 & $-$0.0770 & 0.529 & 0.502 & 4.091 & 4{,}824 & $-$0.260 & $-$0.170 & 0.130 & $-$5.690 \\
 &  & 2 & 20 & $-$0.0223 & $-$0.0514 & $-$0.0230 & $-$0.0586 & $-$0.0685 & 0.503 & 0.553 & 4.013 & 4{,}905 & $-$0.200 & $-$0.185 & 0.075 & $-$5.330 \\
\bottomrule
\end{tabular}
\end{adjustbox}
\end{table}

\section{Statistical Tests}
\label{app:stats}

The tests below are computed from the per-run values of Tables~\ref{tab:main}, \ref{tab:parallel}, and~\ref{tab:runs} and are reported separately for each arm; no test combines runs of different arms. Table~\ref{tab:pair} gives the tests per target ($n=3$ seeds), Table~\ref{tab:stats} the tests per arm over both targets ($n=6$ runs) with the paired comparison against the LLM arm, and Table~\ref{tab:nonseq} the same for the parallel control. The tests over $n=6$ runs are descriptive summaries: the six runs of an arm form a clustered design of two targets $\times$ three data splits on the same frozen weights, not six independent samples, so the per-target tables (Tables~\ref{tab:main}, \ref{tab:parallel}, and~\ref{tab:pair}) are the primary evidence.

\begin{table}[!tbp]
\centering
\caption{Tests per target and arm over the three seeds of the default sequential runs: mean (standard deviation) and the $p$-value of a one-sample $t$-test against zero ($n=3$).}
\label{tab:pair}
\begin{adjustbox}{max width=\textwidth}
\footnotesize
\setlength{\tabcolsep}{3.5pt}
\begin{tabular}{llcccccccc}
\toprule
Target & Arm & $\Delta P^{[A]}$ & $p$ & $\Delta P^{[D]}$ & $p$ & $\Delta S^{[D]}$ & $p$ & KL & Held-out perplexity change (\%) \\
\midrule
Qwen & LLM arm & 0.0221 (0.0059) & 0.023 & 0.0249 (0.0047) & 0.012 & 0.0390 (0.0136) & 0.038 & 0.093 (0.009) & 7.38 (1.19) \\
Qwen & Heuristic & 0.0340 (0.0023) & 0.001 & 0.0227 (0.0062) & 0.024 & 0.0140 (0.0026) & 0.012 & 0.111 (0.003) & 8.15 (0.11) \\
Qwen & Hybrid & 0.0348 (0.0038) & 0.004 & 0.0222 (0.0048) & 0.015 & 0.0170 (0.0086) & 0.076 & 0.109 (0.007) & 8.10 (0.77) \\
\addlinespace
SmolLM2 & LLM arm & 0.0060 (0.0040) & 0.121 & 0.0111 (0.0006) & 0.001 & 0.0198 (0.0059) & 0.028 & 0.091 (0.008) & 7.16 (0.89) \\
SmolLM2 & Heuristic & 0.0097 (0.0034) & 0.038 & 0.0125 (0.0034) & 0.024 & 0.0198 (0.0023) & 0.004 & 0.117 (0.004) & 6.66 (2.31) \\
SmolLM2 & Hybrid & 0.0090 (0.0043) & 0.068 & 0.0120 (0.0043) & 0.041 & 0.0191 (0.0016) & 0.002 & 0.110 (0.007) & 6.78 (2.14) \\
\bottomrule
\end{tabular}
\end{adjustbox}
\end{table}

\begin{table}[!tbp]
\centering
\caption{Tests per arm over both targets and three seeds ($n=6$ runs per arm; descriptive summary over a clustered design of 2 targets $\times$ 3 data splits). Mean, 95\% $t$-interval, $p$-value of a one-sample $t$-test against zero, and the paired difference to the LLM arm (arm minus LLM arm) with the $p$-value of a paired $t$-test over runs matched on target and seed. $p$-values are uncorrected; a test against zero is not meaningful for CKA.}
\label{tab:stats}
\begin{adjustbox}{max width=\textwidth, max totalheight=0.88\textheight}
\footnotesize
\begin{tabular}{llccccc}
\toprule
Metric & Arm & Mean & 95\% interval & $p$ (vs. 0) & Diff. to LLM arm & $p$ (paired) \\
\midrule
$\Delta P$ $[A]$ & LLM arm & 0.0140 & [0.0036, 0.0245] & 0.018 & -- & -- \\
 & Heuristic & 0.0218 & [0.0076, 0.0360] & 0.011 & $+$0.0078 & 0.056 \\
 & Hybrid & 0.0219 & [0.0066, 0.0372] & 0.014 & $+$0.0079 & 0.055 \\
\addlinespace
$\Delta P$ $[B]$ & LLM arm & 0.0214 & [0.0102, 0.0326] & 0.004 & -- & -- \\
 & Heuristic & 0.0221 & [0.0116, 0.0326] & 0.003 & $+$0.0007 & 0.715 \\
 & Hybrid & 0.0217 & [0.0107, 0.0327] & 0.004 & $+$0.0003 & 0.907 \\
\addlinespace
$\Delta P$ $[C]$ & LLM arm & 0.0122 & [0.0042, 0.0201] & 0.011 & -- & -- \\
 & Heuristic & 0.0189 & [0.0104, 0.0275] & 0.002 & $+$0.0068 & 0.039 \\
 & Hybrid & 0.0190 & [0.0096, 0.0285] & 0.004 & $+$0.0069 & 0.037 \\
\addlinespace
$\Delta P$ $[D]$ & LLM arm & 0.0180 & [0.0094, 0.0266] & 0.003 & -- & -- \\
 & Heuristic & 0.0176 & [0.0101, 0.0251] & 0.002 & $-$0.0004 & 0.882 \\
 & Hybrid & 0.0171 & [0.0098, 0.0243] & 0.002 & $-$0.0009 & 0.743 \\
\addlinespace
$\Delta S$ $[D]$ & LLM arm & 0.0294 & [0.0147, 0.0442] & 0.004 & -- & -- \\
 & Heuristic & 0.0169 & [0.0129, 0.0210] & $<$0.001 & $-$0.0125 & 0.140 \\
 & Hybrid & 0.0180 & [0.0121, 0.0240] & $<$0.001 & $-$0.0114 & 0.102 \\
\addlinespace
$\Delta\kappa$ $[D]$ & LLM arm & $-$0.0315 & [$-$0.0540, $-$0.0091] & 0.015 & -- & -- \\
 & Heuristic & $-$0.0238 & [$-$0.0262, $-$0.0214] & $<$0.001 & $+$0.0077 & 0.411 \\
 & Hybrid & $-$0.0245 & [$-$0.0320, $-$0.0169] & $<$0.001 & $+$0.0071 & 0.399 \\
\addlinespace
CKA, probe words & LLM arm & 0.919 & [0.886, 0.953] & -- & -- & -- \\
 & Heuristic & 0.829 & [0.788, 0.871] & -- & $-$0.090 & 0.003 \\
 & Hybrid & 0.848 & [0.808, 0.889] & -- & $-$0.071 & $<$0.001 \\
\addlinespace
CKA, unseen words & LLM arm & 0.928 & [0.900, 0.956] & -- & -- & -- \\
 & Heuristic & 0.856 & [0.823, 0.889] & -- & $-$0.072 & 0.005 \\
 & Hybrid & 0.873 & [0.842, 0.904] & -- & $-$0.055 & 0.003 \\
\addlinespace
CKA, guard tokens & LLM arm & 0.980 & [0.969, 0.990] & -- & -- & -- \\
 & Heuristic & 0.972 & [0.955, 0.988] & -- & $-$0.008 & 0.430 \\
 & Hybrid & 0.972 & [0.955, 0.989] & -- & $-$0.008 & 0.452 \\
\addlinespace
Cumulative KL & LLM arm & 0.092 & [0.084, 0.100] & $<$0.001 & -- & -- \\
 & Heuristic & 0.114 & [0.109, 0.119] & $<$0.001 & $+$0.022 & 0.002 \\
 & Hybrid & 0.109 & [0.103, 0.116] & $<$0.001 & $+$0.017 & 0.025 \\
\addlinespace
Held-out perplexity change (\%) & LLM arm & 7.27 & [6.28, 8.26] & $<$0.001 & -- & -- \\
 & Heuristic & 7.40 & [5.64, 9.16] & $<$0.001 & $+$0.13 & 0.827 \\
 & Hybrid & 7.44 & [5.75, 9.13] & $<$0.001 & $+$0.17 & 0.719 \\
\addlinespace
$\Delta$ cloze, unseen & LLM arm & 0.022 & [0.002, 0.041] & 0.037 & -- & -- \\
 & Heuristic & 0.035 & [0.020, 0.051] & 0.002 & $+$0.014 & 0.298 \\
 & Hybrid & 0.027 & [0.011, 0.044] & 0.007 & $+$0.006 & 0.571 \\
\addlinespace
$\Delta$ HellaSwag & LLM arm & 0.003 & [$-$0.005, 0.011] & 0.328 & -- & -- \\
 & Heuristic & $-$0.010 & [$-$0.028, 0.008] & 0.223 & $-$0.013 & 0.107 \\
 & Hybrid & $-$0.006 & [$-$0.022, 0.011] & 0.402 & $-$0.009 & 0.186 \\
\addlinespace
$\Delta$ ARC-Easy & LLM arm & $-$0.004 & [$-$0.021, 0.013] & 0.558 & -- & -- \\
 & Heuristic & $-$0.012 & [$-$0.029, 0.004] & 0.109 & $-$0.008 & 0.250 \\
 & Hybrid & $-$0.012 & [$-$0.028, 0.005] & 0.128 & $-$0.008 & 0.191 \\
\addlinespace
$\Delta$ TruthfulQA & LLM arm & $-$0.003 & [$-$0.008, 0.001] & 0.102 & -- & -- \\
 & Heuristic & 0.001 & [$-$0.009, 0.011] & 0.842 & $+$0.004 & 0.289 \\
 & Hybrid & $-$0.002 & [$-$0.017, 0.014] & 0.793 & $+$0.002 & 0.793 \\
\addlinespace
$\Delta$ truth margin & LLM arm & $-$0.316 & [$-$0.563, $-$0.069] & 0.022 & -- & -- \\
 & Heuristic & $-$0.363 & [$-$0.521, $-$0.205] & 0.002 & $-$0.047 & 0.735 \\
 & Hybrid & $-$0.290 & [$-$0.401, $-$0.179] & 0.001 & $+$0.026 & 0.822 \\
\bottomrule
\end{tabular}
\end{adjustbox}
\end{table}

\begin{table}[!tbp]
\centering
\caption{Tests of the parallel control per arm over both targets and three seeds ($n=6$ runs per cell; descriptive summary over a clustered design of 2 targets $\times$ 3 data splits). The paired difference compares the parallel with the sequential run of the same arm, target, and seed.}
\label{tab:nonseq}
\begin{adjustbox}{max width=\textwidth}
\footnotesize
\begin{tabular}{lllcccc}
\toprule
Metric & Arm & Mode & Mean & 95\% interval & Diff. to sequential & $p$ (paired) \\
\midrule
Edited layers & LLM arm & Sequential & 16.7 & [12.4, 20.9] & -- & -- \\
 & LLM arm & Parallel & 23.7 & [19.0, 28.3] & $+$7.0 & 0.009 \\
 & Heuristic & Sequential & 19.2 & [16.2, 22.2] & -- & -- \\
 & Heuristic & Parallel & 24.0 & [19.4, 28.6] & $+$4.8 & 0.006 \\
\addlinespace
$\Delta P$ $[D]$ & LLM arm & Sequential & 0.0180 & [0.0094, 0.0266] & -- & -- \\
 & LLM arm & Parallel & 0.0159 & [0.0055, 0.0262] & $-$0.0021 & 0.137 \\
 & Heuristic & Sequential & 0.0176 & [0.0101, 0.0251] & -- & -- \\
 & Heuristic & Parallel & $-$0.0721 & [$-$0.1052, $-$0.0390] & $-$0.0897 & 0.002 \\
\addlinespace
$\Delta S$ $[D]$ & LLM arm & Sequential & 0.0294 & [0.0147, 0.0442] & -- & -- \\
 & LLM arm & Parallel & 0.0733 & [0.0535, 0.0932] & $+$0.0439 & $<$0.001 \\
 & Heuristic & Sequential & 0.0169 & [0.0129, 0.0210] & -- & -- \\
 & Heuristic & Parallel & $-$0.1022 & [$-$0.1295, $-$0.0749] & $-$0.1191 & $<$0.001 \\
\addlinespace
CKA, guard tokens & LLM arm & Sequential & 0.980 & [0.969, 0.990] & -- & -- \\
 & LLM arm & Parallel & 0.910 & [0.900, 0.920] & $-$0.069 & $<$0.001 \\
 & Heuristic & Sequential & 0.972 & [0.955, 0.988] & -- & -- \\
 & Heuristic & Parallel & 0.516 & [0.475, 0.557] & $-$0.455 & $<$0.001 \\
\addlinespace
Cumulative KL & LLM arm & Sequential & 0.092 & [0.084, 0.100] & -- & -- \\
 & LLM arm & Parallel & 0.962 & [0.694, 1.231] & $+$0.870 & $<$0.001 \\
 & Heuristic & Sequential & 0.114 & [0.109, 0.119] & -- & -- \\
 & Heuristic & Parallel & 4.640 & [3.906, 5.375] & $+$4.526 & $<$0.001 \\
\addlinespace
Held-out perplexity change (\%) & LLM arm & Sequential & 7.3 & [6.3, 8.3] & -- & -- \\
 & LLM arm & Parallel & 146.6 & [109.9, 183.4] & $+$139.4 & $<$0.001 \\
 & Heuristic & Sequential & 7.4 & [5.6, 9.2] & -- & -- \\
 & Heuristic & Parallel & 10{,}150 & [3{,}496, 16{,}804] & $+$10{,}142 & 0.011 \\
\addlinespace
$\Delta$ HellaSwag & LLM arm & Sequential & 0.003 & [$-$0.005, 0.011] & -- & -- \\
 & LLM arm & Parallel & $-$0.040 & [$-$0.084, 0.004] & $-$0.043 & 0.038 \\
 & Heuristic & Sequential & $-$0.010 & [$-$0.028, 0.008] & -- & -- \\
 & Heuristic & Parallel & $-$0.230 & [$-$0.253, $-$0.207] & $-$0.220 & $<$0.001 \\
\addlinespace
$\Delta$ cloze, unseen & LLM arm & Sequential & 0.022 & [0.002, 0.041] & -- & -- \\
 & LLM arm & Parallel & $-$0.050 & [$-$0.087, $-$0.012] & $-$0.071 & 0.010 \\
 & Heuristic & Sequential & 0.035 & [0.020, 0.051] & -- & -- \\
 & Heuristic & Parallel & $-$0.320 & [$-$0.462, $-$0.178] & $-$0.355 & $<$0.001 \\
\bottomrule
\end{tabular}
\end{adjustbox}
\end{table}

\section{Full Ablation Tables}
\label{app:ablations}

Tables~\ref{tab:abldesc} to~\ref{tab:ablcatrule} list every run of every ablation, and Table~\ref{tab:pooled} gives the tests of the purity gain under $[D]$ for every variant. The strength and gain-threshold studies were run on the heuristic arm; they concern the verification loop and the heuristic, not the analyst.

\subsection{Extended results}
\label{app:extended}

Under $[D]$, the differences in purity gain between the LLM arm and the default heuristic on the same target and seed ranged from $-0.004$ to $+0.010$. In all six pairs, the analyst's edits had a lower cumulative KL (by 0.005 to 0.030 nats per token) and a higher CKA on unseen words (by 0.004 to 0.107) and on probe words (by 0.021 to 0.133). On Qwen, the analyst's silhouette gain was higher in all three pairs (by 0.008 to 0.034); on SmolLM2, the heuristic's was higher in two of three. In the hybrid arm, analyst edits could contain several operations (for example, 16 relaxed-stage analyst edits with 22 operations on Qwen, seed 2; Table~\ref{tab:ops2}).

In addition, per nat of cumulative KL, the LLM arm reached a purity gain under $[D]$ of 0.23 to 0.32 (Qwen) and 0.12 to 0.13 (SmolLM2), against 0.15 to 0.25 and 0.07 to 0.12 for the heuristic.

Next, the hybrid arm's purity gain under $[D]$ was 0.0169 to 0.0263 (Qwen) and 0.0070 to 0.0147 (SmolLM2), its CKA on unseen words 0.835 to 0.861 and 0.874 to 0.919, its cumulative KL 0.102 to 0.114 and 0.102 to 0.115, and its held-out perplexity change $+7.3\%$ to $+8.8\%$ and $+5.5\%$ to $+9.2\%$. Its CKA and KL lay between the other two arms or overlapped with the heuristic, and its largest benchmark change was $-0.035$.

Beyond the edited layers, under the all-layer averages of Table~\ref{tab:traj}, the ranking between the LLM arm and the default heuristic under $[D]$ is unchanged in all six seed-matched pairs. Across all pairwise comparisons of the three arms, the ranking changes in one case: on SmolLM2 with seed 0, the hybrid arm leads the LLM arm over the edited layers (0.0147 against 0.0111), whereas the LLM arm leads over all layers (0.0131 against 0.0129). On Qwen, the LLM arm's all-layer gain (0.0216 to 0.0357) exceeds its edited-layer gain in all three runs, because its edits also raise unedited downstream layers.

In the parallel control, for the LLM arm, the parallel mode gave a cumulative KL of 0.696 to 0.801 nats per token on Qwen and 1.136 to 1.234 on SmolLM2, and held-out perplexity increases of 103.6\% to 134.4\% and 165.8\% to 185.1\%. For the heuristic, it gave a KL of 4.97 to 5.56 and 3.97 to 4.09, held-out perplexity increases of 12{,}334\% to 19{,}045\% and 4{,}367\% to 4{,}905\%, and HellaSwag drops of 0.23 to 0.25 and 0.20 to 0.26. In the LLM arm's parallel runs, 23 to 26 of 27 to 28 accepted edits (Qwen) and 15 to 18 of 19 to 20 (SmolLM2) used $T_{\mathrm{pcs}}$.

Regarding identity, perplexity, and behavior, the lowest per-edit token CKA was 0.983 for the LLM arm and 0.990 for the heuristic (Table~\ref{tab:decisions}). After all edits, the heuristic's token-level CKA fell to 0.78, 0.38, and 0.55 at layer 8 of SmolLM2 in the three runs, with the same minima for the hybrid arm (0.77, 0.38, and 0.55). The guard-set perplexity rose by 2.6\% to 5.0\% across the 18 default runs (LLM arm: 4.3\% to 5.0\%). Cloze accuracy rose on probe words by 0.041 to 0.084; the one decrease on unseen words was the LLM arm on Qwen, seed 1 ($-0.003$). The truth margin decreased by 0.256 to 0.563 nats (Qwen) and 0.020 to 0.563 nats (SmolLM2) for the LLM arm and by 0.198 to 0.267 and 0.431 to 0.573 nats for the heuristic; the factual-preference accuracy stayed at 0.975 on Qwen and 1.000 on SmolLM2.

Turning to strength and gain threshold, a fixed strength of $\alpha=2$ saturates the operation limits and left 1 to 8 accepted edits (Qwen) and 0 to 8 (SmolLM2). A fixed $\alpha=0.5$ and a strict gain threshold of 0.02 without the relaxed stage each gave a purity gain under $[D]$ in or above the range of the adaptive default on both targets. On Qwen, both reached it at a much lower divergence (KL 0.022 to 0.043 against 0.108 to 0.115); on SmolLM2, the strict threshold lowered the KL (0.030 to 0.076 against 0.112 to 0.121), whereas $\alpha=0.5$ did not (0.089 to 0.130; Table~\ref{tab:headline}).

As for the fidelity of the baked edits, over the 345 applied edits of the 18 default runs, the ridge fit had a median $R^2$ of 0.9995 on the calibration tokens, and 92.5\% of the edits had an $R^2$ of at least 0.99. On Qwen, the $R^2$ of every edit was between 0.991 and 0.9998; on SmolLM2, the mean was 0.963 and the minimum 0.37. The KL divergence between the intervened and the baked model had a median of 0.0001 nats per token over the 345 edits (0.0002 for the analyst's 100 edits, with a maximum of 0.039). Of the analyst's 100 edits, 91 were fitted with the smallest ridge weight ($\lambda=10^{-2}$), 6 with $10^{-1}$, and 3 with $\lambda=1$; their relative weight change was at most 0.096.

Regarding the number of categories, with four categories, the LLM arm's purity gain under $[D]$ was 0.020 to 0.035 (Qwen) and 0.012 to 0.016 (SmolLM2), and the heuristic's 0.045 to 0.082 and 0.024 to 0.050. With 16 categories, the LLM arm's purity gain under $[D]$ was $-0.010$ to $+0.0005$ (Qwen) and $-0.010$ to $-0.005$ (SmolLM2), while its silhouette gain stayed positive (0.019 to 0.025 and 0.005 to 0.014); the heuristic's purity gain was 0.010 to 0.017 and $-0.004$ to $+0.006$.

Finally, as for analyst scale, we conjecture that analysts with stronger reasoning, instruction following, and adherence to the output schema reach higher acceptance rates at the strict gain threshold and better out-of-distribution transfer; the round robin gives a first test (Appendix~\ref{app:rr}). There, the number of accepted edits differed strongly between analysts and was lowest for the two smallest models, but the purity gain did not increase with the parameter count.

\begin{table}[!tbp]
\centering
\caption{Descriptor ablation, LLM arm, every run (LLM arm, sequential; 24 runs of the descriptor study). Ed.: edited layers; CKA$_w$: CKA on unseen words; KL: cumulative KL; $\Delta\mathrm{PPL}$: held-out perplexity change (\%).}
\label{tab:abldesc}
\begin{adjustbox}{max width=\textwidth}
\footnotesize
\begin{tabular}{lllccccccc}
\toprule
Target & Digest & Seed & Ed. & $\Delta P^{[A]}$ & $\Delta P^{[D]}$ & $\Delta S^{[D]}$ & CKA$_w$ & KL & $\Delta\mathrm{PPL}$ \\
\midrule
\multirow{12}{*}{Qwen} & \multirow{3}{*}{Purity + silhouette} & 0 & 17 & 0.0173 & 0.0104 & 0.0159 & 0.889 & 0.106 & 6.01 \\
 &  & 1 & 19 & 0.0249 & 0.0208 & 0.0178 & 0.898 & 0.085 & 6.40 \\
 &  & 2 & 18 & 0.0279 & 0.0147 & 0.0089 & 0.855 & 0.096 & 6.24 \\
\cmidrule(l){2-10}
 & \multirow{3}{*}{+ anisotropy} & 0 & 12 & 0.0128 & 0.0042 & 0.0033 & 0.947 & 0.079 & 5.26 \\
 &  & 1 & 22 & 0.0244 & 0.0159 & 0.0292 & 0.891 & 0.087 & 7.77 \\
 &  & 2 & 12 & 0.0181 & 0.0105 & 0.0147 & 0.917 & 0.088 & 5.87 \\
\cmidrule(l){2-10}
 & \multirow{3}{*}{+ intrinsic dimension, Betti} & 0 & 17 & 0.0206 & 0.0231 & 0.0390 & 0.932 & 0.094 & 7.04 \\
 &  & 1 & 22 & 0.0263 & 0.0270 & 0.0196 & 0.910 & 0.080 & 4.51 \\
 &  & 2 & 23 & 0.0269 & 0.0203 & 0.0136 & 0.864 & 0.097 & 8.65 \\
\cmidrule(l){2-10}
 & \multirow{3}{*}{+ drift, distance ratio (full)} & 0 & 13 & 0.0239 & 0.0287 & 0.0235 & 0.897 & 0.089 & 6.45 \\
 &  & 1 & 20 & 0.0269 & 0.0196 & 0.0484 & 0.935 & 0.086 & 6.96 \\
 &  & 2 & 23 & 0.0156 & 0.0264 & 0.0453 & 0.919 & 0.104 & 8.72 \\
\midrule
\multirow{12}{*}{SmolLM2} & \multirow{3}{*}{Purity + silhouette} & 0 & 15 & 0.0012 & $-$0.0018 & 0.0059 & 0.982 & 0.073 & 5.11 \\
 &  & 1 & 12 & 0.0000 & $-$0.0004 & 0.0273 & 0.880 & 0.133 & 7.28 \\
 &  & 2 & 16 & 0.0017 & $-$0.0026 & 0.0026 & 0.963 & 0.140 & 8.61 \\
\cmidrule(l){2-10}
 & \multirow{3}{*}{+ anisotropy} & 0 & 12 & 0.0035 & 0.0056 & 0.0157 & 0.905 & 0.170 & 10.42 \\
 &  & 1 & 16 & 0.0029 & 0.0067 & 0.0257 & 0.892 & 0.091 & 5.93 \\
 &  & 2 & 15 & 0.0065 & 0.0005 & 0.0128 & 0.910 & 0.141 & 8.58 \\
\cmidrule(l){2-10}
 & \multirow{3}{*}{+ intrinsic dimension, Betti} & 0 & 16 & 0.0067 & 0.0093 & 0.0074 & 0.944 & 0.143 & 7.35 \\
 &  & 1 & 15 & 0.0067 & 0.0088 & 0.0130 & 0.971 & 0.074 & 5.15 \\
 &  & 2 & 17 & 0.0050 & 0.0008 & 0.0099 & 0.886 & 0.127 & 8.31 \\
\cmidrule(l){2-10}
 & \multirow{3}{*}{+ drift, distance ratio (full)} & 0 & 13 & 0.0017 & 0.0111 & 0.0266 & 0.902 & 0.092 & 6.82 \\
 &  & 1 & 15 & 0.0067 & 0.0104 & 0.0166 & 0.962 & 0.083 & 6.50 \\
 &  & 2 & 16 & 0.0095 & 0.0117 & 0.0163 & 0.953 & 0.099 & 8.17 \\
\bottomrule
\end{tabular}
\end{adjustbox}
\end{table}

\begin{table}[!tbp]
\centering
\caption{Strength ablation, heuristic arm, every run (sequential; 42 runs of the strength study). One fixed strength $\alpha$ replaces the adaptive search over six strengths. Factors are clipped to the operation limits, so $\alpha=2$ saturates. Columns as in Table~\ref{tab:abldesc}.}
\label{tab:ablstr}
\begin{adjustbox}{max width=\textwidth, max totalheight=0.88\textheight}
\footnotesize
\begin{tabular}{lllccccccc}
\toprule
Target & Strength & Seed & Ed. & $\Delta P^{[A]}$ & $\Delta P^{[D]}$ & $\Delta S^{[D]}$ & CKA$_w$ & KL & $\Delta\mathrm{PPL}$ \\
\midrule
\multirow{21}{*}{Qwen} & \multirow{3}{*}{$\alpha=0.25$} & 0 & 28 & 0.0159 & 0.0187 & 0.0086 & 0.992 & 0.004 & 0.15 \\
 &  & 1 & 28 & 0.0167 & 0.0170 & 0.0109 & 0.990 & 0.004 & 0.36 \\
 &  & 2 & 28 & 0.0186 & 0.0190 & 0.0108 & 0.990 & 0.005 & 0.16 \\
\cmidrule(l){2-10}
 & \multirow{3}{*}{$\alpha=0.5$} & 0 & 28 & 0.0302 & 0.0293 & 0.0169 & 0.947 & 0.028 & 2.05 \\
 &  & 1 & 28 & 0.0308 & 0.0275 & 0.0210 & 0.947 & 0.022 & 1.85 \\
 &  & 2 & 28 & 0.0306 & 0.0277 & 0.0197 & 0.951 & 0.022 & 1.68 \\
\cmidrule(l){2-10}
 & \multirow{3}{*}{$\alpha=1$} & 0 & 19 & 0.0339 & 0.0306 & 0.0158 & 0.878 & 0.088 & 5.65 \\
 &  & 1 & 20 & 0.0392 & 0.0308 & 0.0221 & 0.863 & 0.099 & 7.94 \\
 &  & 2 & 18 & 0.0356 & 0.0209 & 0.0150 & 0.858 & 0.079 & 5.82 \\
\cmidrule(l){2-10}
 & \multirow{3}{*}{$\alpha=1.5$} & 0 & 11 & 0.0323 & 0.0241 & 0.0122 & 0.873 & 0.109 & 7.02 \\
 &  & 1 & 10 & 0.0308 & 0.0261 & 0.0140 & 0.882 & 0.109 & 6.46 \\
 &  & 2 & 11 & 0.0344 & 0.0224 & 0.0141 & 0.883 & 0.119 & 6.92 \\
\cmidrule(l){2-10}
 & \multirow{3}{*}{$\alpha=1.75$} & 0 & 8 & 0.0183 & 0.0134 & 0.0052 & 0.948 & 0.063 & 4.18 \\
 &  & 1 & 9 & 0.0320 & 0.0265 & 0.0135 & 0.882 & 0.126 & 7.23 \\
 &  & 2 & 7 & 0.0299 & 0.0184 & 0.0076 & 0.910 & 0.119 & 6.45 \\
\cmidrule(l){2-10}
 & \multirow{3}{*}{$\alpha=2$} & 0 & 8 & 0.0176 & 0.0119 & 0.0044 & 0.931 & 0.060 & 3.70 \\
 &  & 1 & 1 & 0.0036 & 0.0035 & 0.0000 & 0.999 & 0.009 & $-$0.05 \\
 &  & 2 & 2 & 0.0083 & 0.0099 & 0.0014 & 0.993 & 0.021 & 1.17 \\
\cmidrule(l){2-10}
 & \multirow{3}{*}{Adaptive} & 0 & 23 & 0.0344 & 0.0285 & 0.0158 & 0.837 & 0.115 & 8.14 \\
 &  & 1 & 21 & 0.0315 & 0.0235 & 0.0153 & 0.849 & 0.110 & 8.05 \\
 &  & 2 & 20 & 0.0360 & 0.0162 & 0.0110 & 0.812 & 0.108 & 8.26 \\
\midrule
\multirow{21}{*}{SmolLM2} & \multirow{3}{*}{$\alpha=0.25$} & 0 & 23 & 0.0095 & 0.0151 & 0.0080 & 0.987 & 0.084 & 4.51 \\
 &  & 1 & 17 & 0.0065 & 0.0102 & 0.0053 & 0.991 & 0.112 & 8.41 \\
 &  & 2 & 23 & 0.0097 & 0.0128 & 0.0091 & 0.986 & 0.088 & 4.83 \\
\cmidrule(l){2-10}
 & \multirow{3}{*}{$\alpha=0.5$} & 0 & 22 & 0.0132 & 0.0176 & 0.0153 & 0.953 & 0.089 & 4.73 \\
 &  & 1 & 19 & 0.0109 & 0.0146 & 0.0149 & 0.957 & 0.092 & 6.19 \\
 &  & 2 & 18 & 0.0088 & 0.0147 & 0.0090 & 0.969 & 0.130 & 8.01 \\
\cmidrule(l){2-10}
 & \multirow{3}{*}{$\alpha=1$} & 0 & 14 & 0.0024 & 0.0035 & 0.0137 & 0.899 & 0.104 & 6.29 \\
 &  & 1 & 11 & 0.0139 & 0.0131 & 0.0159 & 0.922 & 0.113 & 7.22 \\
 &  & 2 & 13 & 0.0032 & 0.0003 & 0.0105 & 0.897 & 0.071 & 5.76 \\
\cmidrule(l){2-10}
 & \multirow{3}{*}{$\alpha=1.5$} & 0 & 11 & 0.0063 & $-$0.0001 & 0.0045 & 0.939 & 0.038 & 2.66 \\
 &  & 1 & 7 & 0.0073 & 0.0060 & 0.0047 & 0.929 & 0.016 & 0.29 \\
 &  & 2 & 3 & 0.0055 & 0.0043 & $-$0.0017 & 0.972 & 0.057 & 4.87 \\
\cmidrule(l){2-10}
 & \multirow{3}{*}{$\alpha=1.75$} & 0 & 9 & 0.0071 & $-$0.0003 & 0.0036 & 0.938 & 0.047 & 3.66 \\
 &  & 1 & 2 & 0.0108 & 0.0169 & 0.0028 & 0.970 & 0.005 & 0.29 \\
 &  & 2 & 3 & 0.0090 & 0.0073 & 0.0007 & 0.978 & 0.043 & 3.74 \\
\cmidrule(l){2-10}
 & \multirow{3}{*}{$\alpha=2$} & 0 & 8 & 0.0074 & 0.0001 & 0.0034 & 0.928 & 0.039 & 2.81 \\
 &  & 1 & 0 & 0.0000 & 0.0000 & 0.0000 & 1.000 & 0.000 & 0.00 \\
 &  & 2 & 2 & 0.0120 & 0.0065 & 0.0016 & 0.969 & 0.008 & 0.59 \\
\cmidrule(l){2-10}
 & \multirow{3}{*}{Adaptive} & 0 & 15 & 0.0097 & 0.0150 & 0.0173 & 0.898 & 0.121 & 4.71 \\
 &  & 1 & 19 & 0.0131 & 0.0139 & 0.0203 & 0.885 & 0.112 & 6.04 \\
 &  & 2 & 17 & 0.0064 & 0.0087 & 0.0219 & 0.853 & 0.119 & 9.22 \\
\bottomrule
\end{tabular}
\end{adjustbox}
\end{table}

\begin{table}[!tbp]
\centering
\caption{Gain-threshold ablation, heuristic arm, every run (sequential; 18 runs of the gain-threshold study). The relaxed stage is switched off so that the threshold binds. Columns as in Table~\ref{tab:abldesc}.}
\label{tab:ablgain}
\footnotesize
\begin{tabular}{lllccccccc}
\toprule
Target & Threshold & Seed & Ed. & $\Delta P^{[A]}$ & $\Delta P^{[D]}$ & $\Delta S^{[D]}$ & CKA$_w$ & KL & $\Delta\mathrm{PPL}$ \\
\midrule
\multirow{9}{*}{Qwen} & \multirow{3}{*}{$\ge 0.005$} & 0 & 11 & 0.0334 & 0.0328 & 0.0120 & 0.900 & 0.100 & 7.03 \\
 &  & 1 & 10 & 0.0373 & 0.0304 & 0.0142 & 0.872 & 0.107 & 7.95 \\
 &  & 2 & 10 & 0.0332 & 0.0184 & 0.0087 & 0.868 & 0.076 & 5.67 \\
\cmidrule(l){2-10}
 & \multirow{3}{*}{$\ge 0.01$} & 0 & 10 & 0.0342 & 0.0344 & 0.0131 & 0.917 & 0.091 & 6.27 \\
 &  & 1 & 8 & 0.0358 & 0.0296 & 0.0126 & 0.889 & 0.089 & 6.17 \\
 &  & 2 & 8 & 0.0333 & 0.0228 & 0.0108 & 0.892 & 0.062 & 4.64 \\
\cmidrule(l){2-10}
 & \multirow{3}{*}{$\ge 0.02$} & 0 & 6 & 0.0321 & 0.0357 & 0.0167 & 0.944 & 0.043 & 2.77 \\
 &  & 1 & 5 & 0.0307 & 0.0261 & 0.0133 & 0.924 & 0.030 & 2.30 \\
 &  & 2 & 5 & 0.0362 & 0.0323 & 0.0199 & 0.932 & 0.026 & 1.21 \\
\midrule
\multirow{9}{*}{SmolLM2} & \multirow{3}{*}{$\ge 0.005$} & 0 & 11 & 0.0150 & 0.0223 & 0.0220 & 0.908 & 0.101 & 6.23 \\
 &  & 1 & 11 & 0.0116 & 0.0135 & 0.0164 & 0.914 & 0.117 & 5.91 \\
 &  & 2 & 9 & 0.0092 & 0.0094 & 0.0140 & 0.902 & 0.090 & 6.12 \\
\cmidrule(l){2-10}
 & \multirow{3}{*}{$\ge 0.01$} & 0 & 10 & 0.0153 & 0.0236 & 0.0222 & 0.921 & 0.091 & 5.34 \\
 &  & 1 & 9 & 0.0138 & 0.0167 & 0.0184 & 0.927 & 0.102 & 5.10 \\
 &  & 2 & 8 & 0.0065 & 0.0064 & 0.0075 & 0.942 & 0.081 & 6.43 \\
\cmidrule(l){2-10}
 & \multirow{3}{*}{$\ge 0.02$} & 0 & 6 & 0.0132 & 0.0219 & 0.0141 & 0.947 & 0.076 & 5.46 \\
 &  & 1 & 4 & 0.0136 & 0.0137 & 0.0187 & 0.940 & 0.030 & 1.29 \\
 &  & 2 & 6 & 0.0092 & 0.0097 & 0.0095 & 0.976 & 0.044 & 2.48 \\
\bottomrule
\end{tabular}
\end{table}

\begin{table}[!tbp]
\centering
\caption{Category-count ablation, LLM arm, every run (sequential; 18 runs of the category study). The first 4, 8, or 16 categories of the probe vocabulary are used (700, 1,359, and 2,369 probe words; 175, 341, and 595 unseen words). Columns as in Table~\ref{tab:abldesc}.}
\label{tab:ablcatllm}
\footnotesize
\begin{tabular}{lllccccccc}
\toprule
Target & Categories & Seed & Ed. & $\Delta P^{[A]}$ & $\Delta P^{[D]}$ & $\Delta S^{[D]}$ & CKA$_w$ & KL & $\Delta\mathrm{PPL}$ \\
\midrule
\multirow{9}{*}{Qwen} & \multirow{3}{*}{4} & 0 & 19 & 0.0159 & 0.0200 & 0.0339 & 0.773 & 0.067 & 8.21 \\
 &  & 1 & 16 & 0.0376 & 0.0325 & 0.0228 & 0.795 & 0.076 & 6.46 \\
 &  & 2 & 15 & 0.0310 & 0.0349 & 0.0463 & 0.857 & 0.084 & 6.14 \\
\cmidrule(l){2-10}
 & \multirow{3}{*}{8} & 0 & 13 & 0.0239 & 0.0287 & 0.0235 & 0.897 & 0.089 & 6.45 \\
 &  & 1 & 20 & 0.0269 & 0.0196 & 0.0484 & 0.935 & 0.086 & 6.96 \\
 &  & 2 & 23 & 0.0156 & 0.0264 & 0.0453 & 0.919 & 0.104 & 8.72 \\
\cmidrule(l){2-10}
 & \multirow{3}{*}{16} & 0 & 25 & $-$0.0009 & $-$0.0044 & 0.0185 & 0.914 & 0.057 & 6.30 \\
 &  & 1 & 18 & $-$0.0027 & $-$0.0101 & 0.0212 & 0.923 & 0.083 & 6.23 \\
 &  & 2 & 23 & 0.0011 & 0.0005 & 0.0247 & 0.909 & 0.087 & 6.97 \\
\midrule
\multirow{9}{*}{SmolLM2} & \multirow{3}{*}{4} & 0 & 17 & 0.0005 & 0.0125 & 0.0141 & 0.896 & 0.081 & 7.59 \\
 &  & 1 & 17 & 0.0056 & 0.0120 & 0.0181 & 0.971 & 0.091 & 6.10 \\
 &  & 2 & 11 & 0.0011 & 0.0155 & 0.0168 & 0.908 & 0.066 & 4.60 \\
\cmidrule(l){2-10}
 & \multirow{3}{*}{8} & 0 & 13 & 0.0017 & 0.0111 & 0.0266 & 0.902 & 0.092 & 6.82 \\
 &  & 1 & 15 & 0.0067 & 0.0104 & 0.0166 & 0.962 & 0.083 & 6.50 \\
 &  & 2 & 16 & 0.0095 & 0.0117 & 0.0163 & 0.953 & 0.099 & 8.17 \\
\cmidrule(l){2-10}
 & \multirow{3}{*}{16} & 0 & 18 & $-$0.0111 & $-$0.0100 & 0.0049 & 0.961 & 0.131 & 7.87 \\
 &  & 1 & 22 & $-$0.0095 & $-$0.0046 & 0.0142 & 0.942 & 0.148 & 7.17 \\
 &  & 2 & 22 & $-$0.0086 & $-$0.0065 & 0.0103 & 0.946 & 0.152 & 6.26 \\
\bottomrule
\end{tabular}
\end{table}

\begin{table}[!tbp]
\centering
\caption{Category-count ablation, heuristic, every run (sequential; 18 runs of the category study). Columns as in Table~\ref{tab:abldesc}.}
\label{tab:ablcatrule}
\footnotesize
\begin{tabular}{lllccccccc}
\toprule
Target & Categories & Seed & Ed. & $\Delta P^{[A]}$ & $\Delta P^{[D]}$ & $\Delta S^{[D]}$ & CKA$_w$ & KL & $\Delta\mathrm{PPL}$ \\
\midrule
\multirow{9}{*}{Qwen} & \multirow{3}{*}{4} & 0 & 21 & 0.0551 & 0.0821 & 0.0696 & 0.798 & 0.112 & 13.81 \\
 &  & 1 & 19 & 0.0520 & 0.0454 & 0.0584 & 0.796 & 0.119 & 14.15 \\
 &  & 2 & 19 & 0.0563 & 0.0535 & 0.0649 & 0.781 & 0.115 & 13.59 \\
\cmidrule(l){2-10}
 & \multirow{3}{*}{8} & 0 & 23 & 0.0344 & 0.0285 & 0.0158 & 0.837 & 0.115 & 8.14 \\
 &  & 1 & 21 & 0.0315 & 0.0235 & 0.0153 & 0.849 & 0.110 & 8.05 \\
 &  & 2 & 20 & 0.0360 & 0.0162 & 0.0110 & 0.812 & 0.108 & 8.26 \\
\cmidrule(l){2-10}
 & \multirow{3}{*}{16} & 0 & 17 & 0.0197 & 0.0167 & 0.0042 & 0.904 & 0.070 & 5.93 \\
 &  & 1 & 19 & 0.0165 & 0.0110 & 0.0084 & 0.897 & 0.066 & 6.94 \\
 &  & 2 & 20 & 0.0217 & 0.0102 & 0.0124 & 0.903 & 0.075 & 6.03 \\
\midrule
\multirow{9}{*}{SmolLM2} & \multirow{3}{*}{4} & 0 & 12 & 0.0225 & 0.0502 & 0.0401 & 0.898 & 0.080 & 7.46 \\
 &  & 1 & 9 & 0.0253 & 0.0236 & 0.0324 & 0.927 & 0.070 & 5.75 \\
 &  & 2 & 14 & 0.0241 & 0.0383 & 0.0371 & 0.920 & 0.123 & 13.23 \\
\cmidrule(l){2-10}
 & \multirow{3}{*}{8} & 0 & 15 & 0.0097 & 0.0150 & 0.0173 & 0.898 & 0.121 & 4.71 \\
 &  & 1 & 19 & 0.0131 & 0.0139 & 0.0203 & 0.885 & 0.112 & 6.04 \\
 &  & 2 & 17 & 0.0064 & 0.0087 & 0.0219 & 0.853 & 0.119 & 9.22 \\
\cmidrule(l){2-10}
 & \multirow{3}{*}{16} & 0 & 14 & 0.0012 & 0.0030 & 0.0064 & 0.915 & 0.051 & 2.68 \\
 &  & 1 & 15 & 0.0047 & 0.0061 & 0.0106 & 0.917 & 0.105 & 6.16 \\
 &  & 2 & 15 & 0.0024 & $-$0.0035 & 0.0068 & 0.931 & 0.080 & 3.90 \\
\bottomrule
\end{tabular}
\end{table}

\begin{table}[!tbp]
\centering
\caption{Tests of the purity gain on unseen words in unseen templates ($\Delta P^{[D]}$) for every ablation variant, per arm over both targets and three seeds ($n=6$; descriptive summary over a clustered design of 2 targets $\times$ 3 data splits; the per-run Tables~\ref{tab:abldesc} to~\ref{tab:ablcatrule} are primary). Mean, 95\% $t$-interval, $p$-value against zero, and the paired difference to the reference variant of the same arm (first row of each block) with its $p$-value. $p$-values are uncorrected.}
\label{tab:pooled}
\begin{adjustbox}{max width=\textwidth}
\footnotesize
\begin{tabular}{lllccccc}
\toprule
Study & Arm & Variant & Mean & 95\% interval & $p$ (vs. 0) & Diff. to ref. & $p$ (paired) \\
\midrule
Descriptors & LLM arm & Full (reference) & 0.0180 & [0.0094, 0.0266] & 0.003 & -- & -- \\
 & LLM arm & Purity + silhouette & 0.0068 & [$-$0.0035, 0.0172] & 0.149 & $-$0.0111 & 0.009 \\
 & LLM arm & + anisotropy & 0.0072 & [0.0016, 0.0128] & 0.021 & $-$0.0108 & 0.024 \\
 & LLM arm & + intrinsic dimension, Betti & 0.0149 & [0.0043, 0.0255] & 0.015 & $-$0.0031 & 0.270 \\
\addlinespace
Strength & Heuristic & Adaptive (reference) & 0.0176 & [0.0101, 0.0251] & 0.002 & -- & -- \\
 & Heuristic & $\alpha=0.25$ & 0.0155 & [0.0118, 0.0191] & $<$0.001 & $-$0.0022 & 0.376 \\
 & Heuristic & $\alpha=0.5$ & 0.0219 & [0.0146, 0.0292] & $<$0.001 & $+$0.0043 & 0.051 \\
 & Heuristic & $\alpha=1$ & 0.0165 & [0.0027, 0.0303] & 0.028 & $-$0.0011 & 0.729 \\
 & Heuristic & $\alpha=1.5$ & 0.0138 & [0.0016, 0.0260] & 0.034 & $-$0.0038 & 0.268 \\
 & Heuristic & $\alpha=1.75$ & 0.0137 & [0.0039, 0.0235] & 0.015 & $-$0.0039 & 0.327 \\
 & Heuristic & $\alpha=2$ & 0.0053 & [0.0001, 0.0106] & 0.047 & $-$0.0123 & 0.007 \\
\addlinespace
Gain threshold & Heuristic & Gain $\ge 0.005$ (reference) & 0.0211 & [0.0114, 0.0308] & 0.003 & -- & -- \\
 & Heuristic & Gain $\ge 0.01$ & 0.0223 & [0.0119, 0.0326] & 0.003 & $+$0.0011 & 0.347 \\
 & Heuristic & Gain $\ge 0.02$ & 0.0233 & [0.0125, 0.0340] & 0.003 & $+$0.0021 & 0.442 \\
\addlinespace
Categories & LLM arm & 4 categories & 0.0212 & [0.0107, 0.0318] & 0.004 & -- & -- \\
 & LLM arm & 8 categories & 0.0180 & [0.0094, 0.0266] & 0.003 & $-$0.0033 & 0.328 \\
 & LLM arm & 16 categories & $-$0.0058 & [$-$0.0100, $-$0.0017] & 0.016 & $-$0.0271 & $<$0.001 \\
 & Heuristic & 4 categories & 0.0489 & [0.0285, 0.0693] & 0.002 & -- & -- \\
 & Heuristic & 8 categories & 0.0176 & [0.0101, 0.0251] & 0.002 & $-$0.0312 & 0.004 \\
 & Heuristic & 16 categories & 0.0073 & [$-$0.0001, 0.0146] & 0.053 & $-$0.0416 & 0.001 \\
\bottomrule
\end{tabular}
\end{adjustbox}
\end{table}

\section{Absolute Behavioral Values}
\label{app:abs}

Table~\ref{tab:abs} lists the absolute behavioral values of every default and parallel run.

\begin{table}[!tbp]
\centering
\caption{Absolute behavioral values for every seed: the unedited target, each arm in the sequential mode, and the parallel control (30 default and parallel runs). Benchmark samples (200 examples each) are redrawn per seed, so the unedited values differ between seeds. PPL: perplexity on the calibration, guard, held-out, and new-domain sentences; Cloze$_p$, Cloze$_u$: cloze accuracy on probe and unseen words; HS, ARC, TQA: benchmark accuracies; Fact: factual-preference accuracy; TM: truth margin (nats).}
\label{tab:abs}
\begin{adjustbox}{max width=\textwidth}
\scriptsize
\setlength{\tabcolsep}{3pt}
\begin{tabular}{lllrrrrrrrrrrr}
\toprule
Target & Seed & Model & $\mathrm{PPL}_{\mathrm{cal}}$ & $\mathrm{PPL}_{\mathrm{guard}}$ & $\mathrm{PPL}_{\mathrm{held}}$ & $\mathrm{PPL}_{\mathrm{new}}$ & Cloze$_p$ & Cloze$_u$ & HS & ARC & TQA & Fact & TM \\
\midrule
\multirow{18}{*}{Qwen} & \multirow{6}{*}{0} & Unedited & 45.98 & 49.11 & 43.72 & 61.36 & 0.638 & 0.710 & 0.620 & 0.750 & 0.275 & 0.975 & 6.92 \\
 &  & LLM arm, seq. & 48.66 & 51.57 & 46.54 & 62.44 & 0.706 & 0.760 & 0.630 & 0.720 & 0.275 & 0.975 & 6.67 \\
 &  & Heuristic, seq. & 50.33 & 51.12 & 47.28 & 60.35 & 0.696 & 0.733 & 0.630 & 0.740 & 0.270 & 0.975 & 6.73 \\
 &  & Hybrid, seq. & 49.96 & 50.40 & 46.90 & 60.75 & 0.679 & 0.733 & 0.625 & 0.730 & 0.255 & 0.975 & 6.67 \\
 &  & LLM arm, par. & 101.38 & 98.12 & 92.51 & 79.41 & 0.634 & 0.613 & 0.600 & 0.690 & 0.285 & 0.950 & 6.13 \\
 &  & Heuristic, par. & 8{,}571 & 9{,}388 & 8{,}370 & 2{,}340 & 0.191 & 0.220 & 0.390 & 0.320 & 0.315 & 0.800 & 2.92 \\
\cmidrule(l){2-14}
 & \multirow{6}{*}{1} & Unedited & 45.98 & 49.11 & 43.72 & 61.36 & 0.628 & 0.765 & 0.640 & 0.755 & 0.255 & 0.975 & 6.92 \\
 &  & LLM arm, seq. & 49.98 & 51.21 & 46.76 & 58.56 & 0.693 & 0.762 & 0.645 & 0.770 & 0.245 & 0.975 & 6.51 \\
 &  & Heuristic, seq. & 49.59 & 51.42 & 47.24 & 60.72 & 0.670 & 0.795 & 0.635 & 0.760 & 0.245 & 0.975 & 6.69 \\
 &  & Hybrid, seq. & 50.17 & 51.55 & 47.31 & 60.02 & 0.674 & 0.780 & 0.640 & 0.755 & 0.240 & 0.975 & 6.61 \\
 &  & LLM arm, par. & 99.53 & 97.94 & 89.02 & 81.07 & 0.663 & 0.689 & 0.645 & 0.730 & 0.250 & 0.950 & 5.91 \\
 &  & Heuristic, par. & 5{,}428 & 5{,}962 & 5{,}436 & 1{,}393 & 0.319 & 0.358 & 0.405 & 0.360 & 0.345 & 0.725 & 2.53 \\
\cmidrule(l){2-14}
 & \multirow{6}{*}{2} & Unedited & 45.98 & 49.11 & 43.72 & 61.36 & 0.636 & 0.724 & 0.585 & 0.775 & 0.315 & 0.975 & 6.92 \\
 &  & LLM arm, seq. & 51.05 & 51.56 & 47.53 & 59.03 & 0.720 & 0.751 & 0.595 & 0.770 & 0.310 & 0.975 & 6.36 \\
 &  & Heuristic, seq. & 49.73 & 50.72 & 47.33 & 60.51 & 0.680 & 0.751 & 0.580 & 0.760 & 0.310 & 0.975 & 6.66 \\
 &  & Hybrid, seq. & 49.78 & 51.49 & 47.57 & 60.98 & 0.691 & 0.771 & 0.585 & 0.760 & 0.315 & 0.975 & 6.79 \\
 &  & LLM arm, par. & 116.31 & 109.09 & 102.48 & 85.80 & 0.636 & 0.666 & 0.590 & 0.775 & 0.310 & 0.900 & 5.73 \\
 &  & Heuristic, par. & 7{,}005 & 6{,}719 & 6{,}787 & 2{,}057 & 0.246 & 0.302 & 0.340 & 0.365 & 0.400 & 0.725 & 2.90 \\
\midrule
\multirow{18}{*}{SmolLM2} & \multirow{6}{*}{0} & Unedited & 42.92 & 48.47 & 40.77 & 55.06 & 0.614 & 0.736 & 0.715 & 0.640 & 0.245 & 1.000 & 8.18 \\
 &  & LLM arm, seq. & 45.83 & 50.80 & 43.56 & 57.36 & 0.664 & 0.748 & 0.720 & 0.645 & 0.245 & 1.000 & 7.62 \\
 &  & Heuristic, seq. & 44.93 & 50.87 & 42.69 & 55.75 & 0.665 & 0.798 & 0.675 & 0.625 & 0.245 & 1.000 & 7.71 \\
 &  & Hybrid, seq. & 45.72 & 50.85 & 43.00 & 57.13 & 0.670 & 0.783 & 0.685 & 0.630 & 0.240 & 1.000 & 7.77 \\
 &  & LLM arm, par. & 110.40 & 113.04 & 108.37 & 136.54 & 0.581 & 0.686 & 0.625 & 0.590 & 0.215 & 0.925 & 6.52 \\
 &  & Heuristic, par. & 2{,}175 & 1{,}879 & 1{,}821 & 1{,}095 & 0.386 & 0.522 & 0.505 & 0.535 & 0.350 & 0.725 & 2.92 \\
\cmidrule(l){2-14}
 & \multirow{6}{*}{1} & Unedited & 42.92 & 48.47 & 40.77 & 55.06 & 0.611 & 0.768 & 0.720 & 0.690 & 0.225 & 1.000 & 8.18 \\
 &  & LLM arm, seq. & 45.77 & 50.89 & 43.42 & 56.95 & 0.673 & 0.801 & 0.710 & 0.675 & 0.225 & 1.000 & 8.10 \\
 &  & Heuristic, seq. & 45.50 & 50.83 & 43.24 & 60.37 & 0.659 & 0.795 & 0.700 & 0.650 & 0.240 & 1.000 & 7.61 \\
 &  & Hybrid, seq. & 45.22 & 50.81 & 43.07 & 60.27 & 0.672 & 0.783 & 0.700 & 0.655 & 0.240 & 1.000 & 7.79 \\
 &  & LLM arm, par. & 114.62 & 113.23 & 113.89 & 145.01 & 0.604 & 0.757 & 0.645 & 0.580 & 0.235 & 0.900 & 5.98 \\
 &  & Heuristic, par. & 2{,}363 & 2{,}076 & 2{,}008 & 1{,}037 & 0.455 & 0.560 & 0.460 & 0.520 & 0.355 & 0.650 & 2.49 \\
\cmidrule(l){2-14}
 & \multirow{6}{*}{2} & Unedited & 42.92 & 48.47 & 40.77 & 55.06 & 0.617 & 0.733 & 0.665 & 0.680 & 0.300 & 1.000 & 8.18 \\
 &  & LLM arm, seq. & 46.28 & 50.84 & 44.11 & 57.91 & 0.683 & 0.745 & 0.665 & 0.685 & 0.295 & 1.000 & 8.16 \\
 &  & Heuristic, seq. & 47.08 & 50.87 & 44.53 & 58.59 & 0.693 & 0.777 & 0.665 & 0.680 & 0.310 & 1.000 & 7.75 \\
 &  & Hybrid, seq. & 46.92 & 50.36 & 44.54 & 58.27 & 0.699 & 0.751 & 0.675 & 0.690 & 0.315 & 1.000 & 7.94 \\
 &  & LLM arm, par. & 118.10 & 118.53 & 116.25 & 146.83 & 0.597 & 0.727 & 0.600 & 0.630 & 0.275 & 0.925 & 6.42 \\
 &  & Heuristic, par. & 2{,}503 & 1{,}998 & 2{,}041 & 1{,}015 & 0.384 & 0.557 & 0.465 & 0.495 & 0.375 & 0.725 & 2.85 \\
\bottomrule
\end{tabular}
\end{adjustbox}
\end{table}

On the new-domain sentences, the perplexity changed by $-4.6\%$ to $+1.8\%$ (Qwen) and $+3.4\%$ to $+5.2\%$ (SmolLM2) in the analyst's sequential runs and by $-1.6\%$ to $-1.0\%$ and $+1.2\%$ to $+9.6\%$ in the heuristic's.

\section{Per-Layer Trajectory Summaries}
\label{app:traj}

Table~\ref{tab:traj} gives the per-layer trajectory summaries, including the all-layer averages referred to in Section~\ref{sec:main}.

\begin{table}[!tbp]
\centering
\caption{Per-layer trajectories under condition $[D]$ for every default sequential run. The purity change (edited minus original) is averaged over all layers of the first, middle, and last third of the depth (layers 1--10, 11--19, and 20--28 for Qwen; 1--8, 9--16, and 17--24 for SmolLM2) and over all layers, edited or not. $P_L$: purity at the last layer in the original and in the edited model. Improved: number of layers whose purity under $[D]$ rose.}
\label{tab:traj}
\begin{adjustbox}{max width=\textwidth}
\small
\begin{tabular}{lllcccccccc}
\toprule
 & & & \multicolumn{4}{c}{$\Delta P^{[D]}$ over all layers} & \multicolumn{3}{c}{Last layer} & \\
\cmidrule(lr){4-7}\cmidrule(lr){8-10}
Target & Arm & Seed & First third & Middle third & Last third & All & $P_L$ orig. & $P_L$ edited & $\Delta P_L$ & Improved \\
\midrule
\multirow{9}{*}{Qwen} & \multirow{3}{*}{LLM arm} & 0 & 0.0188 & 0.0254 & 0.0647 & 0.0357 & 0.599 & 0.661 & $+$0.061 & 28/28 \\
 &  & 1 & 0.0111 & 0.0130 & 0.0417 & 0.0216 & 0.663 & 0.723 & $+$0.060 & 28/28 \\
 &  & 2 & 0.0133 & 0.0226 & 0.0526 & 0.0289 & 0.647 & 0.707 & $+$0.060 & 27/28 \\
\cmidrule(l){2-11}
 & \multirow{3}{*}{Heuristic} & 0 & 0.0224 & 0.0148 & 0.0434 & 0.0267 & 0.599 & 0.664 & $+$0.065 & 28/28 \\
 &  & 1 & 0.0206 & 0.0197 & 0.0364 & 0.0254 & 0.663 & 0.722 & $+$0.059 & 28/28 \\
 &  & 2 & 0.0098 & 0.0000 & 0.0284 & 0.0126 & 0.647 & 0.702 & $+$0.055 & 20/28 \\
\cmidrule(l){2-11}
 & \multirow{3}{*}{Hybrid} & 0 & 0.0173 & 0.0077 & 0.0417 & 0.0221 & 0.599 & 0.665 & $+$0.066 & 26/28 \\
 &  & 1 & 0.0173 & 0.0196 & 0.0418 & 0.0259 & 0.663 & 0.733 & $+$0.070 & 28/28 \\
 &  & 2 & 0.0127 & 0.0030 & 0.0404 & 0.0185 & 0.647 & 0.709 & $+$0.062 & 23/28 \\
\midrule
\multirow{9}{*}{SmolLM2} & \multirow{3}{*}{LLM arm} & 0 & 0.0040 & 0.0126 & 0.0228 & 0.0131 & 0.596 & 0.622 & $+$0.026 & 23/24 \\
 &  & 1 & 0.0040 & 0.0039 & 0.0237 & 0.0105 & 0.655 & 0.705 & $+$0.050 & 23/24 \\
 &  & 2 & 0.0027 & 0.0043 & 0.0197 & 0.0089 & 0.636 & 0.688 & $+$0.052 & 21/24 \\
\cmidrule(l){2-11}
 & \multirow{3}{*}{Heuristic} & 0 & 0.0047 & 0.0130 & 0.0228 & 0.0135 & 0.596 & 0.653 & $+$0.057 & 23/24 \\
 &  & 1 & 0.0062 & 0.0093 & 0.0210 & 0.0121 & 0.655 & 0.716 & $+$0.061 & 22/24 \\
 &  & 2 & $-$0.0011 & 0.0083 & 0.0160 & 0.0077 & 0.636 & 0.679 & $+$0.043 & 19/24 \\
\cmidrule(l){2-11}
 & \multirow{3}{*}{Hybrid} & 0 & 0.0063 & 0.0128 & 0.0196 & 0.0129 & 0.596 & 0.653 & $+$0.057 & 24/24 \\
 &  & 1 & 0.0066 & 0.0096 & 0.0210 & 0.0124 & 0.655 & 0.713 & $+$0.058 & 23/24 \\
 &  & 2 & 0.0004 & 0.0044 & 0.0137 & 0.0062 & 0.636 & 0.678 & $+$0.042 & 20/24 \\
\bottomrule
\end{tabular}
\end{adjustbox}
\end{table}

In every run, the purity change in the last third of the layers exceeded that in the first third, and the change at the last layer, which the output head reads, was between $+0.026$ and $+0.070$. Under $[D]$, purity rose at 150 of the 156 layers of the six models of the LLM arm, at 140 of those of the heuristic, and at 144 of those of the hybrid arm.

Table~\ref{tab:topo} reports the topological descriptors of the unseen words before and after editing. In all 18 runs, both Betti numbers were lower in the edited model. On probe words (condition $[A]$), $\beta_0$ fell in all 18 runs, and $\beta_1$ fell in all runs of the heuristic and hybrid arms and in four of the six runs of the LLM arm.

\begin{table}[!tbp]
\centering
\caption{Topological descriptors under condition $[D]$ for every default sequential run, in the original and in the edited model (mean over the edited layers and over the three unseen templates; class-balanced subsample of 279 unseen words). $\beta_0$, $\beta_1$: Betti numbers at the 25th percentile of the pairwise cosine distances of the respective model; ID: Two-NN intrinsic dimension.}
\label{tab:topo}
\small
\begin{tabular}{lllcccccc}
\toprule
 & & & \multicolumn{2}{c}{$\beta_0$ $\downarrow$} & \multicolumn{2}{c}{$\beta_1$ $\downarrow$} & \multicolumn{2}{c}{ID} \\
\cmidrule(lr){4-5}\cmidrule(lr){6-7}\cmidrule(lr){8-9}
Target & Arm & Seed & orig. & edited & orig. & edited & orig. & edited \\
\midrule
 \multirow{9}{*}{Qwen} & \multirow{3}{*}{LLM arm} & 0 & 6.6 & 4.2 & 2.6 & 2.5 & 21.3 & 21.3 \\
  &  & 1 & 8.1 & 4.2 & 4.2 & 3.2 & 19.1 & 18.7 \\
  &  & 2 & 6.6 & 3.1 & 4.0 & 2.6 & 20.6 & 21.0 \\
\cmidrule(l){2-9}
  & \multirow{3}{*}{Heuristic} & 0 & 9.5 & 5.4 & 2.8 & 2.4 & 21.0 & 21.2 \\
  &  & 1 & 7.7 & 4.4 & 4.3 & 3.0 & 19.1 & 18.8 \\
  &  & 2 & 8.6 & 5.8 & 3.2 & 2.5 & 20.6 & 19.7 \\
\cmidrule(l){2-9}
  & \multirow{3}{*}{Hybrid} & 0 & 8.7 & 5.5 & 2.7 & 2.1 & 21.0 & 21.1 \\
  &  & 1 & 9.8 & 4.2 & 3.8 & 2.9 & 18.8 & 18.4 \\
  &  & 2 & 7.5 & 5.1 & 4.0 & 3.0 & 20.3 & 19.9 \\
\midrule
 \multirow{9}{*}{SmolLM2} & \multirow{3}{*}{LLM arm} & 0 & 15.3 & 11.0 & 2.2 & 1.9 & 19.7 & 19.7 \\
  &  & 1 & 16.2 & 13.8 & 2.6 & 2.0 & 19.1 & 19.0 \\
  &  & 2 & 13.9 & 11.9 & 2.1 & 1.8 & 21.4 & 20.9 \\
\cmidrule(l){2-9}
  & \multirow{3}{*}{Heuristic} & 0 & 16.2 & 12.1 & 2.5 & 1.9 & 19.6 & 19.2 \\
  &  & 1 & 18.2 & 14.5 & 2.9 & 2.4 & 18.2 & 18.0 \\
  &  & 2 & 13.8 & 9.4 & 3.6 & 2.1 & 21.8 & 21.9 \\
\cmidrule(l){2-9}
  & \multirow{3}{*}{Hybrid} & 0 & 15.4 & 11.0 & 2.6 & 1.9 & 19.6 & 19.2 \\
  &  & 1 & 17.6 & 14.8 & 2.6 & 2.4 & 18.3 & 18.2 \\
  &  & 2 & 13.6 & 9.1 & 3.4 & 2.0 & 21.3 & 21.5 \\
\bottomrule
\end{tabular}
\end{table}

\section{Category-Level Breakdown}
\label{app:categories}

Table~\ref{tab:cat} gives the category-level changes of every default sequential run.

\begin{table}[!tbp]
\centering
\caption{Category-level changes under condition $[D]$ for every default sequential run (mean over the edited layers of each run). Columns 0, 1, 2 are the seeds.}
\label{tab:cat}
\begin{adjustbox}{max width=\textwidth}
\footnotesize
\setlength{\tabcolsep}{3pt}
\begin{tabular}{llrrrrrrrrrrrr}
\toprule
 & & \multicolumn{6}{c}{Target: Qwen} & \multicolumn{6}{c}{Target: SmolLM2} \\
\cmidrule(lr){3-8}\cmidrule(lr){9-14}
 & & \multicolumn{3}{c}{$\Delta P$} & \multicolumn{3}{c}{$\Delta S$} & \multicolumn{3}{c}{$\Delta P$} & \multicolumn{3}{c}{$\Delta S$} \\
\cmidrule(lr){3-5}\cmidrule(lr){6-8}\cmidrule(lr){9-11}\cmidrule(lr){12-14}
Category & Arm & 0 & 1 & 2 & 0 & 1 & 2 & 0 & 1 & 2 & 0 & 1 & 2 \\
\midrule
\multirow{3}{*}{Animal} & LLM arm & 0.051 & 0.035 & 0.030 & $-$0.009 & 0.029 & $-$0.015 & 0.007 & 0.017 & 0.000 & $-$0.028 & 0.020 & $-$0.010 \\
 & Heuristic & 0.146 & 0.102 & 0.038 & 0.061 & 0.083 & 0.006 & 0.050 & 0.031 & $-$0.011 & 0.028 & 0.042 & 0.016 \\
 & Hybrid & 0.121 & 0.096 & 0.034 & 0.054 & 0.070 & $-$0.018 & 0.059 & 0.023 & $-$0.004 & 0.038 & 0.026 & 0.021 \\
\midrule
\multirow{3}{*}{Fruit} & LLM arm & $-$0.013 & 0.000 & 0.014 & 0.006 & 0.052 & 0.022 & 0.017 & $-$0.009 & $-$0.008 & 0.017 & 0.003 & 0.006 \\
 & Heuristic & $-$0.054 & $-$0.057 & $-$0.064 & $-$0.029 & $-$0.032 & $-$0.023 & $-$0.039 & $-$0.053 & $-$0.064 & $-$0.027 & $-$0.057 & $-$0.046 \\
 & Hybrid & $-$0.057 & $-$0.046 & $-$0.039 & $-$0.029 & $-$0.043 & $-$0.003 & $-$0.057 & $-$0.049 & $-$0.061 & $-$0.036 & $-$0.044 & $-$0.038 \\
\midrule
\multirow{3}{*}{Vehicle} & LLM arm & $-$0.026 & $-$0.004 & $-$0.006 & $-$0.015 & $-$0.006 & 0.037 & $-$0.004 & 0.004 & 0.026 & $-$0.004 & 0.000 & 0.011 \\
 & Heuristic & 0.040 & 0.034 & 0.066 & 0.044 & 0.024 & 0.051 & 0.010 & 0.027 & 0.052 & 0.034 & 0.059 & 0.084 \\
 & Hybrid & 0.029 & 0.039 & 0.058 & 0.028 & 0.048 & 0.053 & 0.029 & 0.030 & 0.050 & 0.048 & 0.046 & 0.069 \\
\midrule
\multirow{3}{*}{Tool} & LLM arm & 0.090 & 0.046 & 0.046 & 0.033 & 0.073 & 0.066 & 0.059 & 0.003 & 0.004 & 0.055 & 0.011 & 0.021 \\
 & Heuristic & 0.053 & 0.015 & 0.025 & 0.025 & 0.012 & 0.041 & 0.045 & 0.014 & 0.004 & 0.021 & 0.034 & 0.021 \\
 & Hybrid & 0.057 & 0.010 & 0.018 & 0.036 & 0.022 & 0.029 & 0.052 & 0.011 & 0.005 & 0.025 & 0.031 & 0.021 \\
\midrule
\multirow{3}{*}{Color} & LLM arm & $-$0.008 & $-$0.002 & $-$0.030 & 0.014 & 0.001 & $-$0.078 & 0.002 & 0.002 & 0.003 & $-$0.001 & $-$0.002 & $-$0.006 \\
 & Heuristic & $-$0.020 & 0.006 & 0.029 & $-$0.012 & $-$0.001 & $-$0.003 & $-$0.008 & 0.025 & 0.038 & $-$0.003 & 0.008 & 0.012 \\
 & Hybrid & $-$0.029 & $-$0.001 & 0.023 & $-$0.015 & 0.003 & $-$0.002 & $-$0.019 & 0.009 & 0.017 & $-$0.011 & 0.001 & 0.005 \\
\midrule
\multirow{3}{*}{Country} & LLM arm & 0.006 & 0.002 & 0.001 & 0.040 & 0.107 & 0.120 & 0.011 & 0.002 & 0.005 & 0.169 & 0.048 & 0.068 \\
 & Heuristic & $-$0.017 & $-$0.024 & $-$0.018 & $-$0.026 & $-$0.032 & $-$0.033 & $-$0.009 & $-$0.008 & $-$0.006 & 0.001 & 0.007 & 0.006 \\
 & Hybrid & $-$0.014 & $-$0.018 & $-$0.016 & $-$0.033 & 0.004 & $-$0.021 & $-$0.013 & $-$0.006 & $-$0.006 & 0.003 & 0.017 & 0.003 \\
\midrule
\multirow{3}{*}{Emotion} & LLM arm & 0.016 & 0.026 & 0.070 & 0.040 & 0.090 & 0.115 & 0.007 & 0.007 & 0.010 & 0.007 & 0.025 & 0.019 \\
 & Heuristic & $-$0.017 & $-$0.008 & $-$0.018 & $-$0.006 & $-$0.006 & $-$0.005 & 0.003 & 0.004 & 0.006 & 0.004 & 0.004 & 0.009 \\
 & Hybrid & $-$0.016 & 0.000 & $-$0.014 & $-$0.004 & 0.018 & $-$0.004 & $-$0.001 & 0.007 & $-$0.002 & 0.004 & 0.019 & 0.006 \\
\midrule
\multirow{3}{*}{Profession} & LLM arm & 0.100 & 0.043 & 0.057 & 0.074 & 0.015 & 0.017 & $-$0.007 & 0.049 & 0.045 & $-$0.020 & 0.011 & 0.006 \\
 & Heuristic & 0.053 & 0.093 & 0.064 & 0.047 & 0.055 & 0.047 & 0.049 & 0.064 & 0.051 & 0.062 & 0.043 & 0.053 \\
 & Hybrid & 0.052 & 0.099 & 0.064 & 0.035 & 0.062 & 0.062 & 0.039 & 0.074 & 0.047 & 0.054 & 0.046 & 0.035 \\
\bottomrule
\end{tabular}
\end{adjustbox}
\end{table}

The heuristic and the hybrid arm lowered the purity of the fruit category in all twelve of their runs, by 0.039 to 0.064. For the analyst, the purity change of the fruit category was $-0.013$ to $+0.017$, and its largest loss in any category and run was 0.030 (color, on Qwen). In the unedited models, the country and emotion categories already had a purity of 0.94 to 0.98 and 0.80 to 0.90 under condition $[D]$ (mean over all layers); for these two categories, the analyst's silhouette change exceeded its purity change in 11 of the 12 category--run combinations (the exception is emotion on SmolLM2, seed 0, where both changes were 0.007).

\section{Probe Words, Templates, and Evaluation Sentences}
\label{app:data}

For the templates and prompts, the probe template is ``I thought about the \{w\}''. The three unseen templates are ``The technical report included a short section about the \{w\}'', ``During the court hearing, the witness mentioned the \{w\}'', and ``The next slide of the lecture shows a picture of the \{w\}''; a value under conditions $[C]$ and $[D]$ is the mean over the three templates. The cloze prompt is `The word ``\{w\}'' refers to a kind of' for probe words and `Question: what kind of thing is ``\{w\}''? Answer: it is a kind of' for unseen words; the prediction is the category whose first token has the highest logit.

As for the sentence set, the calibration set contains 200 sentences from the validation split of WikiText-103 \cite{merity2017pointer}, and the guard and held-out sets contain 24 and 476 sentences from its test split (sentences of 8 to 40 words, drawn with a fixed seed). An example of a calibration sentence is ``She was indicted by a federal grand jury in 1990 .'', and an example of a held-out sentence is ``Light rainfall also extended into North Carolina .''. The new-domain set contains 500 sentences in five domains (code, medical, legal, chat, and finance) that are template-generated by filling domain-specific sentence frames with domain vocabulary, for example ``A history of hypertension increases the risk of a persistent cough after treatment with the injection.''

For the factual-preference probe, each of the 40 prompts is listed with its true and its false completion: The capital of France is (Paris / Rome); The capital of Japan is (Tokyo / Beijing); The capital of Italy is (Rome / Madrid); The capital of Germany is (Berlin / Vienna); The capital of Egypt is (Cairo / Baghdad); The capital of Canada is (Ottawa / Toronto); The capital of Australia is (Canberra / Sydney); The capital of Russia is (Moscow / Warsaw); The largest planet in the solar system is (Jupiter / Mars); The planet closest to the sun is (Mercury / Venus); Water is made of hydrogen and (oxygen / nitrogen); The chemical symbol for gold is (Au / Ag); The chemical symbol for sodium is (Na / So); Humans breathe in oxygen and breathe out carbon (dioxide / monoxide); The freezing point of water in degrees Celsius is (0 / 32); There are seven days in a (week / month); A triangle has three (sides / corners only); Two plus two equals (four / five); Ten times ten equals one (hundred / thousand); The square root of nine is (three / four); The first month of the year is (January / March); The season after winter is (spring / autumn); The longest river in Africa is the (Nile / Congo); Mount Everest is located in the (Himalayas / Alps); The Great Wall is located in (China / India); The Eiffel Tower is located in (Paris / London); The Pacific is the largest (ocean / desert); The Sahara is a large (desert / forest); Romeo and Juliet was written by William (Shakespeare / Wordsworth); The theory of relativity was developed by Albert (Einstein / Newton); The first person to walk on the moon was Neil (Armstrong / Aldrin); The author of the Harry Potter books is J. K. (Rowling / Tolkien); The currency of Japan is the (yen / yuan); The currency of the United Kingdom is the (pound / euro); The language mainly spoken in Brazil is (Portuguese / Spanish); The language mainly spoken in Mexico is (Spanish / Portuguese); Bees produce (honey / milk); A spider has eight (legs / wings); The heart pumps (blood / air); The opposite of hot is (cold / warm).

Regarding the probe vocabulary, the words of each category are single-word WordNet \cite{miller1995wordnet} lemmas below the root synsets of the category, ranked by frequency, and were not curated by hand, so the lists contain polysemous lemmas and proper names. For each seed, a random 20\% of every category forms the unseen words and the rest the probe words. The category-count ablation adds eight further categories (sport, instrument, clothing, furniture, vegetable, building, flower, and metal). The list below gives a representative subset, the 20 most frequent words of each of the eight core categories.

\begingroup\small\setlength{\parskip}{2pt}
\begin{description}\setlength{\itemsep}{0pt}\setlength{\parsep}{0pt}
\item[Animal (250 words):] world, man, game, head, young, human, baby, blue, dog, entire, female, ass, kid, stock, fish, male, queen, royal, tom, jack, \ldots
\item[Fruit (178 words):] date, apple, orange, empire, delicious, jonathan, cherry, fig, lemon, banana, berry, lime, baldwin, peach, strawberry, grape, blackberry, pineapple, citrus, mango, \ldots
\item[Vehicle (245 words):] car, stage, coach, ship, machine, engine, bottom, bus, van, technical, plane, flat, electric, boat, launch, sub, aircraft, truck, sam, bob, \ldots
\item[Tool (202 words):] bit, saw, bill, style, pick, die, square, file, upset, gang, knife, drove, punch, tap, blade, hammer, stamp, fork, starter, slice, \ldots
\item[Color (103 words):] white, black, red, gold, green, brown, coffee, rose, silver, wine, yellow, navy, pink, chocolate, bone, purple, pearl, olive, blonde, wheat, \ldots
\item[Country (248 words):] North, America, England, China, Union, Australia, India, Canada, France, Russia, Germany, Japan, USA, Israel, Mexico, Britain, Scotland, Italy, Ireland, kingdom, \ldots
\item[Emotion (249 words):] want, love, state, thing, care, hope, heart, fire, sex, security, wish, hate, pain, respect, charge, peace, weight, despite, scene, fear, \ldots
\item[Profession (225 words):] down, general, post, hand, major, private, official, federal, key, page, washington, minister, rock, officer, justice, judge, regular, master, doctor, professor, \ldots
\end{description}
\endgroup

\section{Round-Robin Study}
\label{app:rr}

In the design of this study, each of the five models of the pool (Table~\ref{tab:models}) analyzed the four other models, which gives 20 analyst--target pairs. Every pair was run once, in the default configuration of the LLM arm (sequential loop, adaptive strength, full digest, gain threshold 0.01 with the relaxed stage, and eight categories) with seed 0 and with the data, the limits, and the evaluation of the main study. The heuristic and hybrid arms were not run in this study. Every pair has its own output folder and run identifier, and all values below were computed from the summary, the layer log, and the evaluation files of the respective folder. The two pairs that the main study already contains (Gemma-2-2B on Qwen2.5-1.5B and on SmolLM2-1.7B, seed 0) were executed again and reproduced the earlier outputs exactly: the same layers were edited, and every value of the layer log and of the evaluation files is identical. The 20 runs took 4.7 to 42.7 minutes each and 374 minutes in total. In the tables of this section, Gemma, Llama, Phi, Qwen, and SmolLM2 stand for Gemma-2-2B, Llama-3.2-3B, Phi-3-mini, Qwen2.5-1.5B, and SmolLM2-1.7B; Table~\ref{tab:rrbase} gives the values of the five unedited targets.

\begin{table}[!tbp]
\centering
\caption{Round-robin study: the five unedited targets (seed 0). $L$: number of transformer blocks. Perplexity on the guard, held-out, and new-domain sentences; accuracy on the cloze task (probe and unseen words) and on the three benchmarks (200 examples each); factual-preference accuracy and truth margin (nats); purity under condition $[D]$, averaged over all layers and at the last layer.}
\label{tab:rrbase}
\small
\setlength{\tabcolsep}{4pt}
\begin{adjustbox}{max width=\textwidth}
\begin{tabular}{lccccccccccccc}
\toprule
 & & \multicolumn{3}{c}{Perplexity} & \multicolumn{2}{c}{Cloze} & \multicolumn{3}{c}{Benchmarks} & \multicolumn{2}{c}{Factual probe} & \multicolumn{2}{c}{$P^{[D]}$} \\
\cmidrule(lr){3-5}\cmidrule(lr){6-7}\cmidrule(lr){8-10}\cmidrule(lr){11-12}\cmidrule(lr){13-14}
Target & $L$ & guard & held-out & new domain & probe & unseen & HellaSwag & ARC-Easy & TruthfulQA & acc. & margin & all layers & last layer \\
\midrule
Gemma & 26 & 110.08 & 86.96 & 89.00 & 0.636 & 0.721 & 0.690 & 0.775 & 0.375 & 1.000 & 8.02 & 0.668 & 0.574 \\
Llama & 28 & 58.55 & 50.53 & 66.18 & 0.615 & 0.736 & 0.705 & 0.675 & 0.305 & 1.000 & 7.44 & 0.668 & 0.625 \\
Phi & 32 & 24.20 & 23.87 & 26.57 & 0.675 & 0.748 & 0.760 & 0.725 & 0.350 & 1.000 & 8.46 & 0.609 & 0.544 \\
Qwen & 28 & 49.11 & 43.72 & 61.36 & 0.638 & 0.710 & 0.620 & 0.750 & 0.275 & 0.975 & 6.92 & 0.659 & 0.599 \\
SmolLM2 & 24 & 48.47 & 40.77 & 55.06 & 0.614 & 0.736 & 0.715 & 0.640 & 0.245 & 1.000 & 8.18 & 0.656 & 0.596 \\
\bottomrule
\end{tabular}
\end{adjustbox}
\end{table}

As for the outcome, Table~\ref{tab:rr} lists all 20 pairs. In 16 pairs, the analyst changed its target. In the other four, the target was left unchanged: three analysts (Phi, Qwen, and SmolLM2) had no edit accepted on Gemma, and the two edits of SmolLM2 that were accepted on Qwen were neutral ($T_{\mathrm{sep}}$ operations with factor one, with a relative weight change below $10^{-15}$). In all 16 changed pairs, purity under $[D]$ rose over the edited layers, by 0.0023 to 0.0325 (mean 0.0135). It also rose under $[B]$ in all 16 pairs and under $[A]$ and $[C]$ in 15; the exception is the single edit of Qwen on Phi ($-0.0018$ and $-0.0014$). With the four unchanged pairs counted as zero, the mean over the 20 pairs is 0.0108 (95\% interval 0.0062 to 0.0154); the pairs share analysts and targets, so this interval is descriptive. At the last layer, purity under $[D]$ rose in 15 of the 16 pairs, by 0.003 to 0.076. Silhouette under $[D]$ rose in 15 pairs, anisotropy fell in 13, $\beta_0$ fell in 15, and the intrinsic dimension fell in 12, whereas $\beta_1$ fell in 8 pairs and rose in 7 (Table~\ref{tab:rrbeh}). Purity under $[D]$ rose at 363 of the 446 layers of the 16 changed models (Table~\ref{tab:rrdec}).


\begin{table}[!tbp]
\centering
\caption{Round-robin study, every analyst--target pair (default configuration of the LLM arm, seed 0). Edits: accepted edits; Strict: those accepted at the strict gain threshold. $\Delta P$: purity change under the four conditions, averaged over the edited layers; $\Delta P_L$: purity change at the last layer under $[D]$; $\Delta S$, $\Delta\kappa$: silhouette and anisotropy change under $[D]$. CKA$_w$, CKA$_t$: CKA to the original model on unseen words and on guard tokens. KL: cumulative KL (nats per token). $\Delta\mathrm{PPL}_g$, $\Delta\mathrm{PPL}_h$: perplexity change on the guard and on the held-out sentences (\%). A dash marks a pair without an accepted edit. $^{\ddagger}$Both accepted edits were neutral, and the target is unchanged.}
\label{tab:rr}
\small
\setlength{\tabcolsep}{3.5pt}
\begin{adjustbox}{max width=\textwidth}
\begin{tabular}{llcccccccccccccc}
\toprule
Analyst & Target & Edits & Strict & $\Delta P^{[A]}$ & $\Delta P^{[B]}$ & $\Delta P^{[C]}$ & $\Delta P^{[D]}$ & $\Delta P_L$ & $\Delta S$ & $\Delta\kappa$ & CKA$_w$ & CKA$_t$ & KL & $\Delta\mathrm{PPL}_g$ & $\Delta\mathrm{PPL}_h$ \\
\midrule
\multirow{4}{*}{Gemma} & Llama & 20 & 6 & 0.0091 & 0.0136 & 0.0103 & 0.0164 & $+$0.076 & 0.0327 & $-$0.0503 & 0.914 & 0.978 & 0.085 & $+$4.94 & $+$6.87 \\
 & Phi & 21 & 2 & 0.0132 & 0.0241 & 0.0104 & 0.0218 & $+$0.039 & 0.0119 & $-$0.0023 & 0.933 & 0.994 & 0.059 & $+$4.53 & $+$3.55 \\
 & Qwen & 13 & 3 & 0.0239 & 0.0349 & 0.0206 & 0.0287 & $+$0.061 & 0.0235 & $-$0.0300 & 0.897 & 0.975 & 0.089 & $+$5.00 & $+$6.45 \\
 & SmolLM2 & 13 & 5 & 0.0017 & 0.0145 & 0.0025 & 0.0111 & $+$0.026 & 0.0266 & $-$0.0210 & 0.902 & 0.984 & 0.092 & $+$4.81 & $+$6.82 \\
\addlinespace
\multirow{4}{*}{Llama} & Gemma & 6 & 0 & 0.0058 & 0.0107 & 0.0070 & 0.0103 & $+$0.034 & 0.0081 & $-$0.0219 & 0.942 & 0.992 & 0.039 & $+$2.70 & $+$1.39 \\
 & Phi & 20 & 0 & 0.0077 & 0.0172 & 0.0044 & 0.0103 & $+$0.036 & 0.0086 & $-$0.0123 & 0.965 & 0.996 & 0.070 & $+$4.97 & $+$3.93 \\
 & Qwen & 18 & 2 & 0.0217 & 0.0239 & 0.0191 & 0.0219 & $+$0.072 & 0.0252 & $-$0.0245 & 0.893 & 0.987 & 0.070 & $+$2.88 & $+$4.97 \\
 & SmolLM2 & 13 & 0 & 0.0074 & 0.0108 & 0.0074 & 0.0142 & $+$0.034 & 0.0123 & $-$0.0074 & 0.949 & 0.986 & 0.099 & $+$4.61 & $+$7.91 \\
\addlinespace
\multirow{4}{*}{Phi} & Gemma & 0 & 0 & -- & -- & -- & -- & -- & -- & -- & -- & -- & -- & -- & -- \\
 & Llama & 27 & 1 & 0.0101 & 0.0021 & 0.0089 & 0.0054 & $+$0.024 & 0.0023 & 0.0007 & 0.936 & 0.996 & 0.037 & $+$1.32 & $+$2.05 \\
 & Qwen & 21 & 2 & 0.0172 & 0.0100 & 0.0114 & 0.0068 & $+$0.038 & 0.0143 & $-$0.0199 & 0.896 & 0.992 & 0.076 & $+$4.84 & $+$6.42 \\
 & SmolLM2 & 9 & 5 & 0.0102 & 0.0129 & 0.0062 & 0.0073 & $+$0.026 & 0.0056 & 0.0063 & 0.969 & 0.960 & 0.068 & $+$4.73 & $+$5.42 \\
\addlinespace
\multirow{4}{*}{Qwen} & Gemma & 0 & 0 & -- & -- & -- & -- & -- & -- & -- & -- & -- & -- & -- & -- \\
 & Llama & 1 & 1 & 0.0122 & 0.0246 & 0.0145 & 0.0325 & $+$0.009 & 0.0151 & $-$0.0282 & 0.961 & 0.989 & 0.006 & $-$0.02 & $+$0.37 \\
 & Phi & 1 & 0 & $-$0.0018 & 0.0059 & $-$0.0014 & 0.0029 & $+$0.003 & 0.0016 & 0.0019 & 0.992 & 0.999 & 0.002 & $+$0.24 & $+$0.12 \\
 & SmolLM2 & 1 & 1 & 0.0086 & 0.0040 & 0.0127 & 0.0062 & $-$0.005 & $-$0.0056 & $-$0.0126 & 0.989 & 0.999 & 0.034 & $+$1.67 & $+$1.97 \\
\addlinespace
\multirow{4}{*}{SmolLM2} & Gemma & 0 & 0 & -- & -- & -- & -- & -- & -- & -- & -- & -- & -- & -- & -- \\
 & Llama & 10 & 1 & 0.0011 & 0.0068 & 0.0032 & 0.0023 & $+$0.033 & 0.0194 & $-$0.0360 & 0.885 & 0.993 & 0.032 & $+$4.99 & $+$3.35 \\
 & Phi & 7 & 6 & 0.0113 & 0.0153 & 0.0127 & 0.0181 & $+$0.040 & 0.0082 & $-$0.0111 & 0.952 & 0.991 & 0.037 & $+$4.87 & $+$3.69 \\
 & Qwen & 2$^{\ddagger}$ & 0 & 0.0000 & 0.0000 & 0.0000 & 0.0000 & $+$0.000 & 0.0000 & 0.0000 & 1.000 & 1.000 & 0.000 & $+$0.00 & $+$0.00 \\
\bottomrule
\end{tabular}
\end{adjustbox}
\end{table}

\begin{table}[!tbp]
\centering
\caption{Round-robin study: behavior and topology of the 16 pairs in which the target changed (every change is zero for the other four pairs). Changes are edited minus original. Cloze: accuracy on probe and on unseen words; benchmarks with 200 examples each (standard error about 0.035); Fact acc.: factual-preference accuracy; Margin: truth margin (nats). $\Delta\beta_0$, $\Delta\beta_1$, $\Delta\mathrm{ID}$: change of the Betti numbers at the 25th-percentile scale and of the Two-NN intrinsic dimension under $[D]$, averaged over the edited layers. $\Delta\mathrm{PPL}_n$: perplexity change on the new-domain sentences (\%).}
\label{tab:rrbeh}
\small
\setlength{\tabcolsep}{4pt}
\begin{adjustbox}{max width=\textwidth}
\begin{tabular}{llccccccccccc}
\toprule
 & & \multicolumn{2}{c}{$\Delta$ cloze} & \multicolumn{3}{c}{$\Delta$ benchmark accuracy} & \multicolumn{2}{c}{Factual probe} & \multicolumn{3}{c}{Topology under $[D]$} & \\
\cmidrule(lr){3-4}\cmidrule(lr){5-7}\cmidrule(lr){8-9}\cmidrule(lr){10-12}
Analyst & Target & probe & unseen & HellaSwag & ARC-Easy & TruthfulQA & $\Delta$ acc. & $\Delta$ margin & $\Delta\beta_0$ & $\Delta\beta_1$ & $\Delta\mathrm{ID}$ & $\Delta\mathrm{PPL}_n$ \\
\midrule
\multirow{4}{*}{Gemma} & Llama & $+$0.054 & $-$0.018 & $-$0.005 & $-$0.040 & $+$0.015 & 0.000 & $-$0.390 & $-$1.40 & $-$1.20 & $-$0.92 & $-$1.46 \\
 & Phi & $+$0.003 & $+$0.009 & $-$0.010 & $+$0.000 & $-$0.015 & 0.000 & $-$0.395 & $-$1.76 & $+$0.19 & $-$0.19 & $+$3.70 \\
 & Qwen & $+$0.068 & $+$0.050 & $+$0.010 & $-$0.030 & $+$0.000 & 0.000 & $-$0.256 & $-$2.44 & $-$0.10 & $+$0.07 & $+$1.76 \\
 & SmolLM2 & $+$0.050 & $+$0.012 & $+$0.005 & $+$0.005 & $+$0.000 & 0.000 & $-$0.563 & $-$4.28 & $-$0.36 & $-$0.04 & $+$4.17 \\
\addlinespace
\multirow{4}{*}{Llama} & Gemma & $+$0.042 & $-$0.012 & $-$0.005 & $-$0.005 & $-$0.005 & 0.000 & $-$0.182 & $-$0.17 & $-$1.06 & $-$0.03 & $-$0.18 \\
 & Phi & $+$0.050 & $+$0.044 & $-$0.005 & $-$0.015 & $-$0.010 & 0.000 & $-$0.024 & $-$2.95 & $+$0.55 & $-$0.18 & $+$4.57 \\
 & Qwen & $+$0.004 & $-$0.003 & $-$0.005 & $-$0.010 & $+$0.005 & 0.000 & $-$0.423 & $-$3.59 & $-$0.96 & $-$0.18 & $-$1.87 \\
 & SmolLM2 & $+$0.078 & $+$0.003 & $-$0.015 & $-$0.015 & $-$0.005 & 0.000 & $-$0.328 & $-$1.67 & $+$0.54 & $-$0.07 & $+$4.30 \\
\addlinespace
\multirow{3}{*}{Phi} & Llama & $+$0.035 & $-$0.009 & $+$0.020 & $-$0.025 & $+$0.005 & 0.000 & $-$0.096 & $-$1.07 & $+$0.12 & $-$0.04 & $-$0.88 \\
 & Qwen & $+$0.032 & $+$0.029 & $+$0.000 & $-$0.010 & $-$0.010 & 0.000 & $-$0.105 & $-$3.97 & $-$0.51 & $-$0.06 & $-$0.17 \\
 & SmolLM2 & $+$0.042 & $+$0.035 & $-$0.025 & $+$0.005 & $+$0.000 & 0.000 & $-$0.149 & $-$2.81 & $+$1.26 & $-$0.71 & $+$3.93 \\
\addlinespace
\multirow{3}{*}{Qwen} & Llama & $+$0.020 & $+$0.009 & $+$0.015 & $-$0.015 & $+$0.000 & 0.000 & $+$0.033 & $-$5.00 & $+$19.67 & $+$0.30 & $-$0.33 \\
 & Phi & $+$0.000 & $+$0.000 & $+$0.000 & $+$0.000 & $-$0.010 & 0.000 & $-$0.002 & $-$1.33 & $+$0.00 & $+$0.14 & $+$0.81 \\
 & SmolLM2 & $+$0.010 & $+$0.026 & $-$0.015 & $+$0.000 & $+$0.005 & 0.000 & $-$0.064 & $+$5.67 & $+$5.33 & $+$0.11 & $+$1.26 \\
\addlinespace
\multirow{2}{*}{SmolLM2} & Llama & $-$0.027 & $-$0.056 & $+$0.010 & $-$0.025 & $+$0.000 & 0.000 & $-$0.090 & $-$3.87 & $-$1.17 & $-$0.26 & $-$1.66 \\
 & Phi & $-$0.007 & $+$0.018 & $-$0.015 & $-$0.010 & $+$0.000 & 0.000 & $-$0.081 & $-$1.76 & $-$0.14 & $-$0.21 & $-$0.70 \\
\bottomrule
\end{tabular}
\end{adjustbox}
\end{table}

Regarding identity, behavior, and costs, in the 16 changed pairs, the CKA to the original model was 0.885 to 0.992 on unseen words and 0.960 to 0.999 on guard tokens, and the cumulative KL was 0.002 to 0.099 nats per token, against a budget of 0.30. The guard-set perplexity changed by $-0.02\%$ to $+5.00\%$ and the held-out perplexity by $+0.12\%$ to $+7.91\%$. The largest benchmark change was 0.040 in magnitude (ARC-Easy, Gemma on Llama), and the factual-preference accuracy did not change in any pair. The changes on HellaSwag and TruthfulQA had no consistent sign. ARC-Easy accuracy, however, fell in 11 of the 16 pairs and rose in 2 (mean $-0.012$; one-sample $t$-test over the 16 pairs, $p=0.002$), so a small systematic cost on this benchmark cannot be excluded, although every single change is within about one standard error. The truth margin fell in 15 of the 16 pairs, by at most 0.563 nats (6.9\% of the baseline margin). Cloze accuracy rose on probe words in 13 pairs and on unseen words in 10.

Turning to the analysts, Table~\ref{tab:rranalyst} summarizes the five analysts. Gemma and Llama changed all four of their targets, Phi and Qwen three, and SmolLM2 two. Gemma, Llama, and Phi had 6 to 27 edits accepted on every target that they changed. Qwen had a single edit accepted on each of three targets, and 84 of its 110 layer decisions ended without a proposal that passed. SmolLM2 always proposed four $T_{\mathrm{sep}}$ operations per edit and had 7 and 10 edits accepted on the two targets that it changed. The mean purity gain under $[D]$ over the four targets was largest for Gemma (0.0195) and Llama (0.0142). The ranking does not follow the parameter count: Phi, the largest model, had a mean gain of 0.0049, and the mean of Qwen (0.0104) rests on one edit of one layer of Llama. Gemma was the only analyst that often used $T_{\mathrm{pcs}}$ (24 of its 67 edits); Llama and Phi almost always used $T_{\mathrm{sep}}$. Most accepted edits passed at the relaxed gain threshold (168 of 203). A $T_{\mathrm{wh}}$ operation was accepted six times (four times for Phi and twice for Qwen; three of them had rank one and are neutral), and a $T_{\mathrm{ctr}}$ operation never.

\begin{table}[!tbp]
\centering
\caption{Round-robin study: summary per analyst over its four targets. Params: parameters of the analyst (billions). Changed: targets that the analyst changed. Edits, Strict, Relaxed: accepted edits in total and by stage. $T_{\mathrm{sep}}$, $T_{\mathrm{pcs}}$, $T_{\mathrm{wh}}$: accepted centroid-axis scalings, principal-component rescalings, and partial spectral whitenings. Operations/edit: operations per accepted edit. $\Delta P^{[D]}$: mean purity change under $[D]$ over the four targets (an unchanged target counts as zero) and over the changed targets only. $\Delta S^{[D]}$, KL, and $\Delta\mathrm{PPL}_h$: means over the four targets. Min.: total run time (minutes).}
\label{tab:rranalyst}
\small
\setlength{\tabcolsep}{4pt}
\begin{adjustbox}{max width=\textwidth}
\begin{tabular}{lccccccccccccccc}
\toprule
 & & & & & & \multicolumn{3}{c}{Operations} & & \multicolumn{2}{c}{$\Delta P^{[D]}$} & & & & \\
\cmidrule(lr){7-9}\cmidrule(lr){11-12}
Analyst & Params & Changed & Edits & Strict & Relaxed & $T_{\mathrm{sep}}$ & $T_{\mathrm{pcs}}$ & $T_{\mathrm{wh}}$ & Operations/edit & 4 targets & changed & $\Delta S^{[D]}$ & KL & $\Delta\mathrm{PPL}_h$ & Min. \\
\midrule
Gemma & 2.6 & 4/4 & 67 & 16 & 51 & 43 & 24 & 0 & 1.00 & 0.0195 & 0.0195 & 0.0237 & 0.081 & $+$5.93 & 107.9 \\
Llama & 3.2 & 4/4 & 57 & 2 & 55 & 55 & 2 & 0 & 1.00 & 0.0142 & 0.0142 & 0.0136 & 0.070 & $+$4.55 & 68.7 \\
Phi & 3.8 & 3/4 & 57 & 8 & 49 & 54 & 3 & 4 & 1.07 & 0.0049 & 0.0065 & 0.0055 & 0.045 & $+$3.47 & 81.3 \\
Qwen & 1.5 & 3/4 & 3 & 2 & 1 & 4 & 1 & 2 & 2.33 & 0.0104 & 0.0139 & 0.0028 & 0.010 & $+$0.62 & 39.2 \\
SmolLM2 & 1.7 & 2/4 & 19 & 7 & 12 & 76 & 0 & 0 & 4.00 & 0.0051 & 0.0102 & 0.0069 & 0.017 & $+$1.76 & 76.9 \\
\bottomrule
\end{tabular}
\end{adjustbox}
\end{table}

As for the targets, Llama and Phi, which are larger than the targets of the main study, were changed by all four of their analysts, Qwen by three, and SmolLM2 by four. Gemma was the hardest target. Of the 104 layer decisions of the four analysts on Gemma, 44 ended with the weight budget exceeded and 14 with the CKA limit broken after baking, and only six edits of Llama were accepted, all at the relaxed threshold and at a reduced strength (mean $\alpha$ of 0.33). On Gemma, the baked update needed a much larger weight change for the same requested change: over the baked edits of all analysts, accepted or rejected, the median relative weight change was 0.21, against 0.014 to 0.036 on the other four targets. The verification thus failed closed: where no edit met the limits, the target was left unchanged. We did not investigate why the feed-forward output projection of Gemma is this sensitive.


\begin{table}[!tbp]
\centering
\caption{Round-robin study: layer decisions and per-edit statistics of every pair. Layers: layer decisions; Appl.: applied edits (Strict and Rel.: accepted at the strict and at the relaxed gain threshold); None: no proposal passed; Rev.: weights reverted because the baked edit broke the KL, perplexity, or CKA limit at every tested strength; Weight: weight budget exceeded. $T_{\mathrm{sep}}$, $T_{\mathrm{pcs}}$, $T_{\mathrm{wh}}$: accepted centroid-axis scalings, principal-component rescalings, and partial spectral whitenings. Operations/edit: operations per accepted edit. $\alpha$: mean strength. $R^2$: median fit of the ridge solution. $\lVert\Delta\mathbf{W}\rVert$: largest relative weight change. CKA$_t$: lowest CKA on guard tokens of a single edit (edit) and of the final model at any layer (final). Improved: layers whose purity under $[D]$ rose. $\Delta P^{[D]}_{\mathrm{all}}$: purity change under $[D]$ averaged over all layers.}
\label{tab:rrdec}
\small
\setlength{\tabcolsep}{3pt}
\begin{adjustbox}{max width=\textwidth}
\begin{tabular}{llcccccccccccccccccc}
\toprule
 & & \multicolumn{7}{c}{Layer decisions} & \multicolumn{3}{c}{Operations} & & & & & \multicolumn{2}{c}{CKA$_t$} & & \\
\cmidrule(lr){3-9}\cmidrule(lr){10-12}\cmidrule(lr){17-18}
Analyst & Target & Layers & Appl. & Strict & Rel. & None & Rev. & Weight & $T_{\mathrm{sep}}$ & $T_{\mathrm{pcs}}$ & $T_{\mathrm{wh}}$ & Operations/edit & $\alpha$ & $R^2$ & $\lVert\Delta\mathbf{W}\rVert$ & edit & final & Improved & $\Delta P^{[D]}_{\mathrm{all}}$ \\
\midrule
\multirow{4}{*}{Gemma} & Llama & 28 & 20 & 6 & 14 & 8 & 0 & 0 & 13 & 7 & 0 & 1.00 & 1.13 & 0.9994 & 0.085 & 0.989 & 0.941 & 27/28 & 0.0236 \\
 & Phi & 32 & 21 & 2 & 19 & 9 & 1 & 1 & 13 & 8 & 0 & 1.00 & 0.99 & 0.9991 & 0.068 & 0.996 & 0.990 & 27/32 & 0.0204 \\
 & Qwen & 28 & 13 & 3 & 10 & 15 & 0 & 0 & 11 & 2 & 0 & 1.00 & 1.16 & 0.9994 & 0.089 & 0.983 & 0.918 & 28/28 & 0.0357 \\
 & SmolLM2 & 24 & 13 & 5 & 8 & 9 & 2 & 0 & 6 & 7 & 0 & 1.00 & 1.07 & 0.9997 & 0.084 & 0.989 & 0.948 & 23/24 & 0.0131 \\
\addlinespace
\multirow{4}{*}{Llama} & Gemma & 26 & 6 & 0 & 6 & 12 & 5 & 3 & 6 & 0 & 0 & 1.00 & 0.33 & 0.9993 & 0.076 & 0.982 & 0.974 & 25/26 & 0.0098 \\
 & Phi & 32 & 20 & 0 & 20 & 12 & 0 & 0 & 18 & 2 & 0 & 1.00 & 0.96 & 0.9992 & 0.084 & 0.994 & 0.992 & 32/32 & 0.0110 \\
 & Qwen & 28 & 18 & 2 & 16 & 10 & 0 & 0 & 18 & 0 & 0 & 1.00 & 1.19 & 0.9996 & 0.097 & 0.989 & 0.971 & 28/28 & 0.0231 \\
 & SmolLM2 & 24 & 13 & 0 & 13 & 10 & 1 & 0 & 13 & 0 & 0 & 1.00 & 1.07 & 0.9991 & 0.084 & 0.994 & 0.678 & 24/24 & 0.0111 \\
\addlinespace
\multirow{4}{*}{Phi} & Gemma & 26 & 0 & 0 & 0 & 1 & 4 & 21 & 0 & 0 & 0 & -- & -- & -- & -- & -- & -- & -- & -- \\
 & Llama & 28 & 27 & 1 & 26 & 1 & 0 & 0 & 28 & 0 & 0 & 1.04 & 1.14 & 0.9994 & 0.038 & 0.997 & 0.993 & 19/28 & 0.0053 \\
 & Qwen & 28 & 21 & 2 & 19 & 7 & 0 & 0 & 18 & 3 & 3 & 1.14 & 1.13 & 0.9995 & 0.054 & 0.995 & 0.986 & 16/28 & 0.0043 \\
 & SmolLM2 & 24 & 9 & 5 & 4 & 14 & 1 & 0 & 8 & 0 & 1 & 1.00 & 0.88 & 0.9986 & 0.077 & 0.998 & 0.915 & 23/24 & 0.0050 \\
\addlinespace
\multirow{4}{*}{Qwen} & Gemma & 26 & 0 & 0 & 0 & 3 & 5 & 18 & 0 & 0 & 0 & -- & -- & -- & -- & -- & -- & -- & -- \\
 & Llama & 28 & 1 & 1 & 0 & 27 & 0 & 0 & 0 & 1 & 1 & 2.00 & 1.00 & 0.9995 & 0.087 & 0.989 & 0.989 & 26/28 & 0.0083 \\
 & Phi & 32 & 1 & 0 & 1 & 31 & 0 & 0 & 1 & 0 & 1 & 2.00 & 1.25 & 0.9990 & 0.052 & 0.999 & 0.999 & 19/32 & 0.0006 \\
 & SmolLM2 & 24 & 1 & 1 & 0 & 23 & 0 & 0 & 3 & 0 & 0 & 3.00 & 0.50 & 0.9995 & 0.043 & 0.999 & 0.355 & 4/24 & $-$0.0045 \\
\addlinespace
\multirow{4}{*}{SmolLM2} & Gemma & 26 & 0 & 0 & 0 & 24 & 0 & 2 & 0 & 0 & 0 & -- & -- & -- & -- & -- & -- & -- & -- \\
 & Llama & 28 & 10 & 1 & 9 & 18 & 0 & 0 & 40 & 0 & 0 & 4.00 & 0.88 & 0.9993 & 0.086 & 0.995 & 0.990 & 12/28 & 0.0025 \\
 & Phi & 32 & 7 & 6 & 1 & 23 & 2 & 0 & 28 & 0 & 0 & 4.00 & 0.48 & 0.9987 & 0.094 & 0.992 & 0.976 & 30/32 & 0.0231 \\
 & Qwen & 28 & 2 & 0 & 2 & 26 & 0 & 0 & 8 & 0 & 0 & 4.00 & 1.50 & 0.9196 & 0.000 & 1.000 & 1.000 & 0/28 & 0.0000 \\
\bottomrule
\end{tabular}
\end{adjustbox}
\end{table}

In addition, for Qwen and SmolLM2, the heuristic arm of the main study with seed 0 gives a reference: its purity gain under $[D]$ was 0.0285 and 0.0150, at a cumulative KL of 0.115 and 0.121. On Qwen, Gemma (0.0287 at a KL of 0.089) and Llama (0.0219 at 0.070) came close to this gain at a lower divergence, and Phi stayed below it (0.0068). On SmolLM2, all four analysts stayed below the heuristic (0.0062 to 0.0142). The heuristic was not run on the three other targets.

Finally, three observations on failure cases qualify the positive outcome. (i)~The single edit of Qwen on SmolLM2 (three $T_{\mathrm{sep}}$ operations at layer 1) raised purity under $[D]$ at that layer by 0.006 but lowered it at 20 of the 24 layers (mean over all layers $-0.0045$; last layer $-0.005$) and reduced the CKA on guard tokens at layer 8 to 0.355, although the edit met every per-edit limit and the cumulative KL was 0.034. The edits of Llama on SmolLM2 also left a low CKA at layer 8 (0.678), as did the heuristic in the main study, so this layer of SmolLM2 is sensitive to upstream edits. (ii)~The relaxed gain threshold of $10^{-6}$ admitted edits that change nothing: the two edits of SmolLM2 on Qwen, and two edits of the main study (a $T_{\mathrm{wh}}$ operation of rank one and a $T_{\mathrm{pcs}}$ operation with factor one). A minimum weight change would remove such edits from the counts. (iii)~Averaged over the 16 changed pairs, purity under $[D]$ rose most for tool ($+0.035$), animal ($+0.032$), and profession ($+0.030$) and fell slightly for vehicle ($-0.005$) and fruit ($-0.002$).

\section{Implementation Details and Comparison with Related Work}
\label{app:related}

Regarding the implementation, all runs used a single NVIDIA H100 GPU with 80\,GB of memory (driver 570.133.20) with PyTorch 2.5.1 (CUDA 12.1), transformers 4.57.6, ripser 0.6.15, and scikit-learn 1.7.2.
Targets were held in 32-bit floating point and the analyst in bfloat16. The analyst's reply was constrained to begin with the opening of the JSON object, was read leniently, and was corrected by clipping values to their ranges and dropping operations that could not be applied. A default sequential run took 16.5 to 19.0 min (Qwen) and 15.5 to 17.5 min (SmolLM2) in the LLM arm, 13.1 to 14.6 and 14.8 to 15.7 min with the heuristic, and 18.2 to 20.3 and 17.8 to 22.4 min with the hybrid arm, including the before/after evaluation; a parallel run took 7.9 to 10.0 min, and a run of the round robin 4.7 to 42.7 min. Persistent homology and intrinsic dimension were computed on a class-balanced subsample of at most 300 probe words, and the KL and token-level CKA limits on the 24 and the first 12 guard sentences.

\begin{table}[t]
\centering
\caption{Attributes of the closest reviewed works, as stated in their abstracts or main text. ``Fitted component'' indicates what, if anything, the method fits.}
\label{tab:related}
\small
\begin{tabular}{>{\raggedright\arraybackslash}p{3.6cm}>{\raggedright\arraybackslash}p{3.4cm}>{\raggedright\arraybackslash}p{3.4cm}>{\raggedright\arraybackslash}p{3.3cm}}
\toprule
Work & Input to the editor & Edit target & Fitted component \\
\midrule
Weight-Space Programming \cite{takehi2026weightspace} & Raw weights as text (with tools at 0.8B parameters) & Weights, for a behavioral goal & None (no gradient training) \\
ROME \cite{meng2022locating} & One factual association & One feed-forward weight, rank one & Closed-form update \\
LEACE \cite{belrose2023leace} & Concept labels & Linear map that erases one concept & Closed-form map \\
Activation steering \cite{li2023inference,turner2023steering,rimsky2024steering} & Contrast activations or probes & Hidden states at inference & Probes in \cite{li2023inference}; none (mean difference) in \cite{turner2023steering,rimsky2024steering} \\
Curveball steering \cite{raval2026curveball} & Activation geometry & Hidden states, nonlinear & Kernel PCA on activations \\
Riemannian-manifold steering \cite{oozeer2026riemannian} & Output distances & Hidden states along geodesics & Learned encoder \\
Topology-aware weight baking (ours) & Compressed topological descriptors & Layer geometry, baked into weights & Closed-form ridge regression \\
\bottomrule
\end{tabular}
\end{table}

\end{document}